\pdfoutput=1

\documentclass[11pt]{article}

\usepackage[preprint]{acl}

\usepackage{times}
\usepackage{latexsym}
\usepackage{booktabs}
\usepackage{makecell}
\usepackage{pifont}
\usepackage{threeparttable}
\usepackage{multirow}
\usepackage{tikz}
\usetikzlibrary{positioning,shapes.multipart,fit,backgrounds}
\usepackage{genealogytree}
\usepackage{hyperref}
\usepackage{amsmath}
\usepackage[T1]{fontenc}

\usepackage[utf8]{inputenc}

\usepackage{microtype}

\usepackage{inconsolata}

\usepackage{graphicx}

\definecolor{lightblue}{HTML}{ADD8E6}
\definecolor{starlight}{HTML}{F9F7ED}
\definecolor{lightgray}{HTML}{EEEEEE}

\newcommand{\cmark}{\ding{51}}

\title{Towards Controllable Speech Synthesis in the Era of Large Language Models: A Systematic Survey}

\author{
 \textbf{Tianxin Xie\textsuperscript{1}},
 \textbf{Yan Rong\textsuperscript{1}},
 \textbf{Pengfei Zhang\textsuperscript{1}},
 \textbf{Wenwu Wang\textsuperscript{2}},
 \textbf{Li Liu\textsuperscript{1}},
\\
 \textsuperscript{1}The Hong Kong University of Science and Technology (Guangzhou),
 \textsuperscript{2}University of Surrey
\\
 \small{
   \textbf{Correspondence:} \href{mailto:email@domain}{avrillliu@hkust-gz.edu.cn}
 }
}

\begin{document}
\maketitle
\begin{abstract}
Text-to-speech (TTS) has advanced from generating natural-sounding speech to enabling fine-grained control over attributes like emotion, timbre, and style. Driven by rising industrial demand and breakthroughs in deep learning, e.g., diffusion and large language models (LLMs), controllable TTS has become a rapidly growing research area. This survey provides \textbf{the first} comprehensive review of controllable TTS methods, from traditional control techniques to emerging approaches using natural language prompts. We categorize model architectures, control strategies, and feature representations, while also summarizing challenges, datasets, and evaluations in controllable TTS. This survey aims to guide researchers and practitioners by offering a clear taxonomy and highlighting future directions in this fast-evolving field.
One can visit \url{https://github.com/imxtx/awesome-controllabe-speech-synthesis} for a comprehensive paper list and updates.
\end{abstract}

\section{Introduction} \label{sec:intro}

Speech synthesis, also known as text-to-speech (TTS), aims to generate human-like speech from text~\cite{dutoit1997introduction}, and has found broad applications in personal assistants~\cite{lopez2018alexa}, entertainment~\cite{wang2019comic}, and robotics~\cite{marge2022spoken}. 
Recently, the success of large language models such as ChatGPT~\cite{openai2022chatgpt} has renewed interest in TTS for natural and intuitive human-computer interaction.
Meanwhile, fine-grained control over speech attributes, such as emotion, timbre, and style, has become a key focus in both academia and industry, unlocking more expressive and personalized voice generation.

In the past decade, deep learning has driven remarkable advances in TTS, enabling high-quality synthesis~\cite{tan2024naturalspeech,ren2019fastspeech,du2024cosyvoice} and stronger control over speech attributes~\cite{wang2018style,li2021controllable,zhou2024voxinstruct}.
Recent methods have expanded TTS to multi-modal inputs, including images~\cite{rong2025seeing} and videos~\cite{choi2023diffv2s}.
Meanwhile, the rise of LLMs~\cite{zhao2023survey} has enabled controllable TTS guided by language prompts~\cite{guo2023prompttts,huang2024instructspeech}, opening new possibilities for customized voice synthesis.
Integrating TTS into LLMs has also gained extensive attention~\cite{peng2024survey}.
This rapid progress underscores the need for a comprehensive and timely survey to clarify current trends and guide future directions in controllable TTS.

\begin{figure*}[t]
    \centering
    \includegraphics[width=\linewidth]{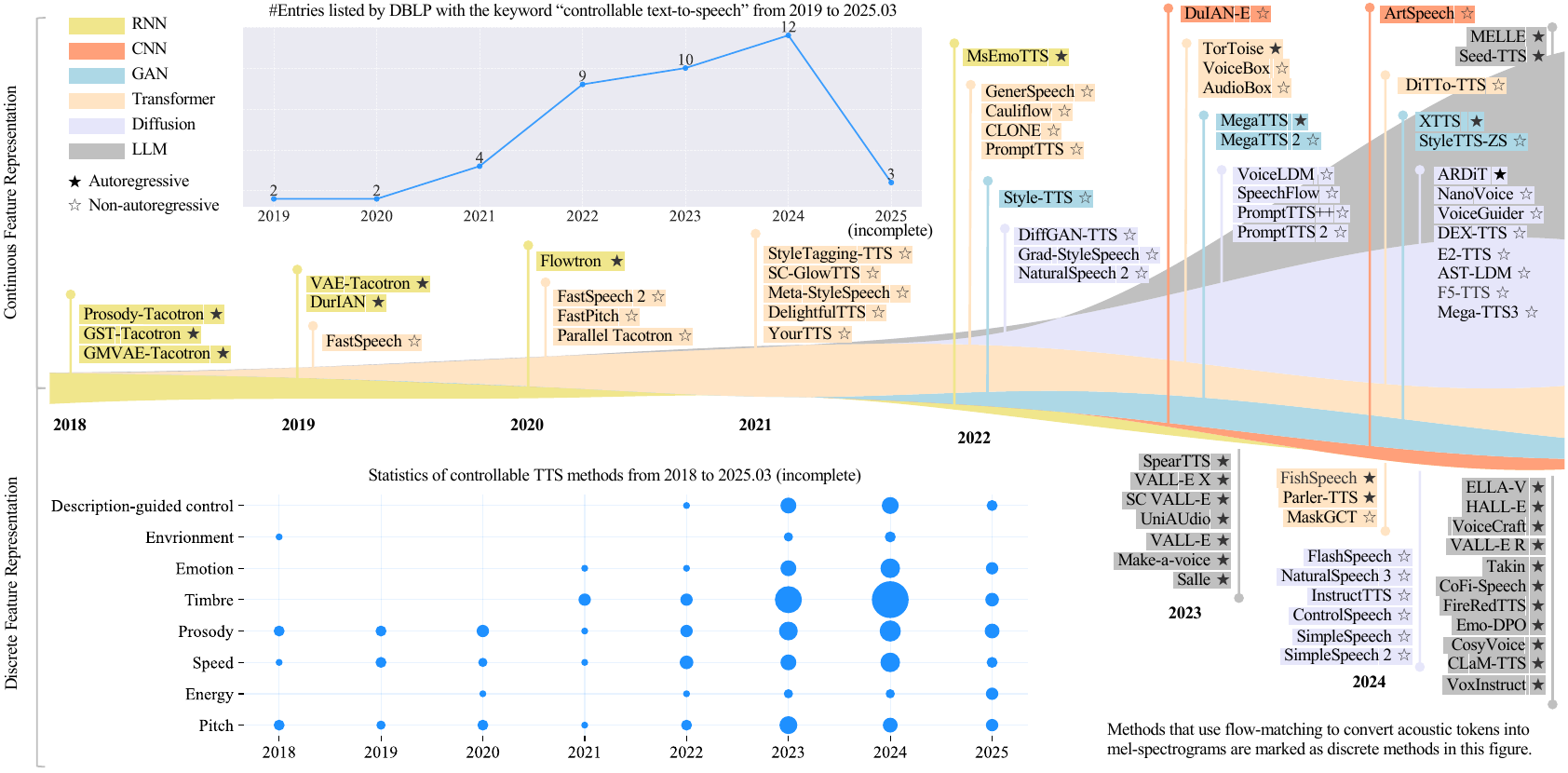}
    \caption{Recent trends in controllable TTS regarding architectures, feature representations, and control abilities.}
    \label{fig:summary}
\end{figure*}

While several surveys have examined parametric~\cite{zen2009statistical} and deep learning–based TTS~\cite{triantafyllopoulos2023overview_survey}, they overlook TTS controllability and recent advances such as description–based methods~\cite{guo2023prompttts,yamamoto2024description}.
The key differences between our survey and earlier work are: \textbf{1) Different Scope}:
~\citet{klatt1987review} provided the first review of formant-based, concatenative, and articulatory TTS, with a strong focus on text analysis.
Later, ~\citet{tabet2011speech,king2014measuring} explored statistics-based techniques.
With the advent of neural networks, ~\citet{ning2019review,tan2021survey,zhang2023survey} surveyed neural model–based TTS, focusing on acoustics and vocoders.
However, they rarely discuss the controllability.
\textbf{2) Closer to Current Demands}:
The need for controllable TTS is rapidly growing in industries like filmmaking, gaming, robotics, and virtual assistants.
Yet, existing surveys rarely explore the gaps between current control techniques and real-world demands.

To fill this gap, we present the first comprehensive survey of emerging controllable TTS methods.
We first define the core tasks (Sec.~\ref{sec:tasks}) and, as shown in Fig.~\ref{fig:summary}, trace the evolution of methods across model architectures (Sec.~\ref{sec:archs}), control strategies (Sec.~\ref{sec:strategies}), and feature representations (Sec.~\ref{sec:features}).
We further summarize relevant datasets and evaluation metrics (Sec.~\ref{sec:datasets_eval}), and discuss current challenges and future research directions (Sec.~\ref{sec:directions}).
For a history of controllable TTS and an overview of the TTS pipeline, see Appendices \ref{sec:appd_history} and \ref{sec:appd_pipeline}.

\section{Main Tasks in Controllable TTS} \label{sec:tasks}


\textbf{Prosody Control} is the most basic task in controllable TTS, aiming to manipulate low-level acoustic features such as pitch~\cite{lancucki2021fastpitch}, duration~\cite{wang2025sparktts}, and energy~\cite{chen2025drawspeech}. Prosody control ensures naturalness and expressiveness in TTS and is essential for rendering emphasis, rhythm, and nuance in speech.

\textbf{Timbre Control} aims to manipulate the acoustic characteristics that define voice quality (e.g., gender, age, nasality), enabling control over how a voice sounds beyond content and prosody. It supports personalized TTS~\cite{du2024cosyvoice}, voice conversion~\cite{zhang2025vevo}, and speaker identity editing~\cite{huang2024instructspeech}.
 
\textbf{Emotion Control} aims to enable the synthesis of emotional speech by manipulating the affective state of the generated voice~\cite{kim2021expressive}. This improves human-computer interaction, storytelling~\cite{rong2025dopamine}, and supports emotionally adaptive systems such as virtual assistants.

\textbf{Style Control} aims to control higher-level attributes of speech such as tone, formality, and discourse mode (e.g., newscast)~\cite{zhou2024voxinstruct,yang2024instructtts}. This is critical for adapting the speaking behavior of TTS systems to different contexts, audiences, and communication goals.

\textbf{Language Control} aims to enable TTS systems to synthesize speech in multiple languages~\cite{zhang2023speak}, dialects~\cite{di2024bailing}, or code-switched contexts~\cite{chen2024f5}. It facilitates cross-lingual communication, multilingual agents, and regionally tailored speech applications.

\textbf{Environment Control} aims to simulate the acoustic characteristics of a specific setting, such as a park, office, or seaside, by conditioning synthesis on background noise and spatial cues~\cite{lee2024voiceldm,kim2024speak}. Speech environment control is useful in filmmaking and audiobooks.


\section{Methods in Controllable TTS} \label{sec:methods}

\begin{table*}[!t]
\renewcommand{\arraystretch}{0.6}
\centering
\resizebox{1.0\textwidth}{!}{
\begin{threeparttable}
    \begin{tabular}{cccccccccccccc}
    \toprule
    \multicolumn{1}{c}{\multirow{2}{*}[-3pt]{Method (Non-autoregressive)}} & \multicolumn{1}{c}{\multirow{2}{*}[-3pt]{\makecell{Zero-shot}}} & \multicolumn{8}{c}{Controlability} & \multicolumn{2}{c}{Model Architectures} & \multicolumn{1}{c}{\multirow{2}{*}[-3pt]{\makecell{Feature}}} & \multicolumn{1}{c}{\multirow{2}{*}[-3pt]{\makecell{Release}}} \\ 
    \cmidrule(rl){3-10} \cmidrule(rl){11-12}
    & & Pit. & Ene. & Spe. & Pro. & Tim. & Emo. & Env. & Des. & Acoustic Model & Vocoder & & \\
    \midrule
    FastSpeech~\cite{ren2019fastspeech} &  &  &  & \cmark & \cmark &  &  &  &  & Transformer & WaveGlow~\cite{prenger2019waveglow} & MelS & 2019.05 \\
    FastSpeech 2~\cite{ren2020fastspeech} &  & \cmark & \cmark & \cmark & \cmark &  &  &  &  & Transformer & Parallel WaveGAN~\cite{yamamoto2020parallelwavegan} & MelS & 2020.06 \\
    FastPitch~\cite{lancucki2021fastpitch} &  & \cmark &  &  & \cmark &  &  &  &  & Transformer & WaveGlow & MelS & 2020.06 \\
    Parallel Tacotron~\cite{elias2021parallel} &  &  &  &  & \cmark &  &  &  &  & Transformer + VAE + CNN & WaveRNN~\cite{kalchbrenner2018wavernn} & MelS & 2020.10 \\
    StyleTagging-TTS~\cite{kim2021expressive} & \cmark &  &  &  &  & \cmark & \cmark &  &  & Transformer + CNN & HiFi-GAN~\cite{kong2020hifigan} & MelS & 2021.04 \\
    SC-GlowTTS~\cite{casanova2021sc} & \cmark &  &  &  &  & \cmark &  &  &  & Transformer + Flow & HiFi-GAN & MelS & 2021.06 \\
    Meta-StyleSpeech~\cite{min2021meta} & \cmark &  &  &  &  & \cmark &  &  &  & Transformer & MelGAN~\cite{kumar2019melgan} & MelS & 2021.06 \\
    DelightfulTTS~\cite{liu2021delightfultts} &  & \cmark &  & \cmark & \cmark &  &  &  &  & Transformer + CNN & HiFiNet~\cite{liu2021delightfultts} & MelS & 2021.11 \\
    YourTTS~\cite{casanova2022yourtts} & \cmark &  &  &  &  & \cmark &  &  &  & Transformer + Flow & HiFi-GAN & LinS & 2021.12 \\
    StyleTTS~\cite{li2025styletts} & \cmark &  &  &  &  & \cmark &  &  &  & CNN + RNN & HiFi-GAN & MelS & 2022.05 \\
    GenerSpeech~\cite{huang2022generspeech} & \cmark &  &  &  &  & \cmark &  &  &  & Transformer + Flow & HiFi-GAN & MelS & 2022.05 \\
    Cauliflow~\cite{abbas2022expressive} &  &  &  & \cmark & \cmark &  &  &  &  & BERT + Flow & UP WaveNet~\cite{jiao2021upwavenet} & MelS & 2022.06 \\
    CLONE~\cite{liu2022controllable} &  & \cmark &  & \cmark & \cmark &  &  &  &  & Transformer + CNN & WaveNet~\cite{van2016wavenet} & MelS + LinS & 2022.07 \\
    PromptTTS~\cite{guo2023prompttts} &  & \cmark & \cmark & \cmark & \cmark & \cmark & \cmark &  & \cmark & BERT + Transformer & HiFi-GAN & MelS & 2022.11 \\
    Grad-StyleSpeech~\cite{kang2023grad} & \cmark &  &  &  &  & \cmark &  &  &  & Score-based Diffusion & HiFi-GAN & MelS & 2022.11 \\
    NaturalSpeech 2~\cite{shen2024naturalspeech2} & \cmark &  &  &  &  & \cmark &  &  &  & Diffusion & RVQ-based~\cite{shen2024naturalspeech2} & Latent Feature & 2023.04 \\
    PromptStyle~\cite{liu2023promptstyle} & \cmark & \cmark &  &  & \cmark & \cmark & \cmark &  & \cmark & VITS + Flow & HiFi-GAN & MelS & 2023.05 \\
    StyleTTS 2~\cite{li2023styletts2} & \cmark &  &  &  & \cmark & \cmark & \cmark &  &  & Flow-based Diffusion + GAN & HifiGAN / iSTFTNet~\cite{kaneko2022istftnet} & MelS & 2023.06 \\
    VoiceBox~\cite{le2024voicebox} & \cmark &  &  &  &  & \cmark &  &  &  & Transformer + Flow & HiFi-GAN & MelS & 2023.06 \\
    MegaTTS 2~\cite{jiang2024mega} & \cmark &  &  &  & \cmark & \cmark & \cmark &  &  & Decoder-only Transformer + GAN & HiFi-GAN & MelS & 2023.07 \\
    PromptTTS 2~\cite{leng2023prompttts2} &  & \cmark & \cmark & \cmark & \cmark & \cmark &  &  & \cmark & Diffusion & RVQ-based~\cite{leng2023prompttts2} & Latent Feature & 2023.09 \\
    VoiceLDM~\cite{lee2024voiceldm} &  & \cmark &  &  & \cmark & \cmark & \cmark & \cmark & \cmark & Diffusion & HiFi-GAN & MelS & 2023.09 \\
    DurIAN-E~\cite{gu2023durian} &  & \cmark &  & \cmark & \cmark &  &  &  &  & CNN + RNN & HiFi-GAN & MelS & 2023.09 \\
    PromptTTS++~\cite{shimizu2024prompttts++} &  & \cmark &  & \cmark & \cmark & \cmark & \cmark &  & \cmark & Transformer + Diffusion & BigVGAN~\cite{lee2023bigvgan} & MelS & 2023.09 \\
    SpeechFlow~\cite{liu2023generative} & \cmark &  &  &  &  & \cmark &  &  &  & Transformer + Flow & HiFi-GAN & MelS & 2023.10 \\
    P-Flow~\cite{kim2024p} & \cmark &  &  &  &  & \cmark &  &  &  & Transformer + Flow & HiFi-GAN & MelS & 2023.10 \\
    E3 TTS~\cite{gao2023e3} & \cmark &  &  &  &  & \cmark &  &  &  & Diffusion & \color{gray}{Not required} & Waveform & 2023.11 \\
    HierSpeech++~\cite{lee2023hierspeech++} & \cmark &  &  &  &  & \cmark &  &  &  & Transformer + VAE + Flow & BigVGAN & MelS & 2023.11 \\
    Audiobox~\cite{vyas2023audiobox} & \cmark & \cmark &  & \cmark & \cmark & \cmark &  & \cmark & \cmark & Transformer + Flow &  HiFi-GAN & MelS & 2023.12 \\
    FlashSpeech~\cite{ye2024flashspeech} & \cmark &  &  &  &  & \cmark &  &  &  & Latent Consistency Model & EnCodec & Token & 2024.04 \\
    NaturalSpeech 3~\cite{ju2024naturalspeech3} & \cmark &  &  & \cmark & \cmark & \cmark &  &  &  & Transformer + Diffusion & FACodec~\cite{ju2024naturalspeech3} & Token & 2024.04 \\
    InstructTTS~\cite{yang2024instructtts} &  & \cmark &  & \cmark & \cmark & \cmark & \cmark &  & \cmark & Transformer + Diffusion & HiFi-GAN & Token & 2024.05 \\
    ControlSpeech~\cite{ji2024controlspeech} & \cmark & \cmark & \cmark & \cmark & \cmark & \cmark & \cmark & & \cmark & Transformer + Diffusion & FACodec & Token & 2024.06 \\
    AST-LDM~\cite{kim2024speak} &  &  &  &  &  & \cmark &  & \cmark & \cmark & Diffusion + VAE & HiFi-GAN & MelS & 2024.06 \\
    SimpleSpeech~\cite{yang2024simplespeech} & \cmark &  &  &  &  & \cmark &  &  &  & Transformer + Diffusion & SQ Codec~\cite{yang2024simplespeech} & Token & 2024.06 \\
    DiTTo-TTS~\cite{lee2025dittotts} & \cmark &  &  & \cmark &  & \cmark &  &  &  & DiT + VAE & BigVGAN & MelS & 2024.06 \\
    E2 TTS~\cite{eskimez2024e2} & \cmark &  &  &  &  & \cmark &  &  &  & Transformer + Flow & BigVGAN & MelS & 2024.06 \\
    MobileSpeech~\cite{ji2024mobilespeech} & \cmark &  &  &  &  & \cmark &  &  &  & Transformer & Vocos~\cite{siuzdak2024vocos} & Token & 2024.06 \\
    DEX-TTS~\cite{park2024dex} & \cmark &  &  &  &  & \cmark &  &  &  & Diffusion & HiFi-GAN & MelS & 2024.06 \\
    ArtSpeech~\cite{wang2024artspeech} & \cmark &  &  &  &  & \cmark &  &  &  & RNN + CNN & HiFI-GAN & MelS + Energy + F0 & 2024.07 \\
    CCSP~\cite{xiao2024contrastive} & \cmark &  &  &  &  & \cmark &  &  &  & Diffusion & RVQ-based~\cite{xiao2024contrastive} & Token & 2024.07 \\
    SimpleSpeech 2~\cite{yang2024simplespeech2} & \cmark &  &  & \cmark &  & \cmark &  &  &  & Flow-based DiT & SQ Codec & Token & 2024.08 \\
    E1 TTS~\cite{liu2024e1} & \cmark &  &  &  &  & \cmark &  &  &  & DiT + Flow & BigVGAN & Token + MelS & 2024.09 \\
    StyleTTS-ZS~\cite{li2024stylettszs} & \cmark &  &  &  &  & \cmark &  &  &  & Flow-based Diffusion + GAN & Mel-based Decoder~\cite{li2024stylettszs} & MelS & 2024.09 \\
    NansyTTS~\cite{yamamoto2024description} & \cmark & \cmark &  & \cmark & \cmark & \cmark &  &  & \cmark & Transformer & NANSY++~\cite{yamamoto2024description} & MelS & 2024.09 \\
    NanoVoice~\cite{park2024nanovoice} & \cmark &  &  &  &  & \cmark &  &  &  & Diffusion & BigVGAN & MelS & 2024.09 \\
    MS$^{2}$KU-VTTS~\cite{he2024multi} &  &  &  &  &  &  &  & \cmark & \cmark & Transformer & BigVGAN & MelS & 2024.10 \\
    MaskGCT~\cite{wang2025maskgct} & \cmark &  &  & \cmark &  & \cmark &  &  &  & Transformer + Flow & Vocos & Token & 2024.10 \\
    EmoSphere++~\cite{cho2024emosphere++} & \cmark &  &  &  & \cmark & \cmark & \cmark &  &  & Transformer + Flow & BigVGAN & MelS & 2024.11 \\
    EmoDubber~\cite{cong2024emodubber} & \cmark &  &  &  & \cmark & \cmark & \cmark &  &  & Transformer + Flow & Flow-based~\cite{cong2024emodubber} & MelS & 2024.12 \\
    HED~\cite{inoue2024HED} & \cmark &  &  &  &  &  & \cmark &  &  & Flow-based Diffusion & Vocos & MelS & 2024.12 \\
    DiffStyleTTS~\cite{liu2025diffstyletts} &  & \cmark & \cmark & \cmark & \cmark & \cmark &  &  &  & Transformer + Diffusion & HiFi-GAN & MelS & 2025.01 \\
    DrawSpeech~\cite{chen2025drawspeech} &  &  & \cmark &  & \cmark &  &  &  &  & Diffusion & HiFi-GAN & MelS & 2025.01 \\
    ProEmo~\cite{zhang2025proemo} &  & \cmark & \cmark &  &  &  & \cmark &  & \cmark & Transformer & HiFi-GAN & MelS & 2025.01 \\
    \midrule
    \multicolumn{1}{c}{\multirow{2}{*}[-3pt]{Method (Autoregressive)}} & \multicolumn{1}{c}{\multirow{2}{*}[-3pt]{\makecell{Zero-shot}}} & \multicolumn{8}{c}{Controlability} & \multicolumn{2}{c}{Model Architectures} & \multicolumn{1}{c}{\multirow{2}{*}[-3pt]{\makecell{Feature}}} & \multicolumn{1}{c}{\multirow{2}{*}[-3pt]{\makecell{Release}}} \\ 
    \cmidrule(rl){3-10} \cmidrule(rl){11-12}
    & & Pit. & Ene. & Spe. & Pro. & Tim. & Emo. & Env. & Des. & Acoustic Model & Vocoder & & \\
    \midrule 
    Prosody-Tacotron~\cite{skerry2018towards} &  & \cmark &  &  & \cmark &  &  &  &  & RNN & WaveNet & MelS & 2018.03 \\
    GST-Tacotron~\cite{stanton2018predicting} &  & \cmark &  &  & \cmark &  &  &  &  & CNN + RNN & Griffin-Lim & LinS & 2018.03 \\
    GMVAE-Tacotron~\cite{hsu2018hierarchical} &  & \cmark &  & \cmark & \cmark &  &  & \cmark &  & VAE & WaveRNN & MelS & 2018.12 \\
    VAE-Tacotron~\cite{zhang2019learning} &  & \cmark &  & \cmark & \cmark &  &  &  &  & VAE + CNN + RNN & WaveNet & MelS & 2019.02 \\
    DurIAN~\cite{yu2020durian} &  & \cmark &  & \cmark & \cmark &  &  &  &  & CNN + RNN &  Multi-band WaveRNN~\cite{yu2020durian} & MelS & 2019.09 \\
    Flowtron~\cite{valle2020flowtron} &  & \cmark &  & \cmark & \cmark &  &  &  &  & CNN + RNN & WaveGlow & MelS & 2020.07 \\
    MsEmoTTS~\cite{lei2022msemotts} &  & \cmark &  &  & \cmark &  & \cmark &  &  & CNN + RNN & WaveRNN & MelS & 2022.01 \\
    VALL-E~\cite{wang2023neural} & \cmark &  &  &  &  & \cmark &  &  &  & Decoder-only Transformer & EnCodec & Token & 2023.01 \\
    SpearTTS~\cite{kharitonov2023speak} & \cmark &  &  &  &  & \cmark &  &  &  & Decoder-only Transformer & SoundStream~\cite{zeghidour2021soundstream} & Token & 2023.02 \\
    VALL-E X~\cite{zhang2023speak} & \cmark &  &  &  &  & \cmark &  &  &  & Decoder-only Transformer & EnCodec & Token & 2023.03 \\
    Make-A-Voice~\cite{huang2023makeavoice} & \cmark &  &  &  &  & \cmark &  &  &  & Encoder-decoder Transformer & Unit-based~\cite{huang2023makeavoice} & Token & 2023.05 \\
    TorToise~\cite{betker2023better} &  &  &  &  &  & \cmark &  &  &  & Decoder-only Transformer + Diffusion & UnivNet~\cite{JangLYKK21univnet} & MelS & 2023.05 \\
    MegaTTS~\cite{jiang2023megavoic} & \cmark &  &  &  &  & \cmark &  &  &  & Decoder-only Transformer + GAN & HiFi-GAN & MelS & 2023.06 \\
    SC VALL-E~\cite{kim2023sc} & \cmark & \cmark &  & \cmark & \cmark & \cmark & \cmark &  &  & Decoder-only Transformer & EnCodec & Token & 2023.07 \\
    Salle~\cite{ji2024textrolspeech} &  & \cmark & \cmark & \cmark & \cmark & \cmark & \cmark &  & \cmark & Decoder-only Transformer & EnCodec & Token & 2023.08 \\
    UniAudio~\cite{yang2023uniaudio} & \cmark & \cmark &  & \cmark & \cmark & \cmark &  &  & \cmark & Decoder-only Transformer & EnCodec & Token & 2023.10 \\
    ELLA-V~\cite{song2024ella} & \cmark &  &  &  &  & \cmark &  &  &  & Decoder-only Transformer & EnCodec & Token & 2024.01 \\
    Base TTS~\cite{lajszczak2024base} & \cmark &  &  &  &  & \cmark &  &  &  & Decoder-only Transformer & HiFi-GAN + BigVGAN & Token & 2024.02 \\
    CLaM-TTS~\cite{kim2024clam} & \cmark &  &  &  &  & \cmark &  &  &  & Encoder-decoder Transformer & BigVGAN & Token + MelS & 2024.04 \\
    RALL-E~\cite{xin2024rall} & \cmark &  &  &  &  & \cmark &  &  &  & Decoder-only Transformer & SoundStream & Token & 2024.05 \\
    ARDiT~\cite{liu2024autoregressive} & \cmark &  &  & \cmark &  & \cmark &  &  &  & Decoder-only DiT & BigVGAN & MelS & 2024.06 \\
    VALL-E R~\cite{han2024vall} & \cmark &  &  &  &  & \cmark &  &  &  & Decoder-only Transformer & Vocos & Token & 2024.06 \\
    VALL-E 2~\cite{chen2024vall} & \cmark &  &  &  &  & \cmark &  &  &  & Decoder-only Transformer & Vocos & Token & 2024.06 \\
    Seed-TTS~\cite{anastassiou2024seed} & \cmark &  &  &  &  & \cmark & \cmark &  &  & Decoder-only Transformer + DiT & \color{gray}{Unknown} & Latent Feature & 2024.06 \\
    VoiceCraft~\cite{peng2024voicecraft} & \cmark &  &  &  &  & \cmark &  &  &  & Decoder-only Transformer & HiFi-GAN & Token & 2024.06 \\
    XTTS~\cite{casanova2024xtts} & \cmark &  &  &  &  & \cmark &  &  &  & Decoder-only Transformer & HiFi-GAN-based~\cite{casanova2024xtts} & Token + MelS & 2024.06 \\
    CosyVoice~\cite{du2024cosyvoice} & \cmark & \cmark &  & \cmark & \cmark & \cmark & \cmark &  & \cmark & Decoder-only Transformer + Flow & HiFi-GAN & Token & 2024.07 \\
    MELLE~\cite{meng2024autoregressive} & \cmark &  &  &  &  & \cmark &  &  &  & Decoder-only Transformer & HiFi-GAN & MelS & 2024.07 \\
    VoxInstruct~\cite{zhou2024voxinstruct} & \cmark & \cmark & \cmark & \cmark & \cmark & \cmark & \cmark &  & \cmark & Decoder-only Transformer & Vocos & Token & 2024.08 \\
    Emo-DPO~\cite{gao2024emo} &  &  &  &  &  &  & \cmark &  & & Decoder-only Transformer & HiFi-GAN & Token + MelS & 2024.09 \\
    FireRedTTS~\cite{guo2024fireredtts} & \cmark &  &  &  & \cmark & \cmark &  &  &  & Decoder-only Transformer + Flow & BigVGAN & Token + MelS & 2024.09 \\
    CoFi-Speech~\cite{guo2024speaking} & \cmark &  &  &  &  & \cmark &  &  &  & Decoder-only Transformer & BigVGAN & Token + MelS & 2024.09 \\
    Takin~\cite{chen2024takin} & \cmark & \cmark &  & \cmark & \cmark & \cmark & \cmark &  & \cmark & Decoder-only Transformer + Flow & HiFi-GAN & Token + MelS & 2024.09 \\
    HALL-E~\cite{nishimura2024hall} & \cmark &  &  &  &  & \cmark &  &  &  & Decoder-only Transformer & EnCodec & Token & 2024.10 \\
    FishSpeech~\cite{liao2024fishspeech} & \cmark &  &  &  &  & \cmark &  &  &  & Decoder-only Transformer & Firefly-GAN~\cite{liao2024fishspeech} & Token & 2024.11 \\
    SLAM-Omni~\cite{chen2024slam-omni} & \cmark &  &  &  & \cmark & \cmark &  &  &  & Decoder-only Transformer & HiFi-GAN & Token + MelS & 2024.12 \\
    IST-LM~\cite{yang2024ist-lm} & \cmark &  &  &  & \cmark & \cmark &  &  &  & Decoder-only Transformer & HiFi-GAN & Token + MelS & 2024.12 \\
    KALL-E~\cite{zhu2024kall-e} & \cmark &  &  &  & \cmark & \cmark & \cmark &  &  & Decoder-only Transformer & WaveVAE~\cite{zhu2024kall-e} & Latent Feature & 2024.12 \\
    IDEA-TTS~\cite{lu2025idea-tts} & \cmark &  &  &  &  & \cmark &  & \cmark &  & Transformer & Flow-based~\cite{lu2025idea-tts} & LinS + MelS & 2024.12 \\
    FleSpeech~\cite{li2025flespeech} & \cmark & \cmark & \cmark & \cmark & \cmark & \cmark & \cmark &  & \cmark & Flow-based DiT & WaveGAN~\cite{donahue2018wavegan} & Latent Feature & 2025.01 \\
    Step-Audio~\cite{huang2025stepaudio} & \cmark &  &  &  & \cmark & \cmark & \cmark &  & \cmark & Decoder-only Transformer & Flow-based~\cite{huang2025stepaudio} & Token & 2025.02 \\
    Vevo~\cite{zhang2025vevo} & \cmark &  &  &  & \cmark & \cmark & \cmark &  &  & Decoder-only Transformer & BigVGAN & Token + MelS & 2025.02 \\
    Spark-TTS~\cite{wang2025sparktts} & \cmark & \cmark &  & \cmark & \cmark & \cmark &  &  & & Decoder-only Transformer & BiCodec~\cite{wang2025sparktts} & Token & 2025.03 \\
    EmoVoice~\cite{yang2025emovoice} & & &  &  &  &  & \cmark &  & \cmark & Decoder-only Transformer & HiFi-GAN & Token & 2025.04 \\
    \bottomrule
    \end{tabular}
    Abbreviations: Pit(ch), Ene(rgy), Spe(ed), Pro(sody), Tim(bre), Emo(tion), Env(ironment), Des(cription). MelS: Mel Spectrogram. LinS: Linear Spectrogram.
\end{threeparttable}}
\caption{A summary of existing controllable neural-based methods.}
\label{tab:methods_all}
\end{table*}

This section reviews controllable TTS from three perspectives: model architectures, feature representations, and control strategies, as shown in Fig.~\ref{fig:summary}.

\subsection{Model Architectures}\label{sec:archs}

The architectures of controllable TTS are primarily divided into two types, i.e., non-autoregressive (NAR) and autoregressive (AR) (See Table \ref{tab:methods_all}).

\subsubsection{Non-Autoregressive Approaches}

Non-autoregressive TTS models generate the entire output speech sequence $\mathbf{y} = (y_1, y_2, \dots, y_T)$ in parallel given the input $\mathbf{x} = (x_1, x_2, \dots, x_T)$:
\begin{equation}
    \underset{\theta}{\arg\max} = P(\mathbf{y} | \mathbf{x}; \theta),
\end{equation}
where $\theta$ denotes model parameters.
In this part, we investigate the transformer, variational autoencoder (VAE), diffusion, and flow-based methods.

\textbf{Transformer-based Methods.}
Transformers enable efficient context modeling and parallel TTS.
FastSpeech~\cite{ren2019fastspeech} introduced a non-autoregressive transformer that improves inference speed and stability via duration prediction.
FastSpeech 2~\cite{ren2020fastspeech} adds pitch and energy control, removing the need for distillation and boosting voice quality.
FastPitch~\cite{lancucki2021fastpitch} further incorporates direct pitch prediction into its architecture, enabling pitch manipulation.

\textbf{VAE-based Methods.}
VAEs enable structured, continuous latent representations by optimizing a variational lower bound.
VAEs have been leveraged to enhance prosody, emotion, and style control.
\citet{hsu2018hierarchical} proposed a hierarchical VAE to control noise and speaking rate.
\citet{zhang2019learning} introduced disentangled VAE representations for effective prosody and emotion transfer.
Parallel Tacotron~\cite{elias2021paralleltacotron} uses a VAE-based residual encoder with iterative spectrogram loss to improve speech naturalness.
CLONE~\cite{liu2022controllable} further improves prosody and energy modeling using conditional VAEs with normalizing flows~\cite{kobyzev2020normalizing} and adversarial training, achieving state-of-the-art quality and control.
These advances underscore VAEs' versatility in expressive and controllable speech synthesis.

\textbf{Diffusion-based Methods.}
Diffusion models~\cite{ho2020denoising} generate speech by reversing a noise injection process: noise is gradually added during the forward phase and removed in the reverse phase to synthesize high-quality audio.
NaturalSpeech 2~\cite{shen2024naturalspeech2} uses a latent diffusion-based codec with quantized latent vectors, while NaturalSpeech 3~\cite{ju2024naturalspeech3} decomposes speech into independent attribute subspaces with factorized diffusion-based codecs.
DEX-TTS~\cite{park2024dex} improves diffusion transformer (DiT)-based networks via overlapping patches and frequency-aware embeddings.
E3 TTS~\cite{gao2023e3} eliminates intermediate features by directly modeling waveforms through diffusion.
Text-to-audio models such as AudioLDM~\cite{liu2023audioldm} and Make-An-Audio~\cite{huang2023makeanaudio} can also generate speech using latent diffusion models.

\textbf{Flow-based Methods.}
Flow models use invertible flows~\cite{rezende2015variational,lipman2023flowmatching} to map speech features to simple distributions, typically Gaussians~\cite{prenger2019waveglow}, enabling direct, high-fidelity generation via inversion.
Recent models adopt flow-matching~\cite{lipman2023flowmatching} for efficient, non-autoregressive synthesis: Audiobox~\cite{vyas2023audiobox}, P-Flow~\cite{kim2024p}, and VoiceBox~\cite{le2024voicebox} consider TTS as speech infilling tasks, predicting masked mel-spectrograms.
FlashSpeech~\cite{ye2024flashspeech} trains a latent consistency model using adversarial training, achieving one- or two-step synthesis.
Inspired by audio infilling, E2 TTS~\cite{eskimez2024e2} uses filler-augmented text sequences to generate mel-spectrograms with human-level quality. F5-TTS~\cite{chen2024f5} builds on this with ConvNeXt v2~\cite{woo2023convnext} to enhance text-speech alignment by directly learning flows conditioned on text and reference speech.
E1 TTS~\cite{liu2024e1} distills rectified flow-based diffusion models~\cite{liu2023flow} into one-step generators via distribution matching, reducing sampling cost.

\subsubsection{Autoregressive Approaches}

Autoregressive TTS models predict the speech sequence $\mathbf{y} = (y_1, \dots, y_T)$ given input $\mathbf{x}$ as:
\begin{equation}
\underset{\theta}{\arg\max} = \prod_{t=1}^{T} P(y_t | y_{<t}, \mathbf{x}; \theta),
\end{equation}
where each frame $y_t$ depends on all previous outputs $y_{<t}$ and the transcript $\mathbf{x}$.
While this enables effective modeling of implicit duration and long-range context, autoregressive TTS models suffer from slower inference, making them more suitable for applications where flexibility is prioritized over speed.
This part investigates recurrent neural networks (RNN) and LLM-based methods.

\textbf{RNN-based Methods.}
RNNs enable natural-sounding speech synthesis with adjustable prosody, pitch, and emotion.
Prosody-Tacotron~\cite{skerry2018towards} extends Tacotron~\cite{wang2017tacotron} by introducing explicit prosodic controls.
\citet{wang2018style} proposed Global Style Tokens (GST), enabling unsupervised style transfer.
Emotion-controllable models such as \citet{li2021controllable} introduced emotion embeddings and style alignment to modulate emotional intensity.
MsEmoTTS~\cite{lei2022msemotts} refined this with a hierarchical structure capturing emotion at global, utterance, and local levels, enabling more nuanced synthesis.
These developments bring synthetic speech significantly closer to human expressiveness.
\begin{figure}[t]
    \centering
    \includegraphics[width=\linewidth]{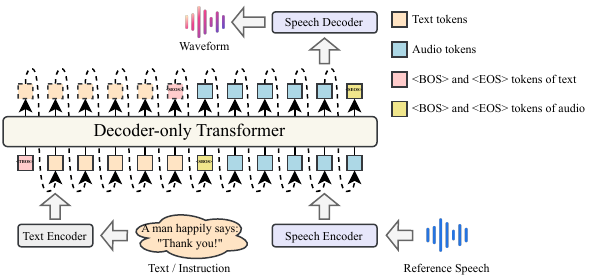}
    \caption{The typical architecture of LLM-based TTS.}
    \label{fig:llm-based}
\end{figure}

\textbf{LLM-based Methods.}
LLM-based TTS is inspired by the success of in-context learning in LLMs.
As illustrated in Fig.~\ref{fig:llm-based}, these approaches typically input target text or instructions with an optional reference speech, using autoregressive decoder-only transformers to generate speech tokens or features, which are then decoded into waveforms.
VALL-E~\cite{wang2023neural} pioneered LLM-based zero-shot TTS by framing it as a conditional language modeling task.
It uses EnCodec~\cite{fossez2023high} to discretize waveforms into tokens and adopts a two-stage pipeline: an autoregressive model generates coarse audio tokens, followed by a non-autoregressive model for iterative refinement.
This hierarchical modeling of semantic and acoustic tokens has laid the groundwork for many subsequent methods, such as VALL-E X~\cite{zhang2023speak}, ELLA-V~\cite{song2024ella}, RALL-E~\cite{xin2024rall}, VALL-E R~\cite{han2024vall}, MELLE~\cite{meng2024autoregressive}, and HALL-E~\cite{nishimura2024hall}.
Beyond the VALL-E series, recent work has further improved text-speech alignment, quality, and robustness. SpearTTS~\cite{kharitonov2023speak} and Make-a-Voice~\cite{huang2023makeavoice} leverage semantic tokens to better bridge text and acoustic features. FireRedTTS~\cite{guo2024fireredtts} refines the tokenizer architecture for improved reconstruction quality.
CoFi-Speech~\cite{guo2024speaking} adopts a coarse-to-fine, multi-scale generation strategy to produce natural, intelligible speech.

\subsubsection{Research Trend}

\begin{figure}[t]
    \centering
    \includegraphics[width=\linewidth]{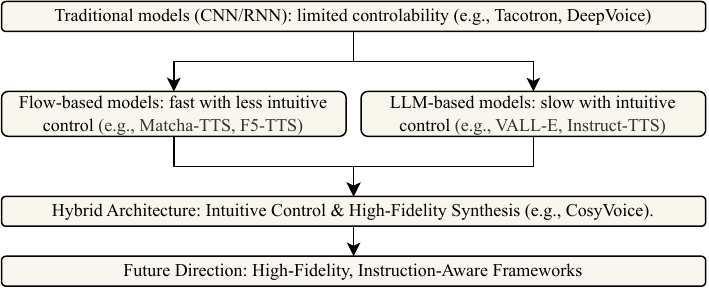}
    \caption{The evolution of TTS model architectures}
    \label{fig:trend_arch}
\end{figure}

\textbf{Traditional CNN/RNN TTS models} face inherent constraints. RNNs (e.g., Tacotron) are slow due to autoregressive inference and struggle with long-term dependencies. CNNs (e.g., Deep Voice) lack global prosody modeling. Both require explicit feature engineering for attributes like emotion and often trade synthesis quality for efficiency (e.g., WaveNet’s high fidelity but high latency).
\textbf{Flow-based models} (e.g., Matcha-TTS, F5-TTS) enable non-autoregressive, parallel synthesis with probabilistic control over acoustic features, improving speed and flexibility but increasing training complexity and dataset requirements. LLM-based models (e.g., VALL-E, InstructTTS) offer natural language-driven control and zero-shot voice cloning, supporting context-aware synthesis, but suffer from high computational cost and potential acoustic artifacts from discrete tokenization.
\textbf{Hybrid architectures} (e.g., CosyVoice) integrate LLM-guided semantic conditioning into flow-based generators, combining high-fidelity synthesis with intuitive, instruction-based control. Users can specify attributes like emotion or speaking style in natural language without compromising audio quality.
Future controllable TTS should balance efficiency, fidelity, and expressiveness, generalizing across voices, styles, and languages. Bridging instruction-based control and acoustic precision remains a key challenge, motivating advances in modular architectures, instruction grounding, and speech-text-instruction alignment.
Fig.~\ref{fig:trend_arch} summarizes the evolution and future direction of TTS model architectures.

\subsection{Control Strategies} \label{sec:strategies}

As illustrated in Fig.~\ref{fig:control_strategies}, control strategies in TTS can be broadly categorized into four types: style tagging, reference speech prompt, natural language descriptions, and instruction-guided control.

\begin{figure*}[t]
\centering
\resizebox{1.0\textwidth}{!}{
\begin{tikzpicture}[scale=1.0]
\genealogytree[
level 2/.style={level size=11.5cm,node box={colback=lightgray}},
level 1/.style={level size=2.3cm,node box={colback=starlight},further distance=1mm},
level 0/.style={level size=2.5cm,node box={colback=lightblue,rotate=90}},
timeflow=left,
processing=tcolorbox,
node size from=5mm to 4cm,
box={size=small,halign=center,valign=center,fontupper=\footnotesize\textrm{}},
edges={foreground=black!25,background=black!5},
class1/.style={box={colback=lightgray}},
class2/.style={box={colback=starlight}},
class3/.style={box={colback=lightblue}},
level distance=5mm,
]
{
    parent{
        g[class3]{\scriptsize Control Strategies}
            parent{
                g[class2]{\scriptsize Style Tagging}
                p[class1,id=styletagging]{\tiny StyleTagging-TTS~\cite{kim2021expressive}, Emo-DPO~\cite{gao2024emo}, Spark-TTS~\cite{wang2025sparktts}, DiffStyleTTS~\cite{liu2025diffstyletts}, DrawSpeech~\cite{chen2025drawspeech}, Cauliflow~\cite{abbas2022expressive}, DurIAN-E~\cite{gu2023durian}, DiTTo-TTS~\cite{lee2025dittotts}}
                }
            parent{
                g[class2]{\scriptsize Reference Speech Prompt}
                p[class1,id=speechprompt]{\tiny MetaStyleSpeech~\cite{min2021meta}, StyleTTS~\cite{li2025styletts}, GenerSpeech~\cite{huang2022generspeech}, SC VALL-E~\cite{kim2023sc}, ArtSpeech~\cite{wang2024artspeech}, CCSP~\cite{xiao2024contrastive}, DEX-TTS~\cite{park2024dex}, StyleTTS-ZS~\cite{li2024stylettszs}, NaturalSpeech 3~\cite{ju2024naturalspeech3}, MegaTTS 2~\cite{jiang2024mega}, ControlSpeech~\cite{ji2024controlspeech}, VoiceCraft~\cite{peng2024voicecraft}, Vevo~\cite{zhang2025vevo}, Spark-TTS~\cite{wang2025sparktts}}
                }
            parent{
                g[class2]{\scriptsize Natural Language Descriptions}
                p[class1]{\tiny PromptTTS~\cite{guo2023prompttts}, InstructTTS~\cite{yang2024instructtts}, PromptStyle~\cite{liu2023promptstyle}, Salle~\cite{ji2024textrolspeech}, NansyTTS~\cite{yamamoto2024description}, PromptTTS++~\cite{shimizu2024prompttts++}, PromptSpeaker~\cite{zhang2023promptspeaker}, PromptTTS 2~\cite{leng2023prompttts2}, ControlSpeech~\cite{ji2024controlspeech}, Audiobox~\cite{vyas2023audiobox}, Takin~\cite{chen2024takin}, VoiceLDM~\cite{lee2024voiceldm}, AST-LDM~\cite{kim2024speak}, FleSpeech~\cite{li2025flespeech}, MS$^{2}$KU-VTTS~\cite{he2024multi}, Parler-TTS~\cite{lyth2024natural}, ProEmo~\cite{zhang2025proemo}, EmoVoice~\cite{yang2025emovoice}}
                }
            parent{
                g[class2]{\scriptsize Instruction-Guided Control}
                p[class1]{\tiny VoxInstruct~\cite{zhou2024voxinstruct}, AudioGPT~\cite{huang2024audiogpt}, SpeechGPT~\cite{zhang2023speechgpt}, FunAudioLLM~\cite{an2024funaudiollm}, CosyVoice~\cite{du2024cosyvoice}, VoiceCraft~\cite{peng2024voicecraft},
                InstructSpeech~\cite{huang2024instructspeech}, Step-Audio~\cite{huang2025stepaudio}}
                }
        }
}
\end{tikzpicture}}
\caption{A taxonomy of controllable TTS from the perspective of control strategies.}
\label{fig:control_strategies}
\end{figure*}

\subsubsection{Style Tagging}

This paradigm enables the adjustment of key attributes such as pitch, energy, speech rate, and emotion, which can be controlled using either categorical labels or continuous values.
``Tagging'' refers to using a control signal to control a specific speech attribute.
1) Some approaches use \emph{discrete labels} to control speech attributes. For example, StyleTagging-TTS~\cite{kim2021expressive} denotes speech styles with short phrases or words (e.g., angry, happy), learning the relationship between linguistic and style embeddings.
Emo-DPO~\cite{gao2024emo} enables emotion control through Direct Preference Optimization (DPO)~\cite{rafailov2023dpo} with LLMs.
Spark-TTS~\cite{wang2025sparktts} provides coarse and fine-grained control, allowing pitch and speaking rate modifications via specially designed tokens and reference speech.
2) Other methods adjust \emph{continuous input signals}. DiffStyleTTS~\cite{liu2025diffstyletts} models prosody hierarchically, enabling control over pitch, energy, duration, and style via guiding scale factors.
DrawSpeech~\cite{chen2025drawspeech} lets users sketch prosody contours, which are refined and converted into detailed speech by a diffusion model, offering control over intonation.
3) Speech attributes can be controlled by \emph{modifying latent features}.
Cauliflow~\cite{abbas2022expressive} adjusts speech rate and pausing through a flow-based model conditioned on user-defined parameters.
DiTTo-TTS~\cite{lee2025dittotts} uses a DiT to control speech rate by modifying latent length predictions.
These methods show great potential in controlling speech attributes by adjusting input signals or latent variables.
However, these methods are limited in expressive diversity, as they can only model a small set of pre-defined attributes.

\subsubsection{Reference Speech Prompt}

This paradigm aims to customize the synthesized voice using only a few seconds of reference speech.
Similar to LLM-based methods, it takes both text and reference speech as input to a conditional TTS model, which generates speech based on both semantic and acoustic features.
MetaStyleSpeech~\cite{min2021meta} employs adaptive normalization for style conditioning, enabling robust zero-shot performance. 
GenerSpeech~\cite{huang2022generspeech} introduces a multilevel style adapter for improved zero-shot style transfer to out-of-domain custom voices.
SC VALL-E~\cite{kim2023sc} integrates style tokens and scale factors for controlling emotion, speaking style, and other acoustic features in the generated speech. 
DEX-TTS~\cite{park2024dex} separates time-invariant and time-variant style components, allowing the extraction of diverse styles.
StyleTTS-ZS~\cite{li2024stylettszs} uses distilled time-varying style diffusion to capture varied speaker identities and prosodies.
MegaTTS 2~\cite{jiang2024mega} introduces an acoustic autoencoder to separate prosody and timbre in the latent space, enabling style transfer to any timbre.
ControlSpeech~\cite{ji2024controlspeech} employs bidirectional attention and parallel decoding to control timbre, style, and content in a zero-shot manner.

\subsubsection{Natural Language Descriptions}

Recent studies have explored controlling speech attributes using natural language descriptions, offering better user-friendliness.
PromptTTS~\cite{guo2023prompttts} uses manually annotated prompts to describe five key speech attributes. 
InstructTTS~\cite{yang2024instructtts} introduces a three-stage training procedure to extract semantics from natural language prompts.
NansyTTS~\cite{yamamoto2024description} enables cross-lingual control by pairing a TTS model with a description controller trained on a different language using shared timbre and style representations.
To address the limitations of textual prompts in capturing speaker characteristics, PromptTTS++~\cite{shimizu2024prompttts++} enhances prompt richness by using additional speaker description prompts. 
PromptTTS 2~\cite{leng2023prompttts2} introduces a variation network to model residual variability beyond the prompt. 
Further efforts extend controllability to the environmental context.
VoiceLDM~\cite{lee2024voiceldm} and AST-LDM~\cite{kim2024speak} extend AudioLDM~\cite{liu2023audioldm} by incorporating content prompts to enable environmental conditioning.
MS$^{2}$KU-VTTS~\cite{he2024multi} further enhances environmental perception by mixing environmental images into the prompt, enabling more immersive speech generation.

\subsubsection{Instruction-Guided Synthesis}

Description-based TTS methods separate inputs into content and description prompts, diverging from the unified instruction formats used in chatbots.
To address this, VoxInstruct~\cite{zhou2024voxinstruct} reframes TTS as a general instruction-to-speech task, where a single natural language prompt conveys both content and style descriptions.
CosyVoice~\cite{du2024cosyvoice} enhances this paradigm using supervised semantic tokens derived from ASR models.
It combines LLM-driven token generation with flow-matching synthesis, enabling precise control over speaker identity, emotion, pitch, speed, and paralinguistic cues through natural language instructions.
AudioGPT~\cite{huang2024audiogpt} is a multimodal LLM-based agent, incorporating multiple modules for speech understanding, synthesis, and style conversion.
StepAudio~\cite{huang2025stepaudio} introduces a speech-text model with an instruction-driven TTS module, enabling dynamic control over dialects, emotions, singing, rapping, and speaking styles.
These advancements push toward more intuitive, instruction-driven speech generation.

\subsubsection{Instruction-Guided Editing}

Some methods also support speech editing via user instructions.
VoiceCraft~\cite{peng2024voicecraft} uses a decoder-only transformer with causal masking and delayed stacking for bidirectional, context-aware instruction-guided editing, such as insertion, deletion, and substitution, while maintaining high naturalness. 
InstructSpeech~\cite{huang2024instructspeech} trains a multi-task LLM on <instruction, input, output> triplets with task embeddings and hierarchical adapters, allowing content and acoustic attributes control.
It supports flexible, free-form speech editing and task adaptation by multi-step reasoning.

\subsubsection{Research Trend}

The evolution of TTS control has moved from basic attribute manipulation to sophisticated, instruction-guided synthesis, reflecting AI’s trend toward intuitive, fine-grained control.
Early methods like \textbf{style tagging} controlled predefined attributes (e.g., pitch, emotion) but offered limited expressive diversity. \textbf{Reference speech prompts} enabled zero-shot TTS and voice cloning, separating timbre from style for greater personalization. To improve user-friendliness, \textbf{natural language descriptions} (e.g., PromptTTS) allowed users to specify vocal characteristics through text.
The latest advance, \textbf{instruction-guided control}, leverages LLMs to interpret free-form instructions combining content and style. Systems like VoxInstruct and CosyVoice generate nuanced speech, including paralinguistic sounds, enabling highly precise, user-centric synthesis.
Overall, the progression from tags to natural instructions shows a clear trajectory toward more \textbf{expressive, personalized, and intuitive TTS}, driven by LLM integration.
Table~\ref{tab:appd_control_strategies} in the Appendix provides a summary of the strengths and weaknesses of each control strategy.

\subsection{Feature Representations} \label{sec:features}

The learning and choice of feature representations critically affect flexibility, naturalness, and controllability.
This subsection discusses speech attribute disentanglement and compares continuous and discrete representations, highlighting their trade-offs.

\textbf{Speech Attribute Disentanglement.}
Attribute disentanglement aims to isolate distinct speech factors, such as speaker identity, emotion, prosody, and content, into separate latent representations. The two main approaches are:
1) \emph{Adversarial training}~\cite{goodfellow2020gan} uses auxiliary classifiers to penalize the presence of unwanted attributes in a latent space.
An encoder learns to ``fool'' these classifiers, resulting in representations that are invariant to specific attributes like speaker~\cite{yang2021ganspeech,Hsu2019Disentangling,lee2021multi}, emotion~\cite{li2022cross}, and style~\cite{li2023styletts2}.
2) \emph{Information bottleneck} uses small-capacity or independent encoder branches to isolate attributes.
Each branch encodes one factor (e.g., content, prosody)~\cite{ju2024naturalspeech3}, often with adversarial or reconstruction losses to discourage leakage of other information.
These methods are often combined.
Regularization via KL divergence~\cite{lu2023speechtriplenet} or  quantization~\cite{zhang2025vevo} also plays a key role in enforcing disentanglement.

\textbf{Continuous Representations.}
Continuous representations model speech in a continuous feature space, preserving acoustic details.
The key advantages are:
1) Fine-grained detail retention for natural and expressive synthesis;
2) Inherent encoding of prosody, pitch, and emotion, aiding controllable and emotional TTS;
3) Enable smooth audio reconstruction without quantization artifacts.
GAN-based~\cite{kong2020hifigan,yamamoto2020parallelwavegan}, VAE-based~\cite{lee2023hierspeech++,lee2025dittotts}, flow-based~\cite{kim2024pflow,casanova2022yourtts}, and diffusion-based methods~\cite{kong2021diffwave,huang2022fastdiff} often utilize continuous feature representations.
However, they are computationally intensive and demand large models and datasets for high-fidelity audio generation.

\textbf{Discrete Tokens.}
Discrete token-based TTS uses quantized units or phoneme-like tokens as acoustic features, which are often derived from quantization~\cite{zeghidour2021soundstream} or learned embeddings~\cite{hsu2021hubert}.
The advantages of discrete tokens are:
1) Discrete tokens can encode phonemes or sub-word units, making them concise and computationally efficient.
2) Discrete tokens often allow TTS systems to require fewer samples to learn and generalize, compared with continuous features.
3) Using discrete tokens simplifies cross-modal TTS applications like description-based TTS, as they are suitable for LLM training.
LLM-based methods~\cite{zhou2024voxinstruct,yang2024instructtts,du2024cosyvoice} often adopt discrete tokens as acoustic features.
However, discrete feature learning may cause information loss or lack the nuanced details in continuous features.

Table~\ref{tab:methods_all} summarizes the features used in existing methods.
We also compare speech quantization and tokenization in Appendix~\ref{sec:appd_feature} and summarize open-source methods in Table~\ref{tab:speech_feature} in the Appendix.

\section{Datasets and Evaluation Methods} \label{sec:datasets_eval}

\subsection{Datasets}
Fully controllable TTS systems require large, diverse, and finely annotated datasets to generate expressive, attribute-controllable speech.
There are mainly three types of datasets for controllable TTS:

\textbf{Tag-based Datasets.}
Tag-based datasets contain speech recordings annotated with predefined discrete attribute labels that describe various aspects of the speech audio~\cite{zhou2022emotional,busso2008iemocap,ringeval2013RECOLA,bagher2018cmu-mosei}. Common attributes include pitch, energy, speaking rate, age, gender, emotion, emphasis, accent, and topic.
By leveraging attribute labels, models can dynamically adjust specific speech characteristics, enabling more expressive synthesis.

\textbf{Description-based Datasets.}
Description-based datasets pair speech samples with rich, free-form textual descriptions that capture nuanced attributes such as intonation, prosody, speaking style, and emotional tone~\cite{guo2023prompttts,ji2024textrolspeech,jin2024speechcraft,lyth2024natural}. Unlike tag-based datasets with predefined categorical labels, these datasets allow models to interpret and generate speech from natural language prompts, enabling context-aware and highly expressive synthesis.

\textbf{Dialogue Datasets.}
Dialogue datasets~\cite{byrne2019taskmaster,lee2023dailytalk,yang2022magicdata} contain multi-turn conversational speech involving two or more speakers, emphasizing natural interaction features such as turn-taking, contextual dependencies, speaker intent, pauses, and prosodic variation. These datasets are essential for generating dynamic and contextually appropriate speech for interactive systems.

By leveraging these datasets, researchers can develop more expressive, context-aware, and highly controllable TTS models.
Table~\ref{tab:datasets} in the Appendix summarizes publicly available datasets.

\subsection{Evaluation Methods}

\subsubsection{Objective and Subjective Metrics}

\textbf{Objective Metrics.}
Objective metrics enable automated and reproducible evaluation. \emph{Mel Cepstral Distortion} (MCD)~\cite{kominek2008synthesizer} quantifies spectral distance between synthesized and reference speech.
MCD below 4 suggests good synthesis, while values above 6 imply distortion. 
\emph{Fréchet DeepSpeech Distance} (FDSD)~\cite{Bińkowski2020High} evaluates speech quality by measuring the distributional distance between synthesized and real speech in the embedding space of a pretrained speech recognition model, such as DeepSpeech~\cite{hannun2014deepspeech}. It compares the mean and covariance of extracted features; thus, a lower FDSD indicates higher perceptual similarity.
\emph{Word Error Rate} (WER)~\cite{enwiki2024wer} quantifies speech intelligibility by comparing recognized and reference transcripts.
\emph{Cosine Similarity} assesses speaker similarity by comparing speaker embeddings (extracted using models like ECAPA-TDNN~\cite{Brecht20ECAPA-TDNN} or x-vectors~\cite{Snyder2018xvectors}) of synthesized and reference speech. Higher values indicate better voice cloning.
\emph{Perceptual Evaluation of Speech Quality} (PESQ)~\cite{Rix2001PESQ} evaluates the intelligibility and distortion of synthesized speech by modeling human auditory perception.

\textbf{Subjective Metrics.}
Subjective metrics assess the perceptual quality of synthesized speech based on human judgments, capturing aspects like expressiveness and style similarity.
\emph{Mean Opinion Score} (MOS) rates synthesized speech (e.g., naturalness) on a 1–5 scale. Though effective in capturing human perception, MOS is costly for large-scale use.
\emph{Comparison MOS} (CMOS)~\cite{loizou2011speech} assesses relative quality by asking participants compare paired samples. Both are averaged across listeners.
\emph{AB/ABX Tests} present listeners with two samples (AB) by different methods or two plus a reference (ABX) to judge preference or closeness to the reference.
They are very common in evaluating fine-grained or zero-shot TTS.
Appendix~\ref{sec:appd_metrics} details the metric computations, while Table~\ref{tab:eval_metrics} summarizes the most commonly used ones.

\subsubsection{Model-based Evaluation}

Model-based evaluation is also an emerging technique, e.g., automatic MOS evaluation~\cite{lian2025apg} and GPT-based evaluation~\cite{rong2025dopamine}.
To evaluate the controllability of existing TTS models, we designed a pipeline using Google Gemini to assess synthesized speech along three dimensions, i.e., instruction following, naturalness, and expressiveness, which are not well captured by traditional metrics. Details of this evaluation are provided in Appendix~\ref{sec:evaluation}.


\section{Challenges and Future Directions}
\label{sec:directions}


\subsection{Challenges}

\textbf{Fine-Grained Attribute Control.}
Emotion and other vocal traits are often intertwined and span multiple granularities, making fine-grained control especially difficult.
This requires high-resolution annotations and advanced models capable of capturing subtle attribute variations.
While description-based methods like VoxInstruct~\cite{zhou2024voxinstruct} allow control via attribute descriptions, precisely targeting a specific granularity or enabling multiscale, fine-grained control remains a big challenge.

\textbf{Feature Disentanglement.}
Fully controllable TTS requires effective feature disentanglement, yet extracting meaningful and independent speech attributes is challenging due to their interdependence and context sensitivity.
For instance, modifying pitch can also affect emotion and naturalness.
To address this, prior work~\cite{an2022disentangling,wang2023generalizable} leverages pre-trained models on tasks like emotion classification and adversarial training to guide feature separation.
However, designing disentanglement methods for more subtle prosodic attributes, such as sarcasm, remains an open challenge and merits further research.

\textbf{Scarcity of Datasets.}
Effective control requires training data that spans a wide range of styles, emotions, accents, and prosodic patterns.
Large-scale datasets like GigaSpeech~\cite{chen2021gigaspeech} and TextrolSpeech~\cite{ji2024textrolspeech} exist, but lack the content and scenario diversity, e.g., comedies, thrillers, and cartoons.
Fine-grained, attribute-specific annotations are another bottleneck. 
Manual labeling is expensive, laborious, requires expertise, and is often inconsistent, especially for subjective traits like emotion.
Most datasets offer only coarse labels (e.g., gender, age, emotions).
While datasets like SpeechCraft~\cite{jin2024speechcraft} and Parler-TTS~\cite{lyth2024natural} include textual descriptions, none provide annotations across varying conditions within the same speaker.

\subsection{Future Directions}


\textbf{Instruction-Guided Fine-Grained Speech Synthesis and Editing.}
Natural language-driven control of fine-grained speech attributes remains underexplored. Most existing methods support only a limited set of controllable attributes. While models like VoxInstruct~\cite{zhou2024voxinstruct} and CosyVoice~\cite{du2024cosyvoice} show promise in controlling emotion, timbre, and style, they often produce speech that deviates from user intent, requiring multiple synthesis attempts.
Similarly, speech editing methods~\cite{tae2022editts,tan2021editspeech} typically rely on conditional models with fixed inputs, offering limited flexibility for fine-grained, instruction-guided modifications. Thus, developing disentangled representations that support precise control through user instructions is promising.

\textbf{Expressive Multimodal Speech Synthesis.}
Synthesizing speech from multimodal inputs such as texts, images, and videos has broad industrial applications in storytelling, film, and gaming. While prior work~\cite{goto2020face2speech,lu2022visualtts,rong2025seeing} explores this direction, current methods struggle to effectively extract and utilize rich multimodal information. Generating expressive, engaging speech for complex visual content remains a promising area for future research.

\textbf{Zero-shot Long Speech and Conversational Synthesis with Emotion Consistency.}
Zero-shot TTS enables voice cloning and style transfer without fine-tuning~\cite{wang2025maskgct,chen2024f5,du2024cosyvoice}, yet struggles with generating long, content-emotion consistent speech due to limited reference input. Overcoming this challenge is key to advancing long speech synthesis.
Besides, conversational TTS, often cascaded and context-unaware, produces robotic and unexpressive speech. Recent advances leverage LLMs and discrete speech tokens~\cite{fang2025llamaomni,zhang2023speechgpt}, but context-aware, emotionally rich conversational TTS remains underexplored.

\textbf{Large-Scale Dataset Generation.}
Dataset construction is critical for both fine-grained control and editing tasks. Researchers can leverage pre-trained speech analysis models to annotate attributes like pitch, energy, emotion, gender, and age. From these annotations, tools like ChatGPT can generate diverse natural language descriptions of speech characteristics. For speech editing, pre-trained models can assist with tasks such as word substitution, pitch adjustment, and emotion conversion, while ChatGPT can provide varied and semantically rich editing instructions for training and evaluation. In addition, multi-agent systems can also be utilized to generate long-form and diverse speech content.


\section{Conclusion} \label{sec:conclusion}

This survey provides a comprehensive review of controllable TTS methods, covering model architectures, control strategies, and feature representations.
We also summarize commonly used datasets and evaluation metrics, discuss major challenges, and highlight promising future directions. 
To the best of our knowledge, this is the first comprehensive survey dedicated to controllable TTS.

\section{Limitations}

We acknowledge several limitations that may affect the completeness of our survey.
First, this survey does not explore the interactions between controllable attributes.
Most existing studies focus on modeling each factor, such as emotion, speaker identity, or prosody, as an independent variable.
However, understanding how these attributes influence one another during synthesis could lead to more effective and flexible control strategies.
Second, we do not address the efficiency of current controllable TTS systems.
It is important to note that approaches guided by descriptions or instructions often involve considerable computational cost, largely due to their reliance on large language model-based codecs and complex cross-modal architectures.
Third, we have not discussed the broader societal implications of controllable TTS, such as the risks associated with deepfake generation or adversarial attacks.
Finally, this survey does not cover related research areas, including speech enhancement, speech separation, speech pretraining, and speech-to-speech translation, which may offer valuable insights or complementary techniques.
Overcoming these limitations presents important opportunities for future research to deepen our understanding and improve the design of controllable TTS systems.

\section{Ethics Statements}

The literature search and review were conducted using sources such as Google Scholar, arXiv, DBLP, Scopus, and ChatGPT. All referenced papers included in this survey were thoroughly read, analyzed, and categorized by the authors.
Portions of the original content in this survey were paraphrased and refined with the assistance of ChatGPT. 

\section{Acknowledgement}

This work was supported by the National Natural Science Foundation of China (No. 62471420), GuangDong Basic and Applied Basic Research Foundation (2025A1515012296), and CCF-Tencent Rhino-Bird Open Research Fund.

\bibliography{custom}

\begin{thebibliography}{237}
\providecommand{\natexlab}[1]{#1}

\bibitem[{Abbas et~al.(2022)Abbas, Merritt, Moinet, Karlapati, Muszynska, Slangen, Gatti, and Drugman}]{abbas2022expressive}
Ammar Abbas, Thomas Merritt, Alexis Moinet, Sri Karlapati, Ewa Muszynska, Simon Slangen, Elia Gatti, and Thomas Drugman. 2022.
\newblock Expressive, variable, and controllable duration modelling in {TTS}.
\newblock \emph{arXiv preprint arXiv:2206.14165}.

\bibitem[{Allen et~al.(1987)Allen, Hunnicutt, Klatt, Armstrong, and Pisoni}]{allen1987mitalk}
Jonathan Allen, M~Sharon Hunnicutt, Dennis~H Klatt, Robert~C Armstrong, and David~B Pisoni. 1987.
\newblock \emph{From Text to Speech: The MITalk System}.
\newblock Cambridge University Press.

\bibitem[{Almeida and Xex{\'e}o(2019)}]{almeida2019word}
Felipe Almeida and Geraldo Xex{\'e}o. 2019.
\newblock Word embeddings: A survey.
\newblock \emph{arXiv preprint arXiv:1901.09069}.

\bibitem[{An et~al.(2024)An, Chen, Deng, Du, Gao, Gao, Gu, He, Hu, Hu et~al.}]{an2024funaudiollm}
Keyu An, Qian Chen, Chong Deng, Zhihao Du, Changfeng Gao, Zhifu Gao, Yue Gu, Ting He, Hangrui Hu, Kai Hu, and 1 others. 2024.
\newblock {FunAudioLLM}: Voice understanding and generation foundation models for natural interaction between humans and {LLMs}.
\newblock \emph{arXiv preprint arXiv:2407.04051}.

\bibitem[{An et~al.(2022)An, Soong, and Xie}]{an2022disentangling}
Xiaochun An, Frank~K Soong, and Lei Xie. 2022.
\newblock Disentangling style and speaker attributes for {TTS} style transfer.
\newblock \emph{IEEE/ACM Transactions on Audio, Speech, and Language Processing}, 30:646--658.

\bibitem[{Anastassiou et~al.(2024)Anastassiou, Chen, Chen, Chen, Chen, Chen, Cong, Deng, Ding, Gao et~al.}]{anastassiou2024seed}
Philip Anastassiou, Jiawei Chen, Jitong Chen, Yuanzhe Chen, Zhuo Chen, Ziyi Chen, Jian Cong, Lelai Deng, Chuang Ding, Lu~Gao, and 1 others. 2024.
\newblock {Seed-TTS}: A family of high-quality versatile speech generation models.
\newblock \emph{arXiv preprint arXiv:2406.02430}.

\bibitem[{Ardila et~al.(2020)Ardila, Branson, Davis, Kohler, Meyer, Henretty, Morais, Saunders, Tyers, and Weber}]{ardila2020commonvoice}
Rosana Ardila, Megan Branson, Kelly Davis, Michael Kohler, Josh Meyer, Michael Henretty, Reuben Morais, Lindsay Saunders, Francis Tyers, and Gregor Weber. 2020.
\newblock {Common Voice}: A massively-multilingual speech corpus.
\newblock In \emph{Proceedings of the Twelfth Language Resources and Evaluation Conference}, pages 4218--4222.

\bibitem[{Baevski et~al.(2022)Baevski, Hsu, Xu, Babu, Gu, and Auli}]{baevski2022data2vec}
Alexei Baevski, Wei-Ning Hsu, Qiantong Xu, Arun Babu, Jiatao Gu, and Michael Auli. 2022.
\newblock Data2vec: A general framework for self-supervised learning in speech, vision and language.
\newblock In \emph{International Conference on Machine Learning}, pages 1298--1312.

\bibitem[{Baevski et~al.(2020{\natexlab{a}})Baevski, Schneider, and Auli}]{baevski2019vqwav2vec}
Alexei Baevski, Steffen Schneider, and Michael Auli. 2020{\natexlab{a}}.
\newblock vq-wav2vec: Self-supervised learning of discrete speech representations.
\newblock In \emph{International Conference on Learning Representations}.

\bibitem[{Baevski et~al.(2020{\natexlab{b}})Baevski, Zhou, Mohamed, and Auli}]{baevski2020wav2vec}
Alexei Baevski, Yuhao Zhou, Abdelrahman Mohamed, and Michael Auli. 2020{\natexlab{b}}.
\newblock wav2vec 2.0: A framework for self-supervised learning of speech representations.
\newblock \emph{Advances in Neural Information Processing Systems}, 33:12449--12460.

\bibitem[{Bagher~Zadeh et~al.(2018)Bagher~Zadeh, Liang, Poria, Cambria, and Morency}]{bagher2018cmu-mosei}
AmirAli Bagher~Zadeh, Paul~Pu Liang, Soujanya Poria, Erik Cambria, and Louis-Philippe Morency. 2018.
\newblock Multimodal language analysis in the wild: {CMU}-{MOSEI} dataset and interpretable dynamic fusion graph.
\newblock In \emph{Proceedings of the 56th Annual Meeting of the Association for Computational Linguistics (Volume 1: Long Papers)}, pages 2236--2246.

\bibitem[{Barrault et~al.(2023)Barrault, Chung, Meglioli, Dale, Dong, Duppenthaler, Duquenne, Ellis, Elsahar, Haaheim et~al.}]{barrault2023seamless}
Lo{\"\i}c Barrault, Yu-An Chung, Mariano~Coria Meglioli, David Dale, Ning Dong, Mark Duppenthaler, Paul-Ambroise Duquenne, Brian Ellis, Hady Elsahar, Justin Haaheim, and 1 others. 2023.
\newblock Seamless: Multilingual expressive and streaming speech translation.
\newblock \emph{arXiv preprint arXiv:2312.05187}.

\bibitem[{Betker(2023)}]{betker2023better}
James Betker. 2023.
\newblock Better speech synthesis through scaling.
\newblock \emph{arXiv preprint arXiv:2305.07243}.

\bibitem[{Bińkowski et~al.(2020)Bińkowski, Donahue, Dieleman, Clark, Elsen, Casagrande, Cobo, and Simonyan}]{Bińkowski2020High}
Mikołaj Bińkowski, Jeff Donahue, Sander Dieleman, Aidan Clark, Erich Elsen, Norman Casagrande, Luis~C. Cobo, and Karen Simonyan. 2020.
\newblock High fidelity speech synthesis with adversarial networks.
\newblock In \emph{International Conference on Learning Representations}.

\bibitem[{Brown et~al.(2020)Brown, Mann, Ryder, Subbiah, Kaplan, Dhariwal, Neelakantan, Shyam, Sastry, Askell, Agarwal, Herbert-Voss, Krueger, Henighan, Child, Ramesh, Ziegler, Wu, Winter, Hesse, Chen, Sigler, Litwin, Gray, Chess, Clark, Berner, McCandlish, Radford, Sutskever, and Amodei}]{brown2020gpt3}
Tom Brown, Benjamin Mann, Nick Ryder, Melanie Subbiah, Jared~D Kaplan, Prafulla Dhariwal, Arvind Neelakantan, Pranav Shyam, Girish Sastry, Amanda Askell, Sandhini Agarwal, Ariel Herbert-Voss, Gretchen Krueger, Tom Henighan, Rewon Child, Aditya Ramesh, Daniel Ziegler, Jeffrey Wu, Clemens Winter, and 12 others. 2020.
\newblock Language models are few-shot learners.
\newblock In \emph{Advances in Neural Information Processing Systems}, volume~33, pages 1877--1901.

\bibitem[{Bulut et~al.(2002)Bulut, Narayanan, and Syrdal}]{bulut2002expressive}
Murtaza Bulut, Shrikanth~S Narayanan, and Ann~K Syrdal. 2002.
\newblock Expressive speech synthesis using a concatenative synthesizer.
\newblock In \emph{Proceedings of the Annual Conference of the International Speech Communication Association}, pages 1265--1268.

\bibitem[{Busso et~al.(2008)Busso, Bulut, Lee, Kazemzadeh, Mower, Kim, Chang, Lee, and Narayanan}]{busso2008iemocap}
Carlos Busso, Murtaza Bulut, Chi-Chun Lee, Abe Kazemzadeh, Emily Mower, Samuel Kim, Jeannette~N Chang, Sungbok Lee, and Shrikanth~S Narayanan. 2008.
\newblock {IEMOCAP}: Interactive emotional dyadic motion capture database.
\newblock \emph{Language Resources and Evaluation}, 42:335--359.

\bibitem[{Byrne et~al.(2019)Byrne, Krishnamoorthi, Sankar, Neelakantan, Duckworth, Yavuz, Goodrich, Dubey, Cedilnik, and Kim}]{byrne2019taskmaster}
Bill Byrne, Karthik Krishnamoorthi, Chinnadhurai Sankar, Arvind Neelakantan, Daniel Duckworth, Semih Yavuz, Ben Goodrich, Amit Dubey, Andy Cedilnik, and Kyu-Young Kim. 2019.
\newblock Taskmaster-1: Toward a realistic and diverse dialog dataset.
\newblock \emph{arXiv preprint arXiv:1909.05358}.

\bibitem[{Casanova et~al.(2024)Casanova, Davis, Gölge, Göknar, Gulea, Hart, Aljafari, Meyer, Morais, Olayemi, and Weber}]{casanova2024xtts}
Edresson Casanova, Kelly Davis, Eren Gölge, Görkem Göknar, Iulian Gulea, Logan Hart, Aya Aljafari, Joshua Meyer, Reuben Morais, Samuel Olayemi, and Julian Weber. 2024.
\newblock {XTTS}: a massively multilingual zero-shot text-to-speech model.
\newblock In \emph{Conference of the International Speech Communication Association}, pages 4978--4982.

\bibitem[{Casanova et~al.(2021)Casanova, Shulby, G{\"o}lge, M{\"u}ller, De~Oliveira, Junior, Soares, Aluisio, and Ponti}]{casanova2021sc}
Edresson Casanova, Christopher Shulby, Eren G{\"o}lge, Nicolas~Michael M{\"u}ller, Frederico~Santos De~Oliveira, Arnaldo~Candido Junior, Anderson da~Silva Soares, Sandra~Maria Aluisio, and Moacir~Antonelli Ponti. 2021.
\newblock {SC-GlowTTS}: An efficient zero-shot multi-speaker text-to-speech model.
\newblock \emph{arXiv preprint arXiv:2104.05557}.

\bibitem[{Casanova et~al.(2022)Casanova, Weber, Shulby, Junior, G{\"o}lge, and Ponti}]{casanova2022yourtts}
Edresson Casanova, Julian Weber, Christopher~D Shulby, Arnaldo~Candido Junior, Eren G{\"o}lge, and Moacir~A Ponti. 2022.
\newblock Your{TTS}: Towards zero-shot multi-speaker {TTS} and zero-shot voice conversion for everyone.
\newblock In \emph{International Conference on Machine Learning}, pages 2709--2720.

\bibitem[{Chen et~al.(2021)Chen, Chai, Wang, Du, Zhang, Weng, Su, Povey, Trmal, Zhang, Jin, Khudanpur, Watanabe, Zhao, Zou, Li, Yao, Wang, You, and Yan}]{chen2021gigaspeech}
Guoguo Chen, Shuzhou Chai, Guan-Bo Wang, Jiayu Du, Wei-Qiang Zhang, Chao Weng, Dan Su, Daniel Povey, Jan Trmal, Junbo Zhang, Mingjie Jin, Sanjeev Khudanpur, Shinji Watanabe, Shuaijiang Zhao, Wei Zou, Xiangang Li, Xuchen Yao, Yongqing Wang, Zhao You, and Zhiyong Yan. 2021.
\newblock {GigaSpeech}: An evolving, multi-domain {ASR} corpus with 10,000 hours of transcribed audio.
\newblock In \emph{Annual Conference of the International Speech Communication Association}, pages 3670--3674.

\bibitem[{Chen et~al.(2024{\natexlab{a}})Chen, Liu, Zhou, Liu, Tan, Li, Zhao, Qian, and Wei}]{chen2024vall}
Sanyuan Chen, Shujie Liu, Long Zhou, Yanqing Liu, Xu~Tan, Jinyu Li, Sheng Zhao, Yao Qian, and Furu Wei. 2024{\natexlab{a}}.
\newblock {VALL-E} 2: Neural codec language models are human parity zero-shot text to speech synthesizers.
\newblock \emph{arXiv preprint arXiv:2406.05370}.

\bibitem[{Chen et~al.(2024{\natexlab{b}})Chen, Feng, He, He, He, Hu, Lin, Lin, Pan, Tan et~al.}]{chen2024takin}
Sijing Chen, Yuan Feng, Laipeng He, Tianwei He, Wendi He, Yanni Hu, Bin Lin, Yiting Lin, Yu~Pan, Pengfei Tan, and 1 others. 2024{\natexlab{b}}.
\newblock Takin: A cohort of superior quality zero-shot speech generation models.
\newblock \emph{arXiv preprint arXiv:2409.12139}.

\bibitem[{Chen et~al.(2025)Chen, Yang, Li, and Wu}]{chen2025drawspeech}
Weidong Chen, Shan Yang, Guangzhi Li, and Xixin Wu. 2025.
\newblock {DrawSpeech}: Expressive speech synthesis using prosodic sketches as control conditions.
\newblock In \emph{IEEE International Conference on Acoustics, Speech and Signal Processing}, pages 1--5.

\bibitem[{Chen et~al.(2024{\natexlab{c}})Chen, Ma, Yan, Liang, Li, Xu, Niu, Zhu, Yang, Liu et~al.}]{chen2024slam-omni}
Wenxi Chen, Ziyang Ma, Ruiqi Yan, Yuzhe Liang, Xiquan Li, Ruiyang Xu, Zhikang Niu, Yanqiao Zhu, Yifan Yang, Zhanxun Liu, and 1 others. 2024{\natexlab{c}}.
\newblock {SLAM-Omni}: Timbre-controllable voice interaction system with single-stage training.
\newblock \emph{arXiv preprint arXiv:2412.15649}.

\bibitem[{Chen et~al.(2015)Chen, Xu, Liu, Sun, and Luan}]{chen2015joint}
Xinxiong Chen, Lei Xu, Zhiyuan Liu, Maosong Sun, and Huan-Bo Luan. 2015.
\newblock Joint learning of character and word embeddings.
\newblock In \emph{Twenty-fourth International Joint Conference on Artificial Intelligence}, pages 1236--1242.

\bibitem[{Chen et~al.(2024{\natexlab{d}})Chen, Niu, Ma, Deng, Wang, Zhao, Yu, and Chen}]{chen2024f5}
Yushen Chen, Zhikang Niu, Ziyang Ma, Keqi Deng, Chunhui Wang, Jian Zhao, Kai Yu, and Xie Chen. 2024{\natexlab{d}}.
\newblock {F5-TTS}: A fairytaler that fakes fluent and faithful speech with flow matching.
\newblock \emph{arXiv preprint arXiv:2410.06885}.

\bibitem[{Cho et~al.(2024)Cho, Oh, Kim, and Lee}]{cho2024emosphere++}
Deok-Hyeon Cho, Hyung-Seok Oh, Seung-Bin Kim, and Seong-Whan Lee. 2024.
\newblock {EmoSphere}++: Emotion-controllable zero-shot text-to-speech via emotion-adaptive spherical vector.
\newblock \emph{arXiv preprint arXiv:2411.02625}.

\bibitem[{Choi et~al.(2023)Choi, Hong, and Ro}]{choi2023diffv2s}
Jeongsoo Choi, Joanna Hong, and Yong~Man Ro. 2023.
\newblock {DiffV2S}: Diffusion-based video-to-speech synthesis with vision-guided speaker embedding.
\newblock In \emph{Proceedings of the IEEE/CVF International Conference on Computer Vision}, pages 7812--7821.

\bibitem[{Chowdhery et~al.(2023)Chowdhery, Narang, Devlin, Bosma, Mishra, Roberts, Barham, Chung, Sutton, Gehrmann et~al.}]{chowdhery2023palm}
Aakanksha Chowdhery, Sharan Narang, Jacob Devlin, Maarten Bosma, Gaurav Mishra, Adam Roberts, Paul Barham, Hyung~Won Chung, Charles Sutton, Sebastian Gehrmann, and 1 others. 2023.
\newblock Pa{LM}: Scaling language modeling with pathways.
\newblock \emph{Journal of Machine Learning Research}, 24(240):1--113.

\bibitem[{Cong et~al.(2024)Cong, Pan, Li, Qi, Peng, Hengel, Yang, and Huang}]{cong2024emodubber}
Gaoxiang Cong, Jiadong Pan, Liang Li, Yuankai Qi, Yuxin Peng, Anton van~den Hengel, Jian Yang, and Qingming Huang. 2024.
\newblock {EmoDubber}: Towards high quality and emotion controllable movie dubbing.
\newblock \emph{arXiv preprint arXiv:2412.08988}.

\bibitem[{Cooper et~al.(2020)Cooper, Lai, Yasuda, Fang, Wang, Chen, and Yamagishi}]{cooper2020zero}
Erica Cooper, Cheng-I Lai, Yusuke Yasuda, Fuming Fang, Xin Wang, Nanxin Chen, and Junichi Yamagishi. 2020.
\newblock Zero-shot multi-speaker text-to-speech with state-of-the-art neural speaker embeddings.
\newblock In \emph{IEEE International Conference on Acoustics, Speech and Signal Processing}, pages 6184--6188.

\bibitem[{D{\'e}fossez et~al.(2023{\natexlab{a}})D{\'e}fossez, Copet, Synnaeve, and Adi}]{fossez2023high}
Alexandre D{\'e}fossez, Jade Copet, Gabriel Synnaeve, and Yossi Adi. 2023{\natexlab{a}}.
\newblock High fidelity neural audio compression.
\newblock \emph{Transactions on Machine Learning Research}, pages 1--19.

\bibitem[{D{\'e}fossez et~al.(2023{\natexlab{b}})D{\'e}fossez, Copet, Synnaeve, and Adi}]{defossez2022encodec}
Alexandre D{\'e}fossez, Jade Copet, Gabriel Synnaeve, and Yossi Adi. 2023{\natexlab{b}}.
\newblock High fidelity neural audio compression.
\newblock \emph{Transactions on Machine Learning Research}.

\bibitem[{D{\'e}fossez et~al.(2024)D{\'e}fossez, Mazar{\'e}, Orsini, Royer, P{\'e}rez, J{\'e}gou, Grave, and Zeghidour}]{defossez2024moshi}
Alexandre D{\'e}fossez, Laurent Mazar{\'e}, Manu Orsini, Am{\'e}lie Royer, Patrick P{\'e}rez, Herv{\'e} J{\'e}gou, Edouard Grave, and Neil Zeghidour. 2024.
\newblock Moshi: a speech-text foundation model for real-time dialogue.
\newblock \emph{arXiv preprint arXiv:2410.00037}.

\bibitem[{Desplanques et~al.(2020)Desplanques, Thienpondt, and Demuynck}]{Brecht20ECAPA-TDNN}
Brecht Desplanques, Jenthe Thienpondt, and Kris Demuynck. 2020.
\newblock {ECAPA-TDNN:} emphasized channel attention, propagation and aggregation in {TDNN} based speaker verification.
\newblock In \emph{21st Annual Conference of the International Speech Communication Association}, pages 3830--3834.

\bibitem[{Devlin et~al.(2019)Devlin, Chang, Lee, and Toutanova}]{devlin2018bert}
Jacob Devlin, Ming-Wei Chang, Kenton Lee, and Kristina Toutanova. 2019.
\newblock {BERT}: Pre-training of deep bidirectional transformers for language understanding.
\newblock In \emph{Proceedings of the Conference of the Association for Computational Linguistics}, pages 4171--4186.

\bibitem[{Di et~al.(2024)Di, Chen, Liang, Zheng, Wang, and Ding}]{di2024bailing}
Xinhan Di, Zihao Chen, Yunming Liang, Junjie Zheng, Yihua Wang, and Chaofan Ding. 2024.
\newblock {Bailing-TTS}: Chinese dialectal speech synthesis towards human-like spontaneous representation.
\newblock \emph{arXiv preprint arXiv:2408.00284}.

\bibitem[{Donahue et~al.(2018)Donahue, McAuley, and Puckette}]{donahue2018wavegan}
Chris Donahue, Julian McAuley, and Miller Puckette. 2018.
\newblock Adversarial audio synthesis.
\newblock In \emph{International Conference on Learning Representations}.

\bibitem[{Du et~al.(2024)Du, Chen, Zhang, Hu, Lu, Yang, Hu, Zheng, Gu, Ma et~al.}]{du2024cosyvoice}
Zhihao Du, Qian Chen, Shiliang Zhang, Kai Hu, Heng Lu, Yexin Yang, Hangrui Hu, Siqi Zheng, Yue Gu, Ziyang Ma, and 1 others. 2024.
\newblock {CosyVoice}: A scalable multilingual zero-shot text-to-speech synthesizer based on supervised semantic tokens.
\newblock \emph{arXiv preprint arXiv:2407.05407}.

\bibitem[{Dutoit(1997)}]{dutoit1997introduction}
Thierry Dutoit. 1997.
\newblock \emph{{An Introduction to Text-to-Speech Synthesis}}, volume~3.
\newblock Springer Science \& Business Media.

\bibitem[{Elias et~al.(2021{\natexlab{a}})Elias, Zen, Shen, Zhang, Jia, Weiss, and Wu}]{elias2021parallel}
Isaac Elias, Heiga Zen, Jonathan Shen, Yu~Zhang, Ye~Jia, Ron~J Weiss, and Yonghui Wu. 2021{\natexlab{a}}.
\newblock Parallel {Tacotron}: Non-autoregressive and controllable {TTS}.
\newblock In \emph{IEEE International Conference on Acoustics, Speech and Signal Processing}, pages 5709--5713.

\bibitem[{Elias et~al.(2021{\natexlab{b}})Elias, Zen, Shen, Zhang, Jia, Weiss, and Wu}]{elias2021paralleltacotron}
Isaac Elias, Heiga Zen, Jonathan Shen, Yu~Zhang, Ye~Jia, Ron~J Weiss, and Yonghui Wu. 2021{\natexlab{b}}.
\newblock Parallel {Tacotron}: Non-autoregressive and controllable {TTS}.
\newblock In \emph{IEEE International Conference on Acoustics, Speech and Signal Processing}, pages 5709--5713.

\bibitem[{Eskimez et~al.(2024)Eskimez, Wang, Thakker, Li, Tsai, Xiao, Yang, Zhu, Tang, Tan et~al.}]{eskimez2024e2}
Sefik~Emre Eskimez, Xiaofei Wang, Manthan Thakker, Canrun Li, Chung-Hsien Tsai, Zhen Xiao, Hemin Yang, Zirun Zhu, Min Tang, Xu~Tan, and 1 others. 2024.
\newblock E2 {TTS}: Embarrassingly easy fully non-autoregressive zero-shot {TTS}.
\newblock In \emph{IEEE Spoken Language Technology Workshop}, pages 682--689.

\bibitem[{Fan et~al.(2015)Fan, Qian, Soong, and He}]{fan2015multi}
Yuchen Fan, Yao Qian, Frank~K Soong, and Lei He. 2015.
\newblock Multi-speaker modeling and speaker adaptation for {DNN}-based {TTS} synthesis.
\newblock In \emph{IEEE International Conference on Acoustics, Speech and Signal Processing}, pages 4475--4479.

\bibitem[{Fan et~al.(2014)Fan, Qian, Xie, and Soong}]{fan2014tts}
Yuchen Fan, Yao Qian, Feng-Long Xie, and Frank~K Soong. 2014.
\newblock {TTS} synthesis with bidirectional {LSTM} based recurrent neural networks.
\newblock In \emph{Proceedings of the Annual Conference of the International Speech Communication Association}, pages 1964--1968.

\bibitem[{Fang et~al.(2025)Fang, Guo, Zhou, Ma, Zhang, and Feng}]{fang2025llamaomni}
Qingkai Fang, Shoutao Guo, Yan Zhou, Zhengrui Ma, Shaolei Zhang, and Yang Feng. 2025.
\newblock {LLaMA-Omni}: Seamless speech interaction with large language models.
\newblock In \emph{International Conference on Learning Representations}, pages 1--18.

\bibitem[{Fukada et~al.(1992)Fukada, Tokuda, Kobayashi, and Imai}]{fukada1992adaptive}
T~Fukada, K~Tokuda, T~Kobayashi, and S~Imai. 1992.
\newblock An adaptive algorithm for mel-cepstral analysis of speech.
\newblock In \emph{IEEE International Conference on Acoustics, Speech, and Signal Processing}, volume~1, pages 137--140.

\bibitem[{Gao et~al.(2024)Gao, Zhang, Chen, Zhang, and Chen}]{gao2024emo}
Xiaoxue Gao, Chen Zhang, Yiming Chen, Huayun Zhang, and Nancy~F Chen. 2024.
\newblock {Emo-DPO}: Controllable emotional speech synthesis through direct preference optimization.
\newblock \emph{arXiv preprint arXiv:2409.10157}.

\bibitem[{Gao et~al.(2023)Gao, Morioka, Zhang, and Chen}]{gao2023e3}
Yuan Gao, Nobuyuki Morioka, Yu~Zhang, and Nanxin Chen. 2023.
\newblock E3 {TTS}: Easy end-to-end diffusion-based text to speech.
\newblock In \emph{IEEE Automatic Speech Recognition and Understanding Workshop}, pages 1--8.

\bibitem[{gil Lee et~al.(2023)gil Lee, Ping, Ginsburg, Catanzaro, and Yoon}]{lee2023bigvgan}
Sang gil Lee, Wei Ping, Boris Ginsburg, Bryan Catanzaro, and Sungroh Yoon. 2023.
\newblock Big{VGAN}: A universal neural vocoder with large-scale training.
\newblock In \emph{The Eleventh International Conference on Learning Representations}.

\bibitem[{Goodfellow et~al.(2020)Goodfellow, Pouget-Abadie, Mirza, Xu, Warde-Farley, Ozair, Courville, and Bengio}]{goodfellow2020gan}
Ian Goodfellow, Jean Pouget-Abadie, Mehdi Mirza, Bing Xu, David Warde-Farley, Sherjil Ozair, Aaron Courville, and Yoshua Bengio. 2020.
\newblock Generative adversarial networks.
\newblock \emph{Communications of the ACM}, 63(11):139--144.

\bibitem[{Goto et~al.(2020)Goto, Onishi, Saito, Tachibana, and Mori}]{goto2020face2speech}
Shunsuke Goto, Kotaro Onishi, Yuki Saito, Kentaro Tachibana, and Koichiro Mori. 2020.
\newblock {Face2Speech}: Towards multi-speaker text-to-speech synthesis using an embedding vector predicted from a face image.
\newblock In \emph{Proceedings of the Annual Conference of the International Speech Communication Association}, pages 1321--1325.

\bibitem[{Gu et~al.(2023)Gu, Bian, Lei, Weng, and Su}]{gu2023durian}
Yu~Gu, Yianrao Bian, Guangzhi Lei, Chao Weng, and Dan Su. 2023.
\newblock {DurIAN-E}: Duration informed attention network for expressive text-to-speech synthesis.
\newblock \emph{arXiv preprint arXiv:2309.12792}.

\bibitem[{Guo et~al.(2024{\natexlab{a}})Guo, Liu, Shen, Wu, Xie, Xie, and Xu}]{guo2024fireredtts}
Hao-Han Guo, Kun Liu, Fei-Yu Shen, Yi-Chen Wu, Feng-Long Xie, Kun Xie, and Kai-Tuo Xu. 2024{\natexlab{a}}.
\newblock {FireRedTTS}: A foundation text-to-speech framework for industry-level generative speech applications.
\newblock \emph{arXiv preprint arXiv:2409.03283}.

\bibitem[{Guo et~al.(2024{\natexlab{b}})Guo, Xie, Yang, Wu, and Meng}]{guo2024speaking}
Haohan Guo, Fenglong Xie, Dongchao Yang, Xixin Wu, and Helen Meng. 2024{\natexlab{b}}.
\newblock Speaking from coarse to fine: Improving neural codec language model via multi-scale speech coding and generation.
\newblock \emph{arXiv preprint arXiv:2409.11630}.

\bibitem[{Guo et~al.(2023)Guo, Leng, Wu, Zhao, and Tan}]{guo2023prompttts}
Zhifang Guo, Yichong Leng, Yihan Wu, Sheng Zhao, and Xu~Tan. 2023.
\newblock {PromptTTS}: Controllable text-to-speech with text descriptions.
\newblock In \emph{IEEE International Conference on Acoustics, Speech and Signal Processing}, pages 1--5.

\bibitem[{Han et~al.(2024)Han, Zhou, Liu, Chen, Meng, Qian, Liu, Zhao, Li, and Wei}]{han2024vall}
Bing Han, Long Zhou, Shujie Liu, Sanyuan Chen, Lingwei Meng, Yanming Qian, Yanqing Liu, Sheng Zhao, Jinyu Li, and Furu Wei. 2024.
\newblock {VALL-E R}: Robust and efficient zero-shot text-to-speech synthesis via monotonic alignment.
\newblock \emph{arXiv preprint arXiv:2406.07855}.

\bibitem[{Hannun et~al.(2014)Hannun, Case, Casper, Catanzaro, Diamos, Elsen, Prenger, Satheesh, Sengupta, Coates et~al.}]{hannun2014deepspeech}
Awni Hannun, Carl Case, Jared Casper, Bryan Catanzaro, Greg Diamos, Erich Elsen, Ryan Prenger, Sanjeev Satheesh, Shubho Sengupta, Adam Coates, and 1 others. 2014.
\newblock {Deep Speech}: Scaling up end-to-end speech recognition.
\newblock \emph{arXiv preprint arXiv:1412.5567}.

\bibitem[{He et~al.(2024)He, Liu, and Li}]{he2024multi}
Shuwei He, Rui Liu, and Haizhou Li. 2024.
\newblock Multi-source spatial knowledge understanding for immersive visual text-to-speech.
\newblock \emph{arXiv preprint arXiv:2410.14101}.

\bibitem[{Heusel et~al.(2017)Heusel, Ramsauer, Unterthiner, Nessler, and Hochreiter}]{heusel2017fid}
Martin Heusel, Hubert Ramsauer, Thomas Unterthiner, Bernhard Nessler, and Sepp Hochreiter. 2017.
\newblock {GANs} trained by a two time-scale update rule converge to a local {Nash} equilibrium.
\newblock \emph{Advances in Neural Information Processing Systems}, 30.

\bibitem[{Ho et~al.(2020)Ho, Jain, and Abbeel}]{ho2020denoising}
Jonathan Ho, Ajay Jain, and Pieter Abbeel. 2020.
\newblock Denoising diffusion probabilistic models.
\newblock \emph{Advances in Neural Information Processing Systems}, 33:6840--6851.

\bibitem[{Hsu et~al.(2021)Hsu, Bolte, Tsai, Lakhotia, Salakhutdinov, and Mohamed}]{hsu2021hubert}
Wei-Ning Hsu, Benjamin Bolte, Yao-Hung~Hubert Tsai, Kushal Lakhotia, Ruslan Salakhutdinov, and Abdelrahman Mohamed. 2021.
\newblock {HuBERT}: Self-supervised speech representation learning by masked prediction of hidden units.
\newblock \emph{IEEE/ACM Transactions on Audio, Speech, and Language Processing}, 29:3451--3460.

\bibitem[{Hsu et~al.(2019)Hsu, Zhang, Weiss, Chung, Wang, Wu, and Glass}]{Hsu2019Disentangling}
Wei-Ning Hsu, Yu~Zhang, Ron~J. Weiss, Yu-An Chung, Yuxuan Wang, Yonghui Wu, and James Glass. 2019.
\newblock \href {https://doi.org/10.1109/ICASSP.2019.8683561} {Disentangling correlated speaker and noise for speech synthesis via data augmentation and adversarial factorization}.
\newblock In \emph{ICASSP 2019 - 2019 IEEE International Conference on Acoustics, Speech and Signal Processing (ICASSP)}, pages 5901--5905.

\bibitem[{Hsu et~al.(2018)Hsu, Zhang, Weiss, Zen, Wu, Wang, Cao, Jia, Chen, Shen et~al.}]{hsu2018hierarchical}
Wei-Ning Hsu, Yu~Zhang, Ron~J Weiss, Heiga Zen, Yonghui Wu, Yuxuan Wang, Yuan Cao, Ye~Jia, Zhifeng Chen, Jonathan Shen, and 1 others. 2018.
\newblock Hierarchical generative modeling for controllable speech synthesis.
\newblock In \emph{International Conference on Learning Representations}.

\bibitem[{Huang et~al.(2025)Huang, Wu, Wang, Yan, Hu, Feng, Tian, Shen, Li, Chen et~al.}]{huang2025stepaudio}
Ailin Huang, Boyong Wu, Bruce Wang, Chao Yan, Chen Hu, Chengli Feng, Fei Tian, Feiyu Shen, Jingbei Li, Mingrui Chen, and 1 others. 2025.
\newblock {Step-Audio}: Unified understanding and generation in intelligent speech interaction.
\newblock \emph{arXiv preprint arXiv:2502.11946}.

\bibitem[{Huang et~al.(2024{\natexlab{a}})Huang, Hu, Wang, Wang, Cheng, Jiang, Ye, Yang, Liu, Gao, and Zhao}]{huang2024instructspeech}
Rongjie Huang, Ruofan Hu, Yongqi Wang, Zehan Wang, Xize Cheng, Ziyue Jiang, Zhenhui Ye, Dongchao Yang, Luping Liu, Peng Gao, and Zhou Zhao. 2024{\natexlab{a}}.
\newblock {InstructSpeech}: Following speech editing instructions via large language models.
\newblock In \emph{Forty-first International Conference on Machine Learning}.

\bibitem[{Huang et~al.(2023{\natexlab{a}})Huang, Huang, Yang, Ren, Liu, Li, Ye, Liu, Yin, and Zhao}]{huang2023makeanaudio}
Rongjie Huang, Jiawei Huang, Dongchao Yang, Yi~Ren, Luping Liu, Mingze Li, Zhenhui Ye, Jinglin Liu, Xiang Yin, and Zhou Zhao. 2023{\natexlab{a}}.
\newblock {Make-An-Audio}: Text-to-audio generation with prompt-enhanced diffusion models.
\newblock In \emph{International Conference on Machine Learning}, pages 13916--13932.

\bibitem[{Huang et~al.(2022{\natexlab{a}})Huang, Lam, Wang, Su, Yu, Ren, and Zhao}]{huang2022fastdiff}
Rongjie Huang, Max W.~Y. Lam, Jun Wang, Dan Su, Dong Yu, Yi~Ren, and Zhou Zhao. 2022{\natexlab{a}}.
\newblock {FastDiff}: A fast conditional diffusion model for high-quality speech synthesis.
\newblock In \emph{Proceedings of the Thirty-First International Joint Conference on Artificial Intelligence}, pages 4157--4163.

\bibitem[{Huang et~al.(2024{\natexlab{b}})Huang, Li, Yang, Shi, Chang, Ye, Wu, Hong, Huang, Liu et~al.}]{huang2024audiogpt}
Rongjie Huang, Mingze Li, Dongchao Yang, Jiatong Shi, Xuankai Chang, Zhenhui Ye, Yuning Wu, Zhiqing Hong, Jiawei Huang, Jinglin Liu, and 1 others. 2024{\natexlab{b}}.
\newblock {AudioGPT}: Understanding and generating speech, music, sound, and talking head.
\newblock In \emph{Proceedings of the AAAI Conference on Artificial Intelligence}, volume~38, pages 23802--23804.

\bibitem[{Huang et~al.(2022{\natexlab{b}})Huang, Ren, Liu, Cui, and Zhao}]{huang2022generspeech}
Rongjie Huang, Yi~Ren, Jinglin Liu, Chenye Cui, and Zhou Zhao. 2022{\natexlab{b}}.
\newblock {GenerSpeech}: Towards style transfer for generalizable out-of-domain text-to-speech.
\newblock In \emph{Advances in Neural Information Processing Systems}, pages 1--14.

\bibitem[{Huang et~al.(2023{\natexlab{b}})Huang, Zhang, Wang, Yang, Liu, Ye, Jiang, Weng, Zhao, and Yu}]{huang2023makeavoice}
Rongjie Huang, Chunlei Zhang, Yongqi Wang, Dongchao Yang, Luping Liu, Zhenhui Ye, Ziyue Jiang, Chao Weng, Zhou Zhao, and Dong Yu. 2023{\natexlab{b}}.
\newblock {Make-A-Voice}: Unified voice synthesis with discrete representation.
\newblock \emph{arXiv preprint arXiv:2305.19269}.

\bibitem[{Hunt and Black(1996)}]{Hunt1996373}
Andrew~J. Hunt and Alan~W. Black. 1996.
\newblock Unit selection in a concatenative speech synthesis system using a large speech database.
\newblock In \emph{IEEE International Conference on Acoustics, Speech, and Signal Processing}, volume~1, pages 373--376.

\bibitem[{Inoue et~al.(2024)Inoue, Zhou, Wang, and Li}]{inoue2024HED}
Sho Inoue, Kun Zhou, Shuai Wang, and Haizhou Li. 2024.
\newblock Hierarchical control of emotion rendering in speech synthesis.
\newblock \emph{arXiv preprint arXiv:2412.12498}.

\bibitem[{Itakura(1975)}]{itakura1975line}
Fumitada Itakura. 1975.
\newblock Line spectrum representation of linear predictor coefficients of speech signals.
\newblock \emph{The Journal of the Acoustical Society of America}, 57(S1):S35--S35.

\bibitem[{Jang et~al.(2021)Jang, Lim, Yoon, Kim, and Kim}]{JangLYKK21univnet}
Won Jang, Dan Lim, Jaesam Yoon, Bongwan Kim, and Juntae Kim. 2021.
\newblock {UnivNet}: A neural vocoder with multi-resolution spectrogram discriminators for high-fidelity waveform generation.
\newblock In \emph{Annual Conference of the International Speech Communication Association}, pages 2207--2211.

\bibitem[{Jeong et~al.(2021)Jeong, Kim, Cheon, Choi, and Kim}]{jeong2021difftts}
Myeonghun Jeong, Hyeongju Kim, Sung~Jun Cheon, Byoung~Jin Choi, and Nam~Soo Kim. 2021.
\newblock {Diff-TTS}: A denoising diffusion model for text-to-speech.
\newblock In \emph{Proceedings of the Annual Conference of the International Speech Communication Association}, pages 3605--3609.

\bibitem[{Ji et~al.(2024{\natexlab{a}})Ji, Jiang, Wang, Zuo, and Zhao}]{ji2024mobilespeech}
Shengpeng Ji, Ziyue Jiang, Hanting Wang, Jialong Zuo, and Zhou Zhao. 2024{\natexlab{a}}.
\newblock {MobileSpeech}: A fast and high-fidelity framework for mobile zero-shot text-to-speech.
\newblock In \emph{The 62nd Annual Meeting of the Association for Computational Linguistics}, pages 13588--13600.

\bibitem[{Ji et~al.(2025)Ji, Jiang, Wang, Chen, Fang, Zuo, Yang, Cheng, Wang, Li, Zhang, Yang, Huang, Jiang, Chen, Zheng, and Zhao}]{ji2025wavtokenizer}
Shengpeng Ji, Ziyue Jiang, Wen Wang, Yifu Chen, Minghui Fang, Jialong Zuo, Qian Yang, Xize Cheng, Zehan Wang, Ruiqi Li, Ziang Zhang, Xiaoda Yang, Rongjie Huang, Yidi Jiang, Qian Chen, Siqi Zheng, and Zhou Zhao. 2025.
\newblock {WavTokenizer}: an efficient acoustic discrete codec tokenizer for audio language modeling.
\newblock In \emph{The Thirteenth International Conference on Learning Representations}.

\bibitem[{Ji et~al.(2024{\natexlab{b}})Ji, Zuo, Fang, Jiang, Chen, Duan, Huai, and Zhao}]{ji2024textrolspeech}
Shengpeng Ji, Jialong Zuo, Minghui Fang, Ziyue Jiang, Feiyang Chen, Xinyu Duan, Baoxing Huai, and Zhou Zhao. 2024{\natexlab{b}}.
\newblock {TextrolSpeech}: A text style control speech corpus with codec language text-to-speech models.
\newblock In \emph{IEEE International Conference on Acoustics, Speech and Signal Processing}, pages 10301--10305.

\bibitem[{Ji et~al.(2024{\natexlab{c}})Ji, Zuo, Wang, Fang, Zheng, Chen, Jiang, Huang, Wang, Cheng et~al.}]{ji2024controlspeech}
Shengpeng Ji, Jialong Zuo, Wen Wang, Minghui Fang, Siqi Zheng, Qian Chen, Ziyue Jiang, Hai Huang, Zehan Wang, Xize Cheng, and 1 others. 2024{\natexlab{c}}.
\newblock {ControlSpeech}: Towards simultaneous zero-shot speaker cloning and zero-shot language style control with decoupled codec.
\newblock \emph{arXiv preprint arXiv:2406.01205}.

\bibitem[{Jiang et~al.(2024)Jiang, Liu, Ren, He, Ye, Ji, Yang, Zhang, Wei, Wang et~al.}]{jiang2024mega}
Ziyue Jiang, Jinglin Liu, Yi~Ren, Jinzheng He, Zhenhui Ye, Shengpeng Ji, Qian Yang, Chen Zhang, Pengfei Wei, Chunfeng Wang, and 1 others. 2024.
\newblock Mega-{TTS} 2: Boosting prompting mechanisms for zero-shot speech synthesis.
\newblock In \emph{The Twelfth International Conference on Learning Representations}.

\bibitem[{Jiang et~al.(2023)Jiang, Ren, Ye, Liu, Zhang, Yang, Ji, Huang, Wang, Yin et~al.}]{jiang2023megavoic}
Ziyue Jiang, Yi~Ren, Zhenhui Ye, Jinglin Liu, Chen Zhang, Qian Yang, Shengpeng Ji, Rongjie Huang, Chunfeng Wang, Xiang Yin, and 1 others. 2023.
\newblock Mega-{TTS}: Zero-shot text-to-speech at scale with intrinsic inductive bias.
\newblock \emph{arXiv preprint arXiv:2306.03509}.

\bibitem[{Jiao et~al.(2021)Jiao, Gabry{\'s}, Tinchev, Putrycz, Korzekwa, and Klimkov}]{jiao2021upwavenet}
Yunlong Jiao, Adam Gabry{\'s}, Georgi Tinchev, Bartosz Putrycz, Daniel Korzekwa, and Viacheslav Klimkov. 2021.
\newblock Universal neural vocoding with parallel {WaveNet}.
\newblock In \emph{IEEE International Conference on Acoustics, Speech and Signal Processing}, pages 6044--6048.

\bibitem[{Jin et~al.(2024)Jin, Jia, Wang, Li, Zhou, Zhou, Qin, and Wu}]{jin2024speechcraft}
Zeyu Jin, Jia Jia, Qixin Wang, Kehan Li, Shuoyi Zhou, Songtao Zhou, Xiaoyu Qin, and Zhiyong Wu. 2024.
\newblock {SpeechCraft}: A fine-grained expressive speech dataset with natural language description.
\newblock In \emph{Proceedings of the 32nd ACM International Conference on Multimedia}, pages 1255--1264.

\bibitem[{Ju et~al.(2024)Ju, Wang, Shen, Tan, Xin, Yang, Liu, Leng, Song, Tang, Wu, Qin, Li, Ye, Zhang, Bian, He, Li, and sheng zhao}]{ju2024naturalspeech3}
Zeqian Ju, Yuancheng Wang, Kai Shen, Xu~Tan, Detai Xin, Dongchao Yang, Eric Liu, Yichong Leng, Kaitao Song, Siliang Tang, Zhizheng Wu, Tao Qin, Xiangyang Li, Wei Ye, Shikun Zhang, Jiang Bian, Lei He, Jinyu Li, and sheng zhao. 2024.
\newblock {NaturalSpeech} 3: Zero-shot speech synthesis with factorized codec and diffusion models.
\newblock In \emph{International Conference on Machine Learning}, pages 1--19.

\bibitem[{Kalchbrenner et~al.(2018)Kalchbrenner, Elsen, Simonyan, Noury, Casagrande, Lockhart, Stimberg, Oord, Dieleman, and Kavukcuoglu}]{kalchbrenner2018wavernn}
Nal Kalchbrenner, Erich Elsen, Karen Simonyan, Seb Noury, Norman Casagrande, Edward Lockhart, Florian Stimberg, Aaron Oord, Sander Dieleman, and Koray Kavukcuoglu. 2018.
\newblock Efficient neural audio synthesis.
\newblock In \emph{International Conference on Machine Learning}, pages 2410--2419.

\bibitem[{Kaneko et~al.(2022)Kaneko, Tanaka, Kameoka, and Seki}]{kaneko2022istftnet}
Takuhiro Kaneko, Kou Tanaka, Hirokazu Kameoka, and Shogo Seki. 2022.
\newblock istftnet: Fast and lightweight mel-spectrogram vocoder incorporating inverse short-time {Fourier} transform.
\newblock In \emph{IEEE International Conference on Acoustics, Speech and Signal Processing}, pages 6207--6211.

\bibitem[{Kang et~al.(2023)Kang, Min, and Hwang}]{kang2023grad}
Minki Kang, Dongchan Min, and Sung~Ju Hwang. 2023.
\newblock {Grad-StyleSpeech}: Any-speaker adaptive text-to-speech synthesis with diffusion models.
\newblock In \emph{IEEE International Conference on Acoustics, Speech and Signal Processing}, pages 1--5.

\bibitem[{Kawahara et~al.(1999)Kawahara, Masuda-Katsuse, and De~Cheveigne}]{kawahara1999restructuring}
Hideki Kawahara, Ikuyo Masuda-Katsuse, and Alain De~Cheveigne. 1999.
\newblock Restructuring speech representations using a pitch-adaptive time--frequency smoothing and an instantaneous-frequency-based {F0} extraction: Possible role of a repetitive structure in sounds.
\newblock \emph{Speech Communication}, 27(3-4):187--207.

\bibitem[{Kharitonov et~al.(2023)Kharitonov, Vincent, Borsos, Marinier, Girgin, Pietquin, Sharifi, Tagliasacchi, and Zeghidour}]{kharitonov2023speak}
Eugene Kharitonov, Damien Vincent, Zal{\'a}n Borsos, Rapha{\"e}l Marinier, Sertan Girgin, Olivier Pietquin, Matt Sharifi, Marco Tagliasacchi, and Neil Zeghidour. 2023.
\newblock Speak, read and prompt: High-fidelity text-to-speech with minimal supervision.
\newblock \emph{Transactions of the Association for Computational Linguistics}, 11:1703--1718.

\bibitem[{Kim et~al.(2023)Kim, Hong, and Choi}]{kim2023sc}
Daegyeom Kim, Seongho Hong, and Yong-Hoon Choi. 2023.
\newblock {SC VALL-E}: Style-controllable zero-shot text to speech synthesizer.
\newblock \emph{arXiv preprint arXiv:2307.10550}.

\bibitem[{Kim et~al.(2024{\natexlab{a}})Kim, Lee, Chung, and Cho}]{kim2024clam}
Jaehyeon Kim, Keon Lee, Seungjun Chung, and Jaewoong Cho. 2024{\natexlab{a}}.
\newblock {CLaM-TTS}: Improving neural codec language model for zero-shot text-to-speech.
\newblock \emph{arXiv preprint arXiv:2404.02781}.

\bibitem[{Kim et~al.(2021)Kim, Cheon, Choi, Kim, and Kim}]{kim2021expressive}
Minchan Kim, Sung~Jun Cheon, Byoung~Jin Choi, Jong~Jin Kim, and Nam~Soo Kim. 2021.
\newblock Expressive text-to-speech using style tag.
\newblock \emph{Annual Conference of the International Speech Communication Association}, pages 4663--4667.

\bibitem[{Kim et~al.(2024{\natexlab{b}})Kim, Chung, Ji, Kang, and Choi}]{kim2024speak}
Miseul Kim, Soo-Whan Chung, Youna Ji, Hong-Goo Kang, and Min-Seok Choi. 2024{\natexlab{b}}.
\newblock Speak in the {Scene}: Diffusion-based acoustic scene transfer toward immersive speech generation.
\newblock In \emph{Annual Conference of the International Speech Communication Association}, pages 4883--4887.

\bibitem[{Kim et~al.(2024{\natexlab{c}})Kim, Shih, Santos, Bakhturina, Desta, Valle, Yoon, Catanzaro et~al.}]{kim2024p}
Sungwon Kim, Kevin Shih, Joao~Felipe Santos, Evelina Bakhturina, Mikyas Desta, Rafael Valle, Sungroh Yoon, Bryan Catanzaro, and 1 others. 2024{\natexlab{c}}.
\newblock {P-Flow}: a fast and data-efficient zero-shot {TTS} through speech prompting.
\newblock \emph{Advances in Neural Information Processing Systems}, 36.

\bibitem[{Kim et~al.(2024{\natexlab{d}})Kim, Shih, Santos, Bakhturina, Desta, Valle, Yoon, Catanzaro et~al.}]{kim2024pflow}
Sungwon Kim, Kevin Shih, Joao~Felipe Santos, Evelina Bakhturina, Mikyas Desta, Rafael Valle, Sungroh Yoon, Bryan Catanzaro, and 1 others. 2024{\natexlab{d}}.
\newblock {P-Flow}: a fast and data-efficient zero-shot {TTS} through speech prompting.
\newblock \emph{Advances in Neural Information Processing Systems}, 36:74213--74228.

\bibitem[{King(2014)}]{king2014measuring}
Simon King. 2014.
\newblock Measuring a decade of progress in text-to-speech.
\newblock \emph{Loquens}, 1(1):e006--e006.

\bibitem[{Klatt(1987)}]{klatt1987review}
Dennis~H Klatt. 1987.
\newblock Review of text-to-speech conversion for english.
\newblock \emph{The Journal of the Acoustical Society of America}, 82(3):737--793.

\bibitem[{Kobyzev et~al.(2020)Kobyzev, Prince, and Brubaker}]{kobyzev2020normalizing}
Ivan Kobyzev, Simon~JD Prince, and Marcus~A Brubaker. 2020.
\newblock Normalizing flows: An introduction and review of current methods.
\newblock \emph{IEEE Transactions on Pattern Analysis and Machine Intelligence}, 43(11):3964--3979.

\bibitem[{Kominek et~al.(2008)Kominek, Schultz, and Black}]{kominek2008synthesizer}
John Kominek, Tanja Schultz, and Alan~W Black. 2008.
\newblock Synthesizer voice quality of new languages calibrated with mean mel cepstral distortion.
\newblock In \emph{Spoken Language Technologies for Under-Resourced Languages}, pages 63--68.

\bibitem[{Kong et~al.(2020)Kong, Kim, and Bae}]{kong2020hifigan}
Jungil Kong, Jaehyeon Kim, and Jaekyoung Bae. 2020.
\newblock {HiFi-GAN}: Generative adversarial networks for efficient and high fidelity speech synthesis.
\newblock \emph{Advances in Neural Information Processing Systems}, 33:17022--17033.

\bibitem[{Kong et~al.(2021)Kong, Ping, Huang, Zhao, and Catanzaro}]{kong2021diffwave}
Zhifeng Kong, Wei Ping, Jiaji Huang, Kexin Zhao, and Bryan Catanzaro. 2021.
\newblock {DiffWave}: A versatile diffusion model for audio synthesis.
\newblock In \emph{International Conference on Learning Representations}, pages 1--17.

\bibitem[{Kumar et~al.(2019)Kumar, Kumar, De~Boissiere, Gestin, Teoh, Sotelo, De~Brebisson, Bengio, and Courville}]{kumar2019melgan}
Kundan Kumar, Rithesh Kumar, Thibault De~Boissiere, Lucas Gestin, Wei~Zhen Teoh, Jose Sotelo, Alexandre De~Brebisson, Yoshua Bengio, and Aaron~C Courville. 2019.
\newblock {MelGAN}: Generative adversarial networks for conditional waveform synthesis.
\newblock \emph{Advances in Neural Information Processing Systems}, 32:14920--14921.

\bibitem[{Kumar et~al.(2024)Kumar, Seetharaman, Luebs, Kumar, and Kumar}]{kumar2024dac}
Rithesh Kumar, Prem Seetharaman, Alejandro Luebs, Ishaan Kumar, and Kundan Kumar. 2024.
\newblock High-fidelity audio compression with improved {RVQGAN}.
\newblock \emph{Advances in Neural Information Processing Systems}, 36.

\bibitem[{{\L}ajszczak et~al.(2024){\L}ajszczak, C{\'a}mbara, Li, Beyhan, van Korlaar, Yang, Joly, Mart{\'\i}n-Cortinas, Abbas, Michalski et~al.}]{lajszczak2024base}
Mateusz {\L}ajszczak, Guillermo C{\'a}mbara, Yang Li, Fatih Beyhan, Arent van Korlaar, Fan Yang, Arnaud Joly, {\'A}lvaro Mart{\'\i}n-Cortinas, Ammar Abbas, Adam Michalski, and 1 others. 2024.
\newblock {BASE TTS}: Lessons from building a billion-parameter text-to-speech model on 100k hours of data.
\newblock \emph{arXiv preprint arXiv:2402.08093}.

\bibitem[{{\L}a{\'n}cucki(2021)}]{lancucki2021fastpitch}
Adrian {\L}a{\'n}cucki. 2021.
\newblock {FastPitch}: Parallel text-to-speech with pitch prediction.
\newblock In \emph{IEEE International Conference on Acoustics, Speech and Signal Processing}, pages 6588--6592.

\bibitem[{Le et~al.(2024)Le, Vyas, Shi, Karrer, Sari, Moritz, Williamson, Manohar, Adi, Mahadeokar et~al.}]{le2024voicebox}
Matthew Le, Apoorv Vyas, Bowen Shi, Brian Karrer, Leda Sari, Rashel Moritz, Mary Williamson, Vimal Manohar, Yossi Adi, Jay Mahadeokar, and 1 others. 2024.
\newblock Voicebox: Text-guided multilingual universal speech generation at scale.
\newblock \emph{Advances in Neural Information Processing Systems}, 36.

\bibitem[{Lee et~al.(2025)Lee, Kim, Kim, Chung, and Cho}]{lee2025dittotts}
Keon Lee, Dong~Won Kim, Jaehyeon Kim, Seungjun Chung, and Jaewoong Cho. 2025.
\newblock \href {https://openreview.net/forum?id=hQvX9MBowC} {Di{TT}o-{TTS}: Diffusion transformers for scalable text-to-speech without domain-specific factors}.
\newblock In \emph{The Thirteenth International Conference on Learning Representations}.

\bibitem[{Lee et~al.(2023{\natexlab{a}})Lee, Park, and Kim}]{lee2023dailytalk}
Keon Lee, Kyumin Park, and Daeyoung Kim. 2023{\natexlab{a}}.
\newblock {DailyTalk}: Spoken dialogue dataset for conversational text-to-speech.
\newblock In \emph{IEEE International Conference on Acoustics, Speech and Signal Processing}, pages 1--5.

\bibitem[{Lee et~al.(2023{\natexlab{b}})Lee, Choi, Kim, and Lee}]{lee2023hierspeech++}
Sang-Hoon Lee, Ha-Yeong Choi, Seung-Bin Kim, and Seong-Whan Lee. 2023{\natexlab{b}}.
\newblock {HierSpeech++}: Bridging the gap between semantic and acoustic representation of speech by hierarchical variational inference for zero-shot speech synthesis.
\newblock \emph{arXiv preprint arXiv:2311.12454}.

\bibitem[{Lee et~al.(2021)Lee, Yoon, Noh, Kim, and Lee}]{lee2021multi}
Sang-Hoon Lee, Hyun-Wook Yoon, Hyeong-Rae Noh, Ji-Hoon Kim, and Seong-Whan Lee. 2021.
\newblock Multi-spectrogan: High-diversity and high-fidelity spectrogram generation with adversarial style combination for speech synthesis.
\newblock In \emph{Proceedings of the AAAI Conference on Artificial Intelligence}, volume~35, pages 13198--13206.

\bibitem[{Lee et~al.(2024)Lee, Yeon, Nam, and Chung}]{lee2024voiceldm}
Yeonghyeon Lee, Inmo Yeon, Juhan Nam, and Joon~Son Chung. 2024.
\newblock {VoiceLDM}: Text-to-speech with environmental context.
\newblock In \emph{IEEE International Conference on Acoustics, Speech and Signal Processing}, pages 12566--12571.

\bibitem[{Lei et~al.(2022)Lei, Yang, Wang, and Xie}]{lei2022msemotts}
Yi~Lei, Shan Yang, Xinsheng Wang, and Lei Xie. 2022.
\newblock {MsEmoTTS}: Multi-scale emotion transfer, prediction, and control for emotional speech synthesis.
\newblock \emph{IEEE/ACM Transactions on Audio, Speech, and Language Processing}, 30:853--864.

\bibitem[{Leng et~al.(2023)Leng, Guo, Shen, Tan, Ju, Liu, Liu, Yang, Zhang, Song et~al.}]{leng2023prompttts2}
Yichong Leng, Zhifang Guo, Kai Shen, Xu~Tan, Zeqian Ju, Yanqing Liu, Yufei Liu, Dongchao Yang, Leying Zhang, Kaitao Song, and 1 others. 2023.
\newblock {PromptTTS} 2: Describing and generating voices with text prompt.
\newblock In \emph{The Twelfth International Conference on Learning Representations}.

\bibitem[{Li et~al.(2025{\natexlab{a}})Li, Li, Wang, Hu, Xie, Yang, and Xie}]{li2025flespeech}
Hanzhao Li, Yuke Li, Xinsheng Wang, Jingbin Hu, Qicong Xie, Shan Yang, and Lei Xie. 2025{\natexlab{a}}.
\newblock {FleSpeech}: Flexibly controllable speech generation with various prompts.
\newblock \emph{arXiv preprint arXiv:2501.04644}.

\bibitem[{Li et~al.(2022)Li, Wang, Xie, Wang, and Xie}]{li2022cross}
Tao Li, Xinsheng Wang, Qicong Xie, Zhichao Wang, and Lei Xie. 2022.
\newblock Cross-speaker emotion disentangling and transfer for end-to-end speech synthesis.
\newblock \emph{IEEE/ACM Transactions on Audio, Speech, and Language Processing}, 30:1448--1460.

\bibitem[{Li et~al.(2021)Li, Yang, Xue, and Xie}]{li2021controllable}
Tao Li, Shan Yang, Liumeng Xue, and Lei Xie. 2021.
\newblock Controllable emotion transfer for end-to-end speech synthesis.
\newblock In \emph{12th International Symposium on Chinese Spoken Language Processing}, pages 1--5.

\bibitem[{Li et~al.(2025{\natexlab{b}})Li, Han, and Mesgarani}]{li2025styletts}
Yinghao~Aaron Li, Cong Han, and Nima Mesgarani. 2025{\natexlab{b}}.
\newblock {StyleTTS}: A style-based generative model for natural and diverse text-to-speech synthesis.
\newblock \emph{IEEE Journal of Selected Topics in Signal Processing}, 19(1):283--296.

\bibitem[{Li et~al.(2023)Li, Han, Raghavan, Mischler, and Mesgarani}]{li2023styletts2}
Yinghao~Aaron Li, Cong Han, Vinay~S Raghavan, Gavin Mischler, and Nima Mesgarani. 2023.
\newblock Style{TTS} 2: Towards human-level text-to-speech through style diffusion and adversarial training with large speech language models.
\newblock In \emph{37th Conference on Neural Information Processing Systems}, pages 1--28.

\bibitem[{Li et~al.(2024)Li, Jiang, Han, and Mesgarani}]{li2024stylettszs}
Yinghao~Aaron Li, Xilin Jiang, Cong Han, and Nima Mesgarani. 2024.
\newblock {StyleTTS-ZS}: Efficient high-quality zero-shot text-to-speech synthesis with distilled time-varying style diffusion.
\newblock \emph{arXiv preprint arXiv:2409.10058}.

\bibitem[{Lian et~al.(2025)Lian, Wang, and Huang}]{lian2025apg}
Zhicheng Lian, Lizhi Wang, and Hua Huang. 2025.
\newblock {APG-MOS}: Auditory perception guided-{MOS} predictor for synthetic speech.
\newblock \emph{arXiv preprint arXiv:2504.20447}.

\bibitem[{Liao et~al.(2024)Liao, Wang, Li, Cheng, Zhang, Zhou, and Xing}]{liao2024fishspeech}
Shijia Liao, Yuxuan Wang, Tianyu Li, Yifan Cheng, Ruoyi Zhang, Rongzhi Zhou, and Yijin Xing. 2024.
\newblock {Fish-Speech}: Leveraging large language models for advanced multilingual text-to-speech synthesis.
\newblock \emph{arXiv preprint arXiv:2411.01156}.

\bibitem[{Ling et~al.(2009)Ling, Richmond, Yamagishi, and Wang}]{ling2009integrating}
Zhen-Hua Ling, Korin Richmond, Junichi Yamagishi, and Ren-Hua Wang. 2009.
\newblock Integrating articulatory features into {HMM}-based parametric speech synthesis.
\newblock \emph{IEEE Transactions on Audio, Speech, and Language Processing}, 17(6):1171--1185.

\bibitem[{Lipman et~al.(2023)Lipman, Chen, Ben-Hamu, Nickel, and Le}]{lipman2023flowmatching}
Yaron Lipman, Ricky T.~Q. Chen, Heli Ben-Hamu, Maximilian Nickel, and Matthew Le. 2023.
\newblock \href {https://openreview.net/forum?id=PqvMRDCJT9t} {Flow matching for generative modeling}.
\newblock In \emph{The Eleventh International Conference on Learning Representations}.

\bibitem[{Liu et~al.(2024{\natexlab{a}})Liu, Le, Vyas, Shi, Tjandra, and Hsu}]{liu2023generative}
Alexander~H Liu, Matthew Le, Apoorv Vyas, Bowen Shi, Andros Tjandra, and Wei-Ning Hsu. 2024{\natexlab{a}}.
\newblock Generative pre-training for speech with flow matching.
\newblock In \emph{The Twelfth International Conference on Learning Representations}.

\bibitem[{Liu et~al.(2023{\natexlab{a}})Liu, Zhang, Lei, Chen, Wang, Li, and Xie}]{liu2023promptstyle}
Guanghou Liu, Yongmao Zhang, Yi~Lei, Yunlin Chen, Rui Wang, Zhifei Li, and Lei Xie. 2023{\natexlab{a}}.
\newblock {PromptStyle}: Controllable style transfer for text-to-speech with natural language descriptions.
\newblock \emph{arXiv preprint arXiv:2305.19522}.

\bibitem[{Liu et~al.(2023{\natexlab{b}})Liu, Chen, Yuan, Mei, Liu, Mandic, Wang, and Plumbley}]{liu2023audioldm}
Haohe Liu, Zehua Chen, Yi~Yuan, Xinhao Mei, Xubo Liu, Danilo Mandic, Wenwu Wang, and Mark~D Plumbley. 2023{\natexlab{b}}.
\newblock {AudioLDM}: text-to-audio generation with latent diffusion models.
\newblock In \emph{Proceedings of the 40th International Conference on Machine Learning}, pages 21450--21474.

\bibitem[{Liu et~al.(2025{\natexlab{a}})Liu, Liu, Hu, Gao, Zhang, and Ling}]{liu2025diffstyletts}
Jiaxuan Liu, Zhaoci Liu, Yajun Hu, Yingying Gao, Shilei Zhang, and Zhenhua Ling. 2025{\natexlab{a}}.
\newblock {DiffStyleTTS}: Diffusion-based hierarchical prosody modeling for text-to-speech with diverse and controllable styles.
\newblock In \emph{Proceedings of the 31st International Conference on Computational Linguistics}, pages 5265--5272.

\bibitem[{Liu et~al.(2023{\natexlab{c}})Liu, Gong, and qiang liu}]{liu2023flow}
Xingchao Liu, Chengyue Gong, and qiang liu. 2023{\natexlab{c}}.
\newblock \href {https://openreview.net/forum?id=XVjTT1nw5z} {Flow straight and fast: Learning to generate and transfer data with rectified flow}.
\newblock In \emph{The Eleventh International Conference on Learning Representations}.

\bibitem[{Liu et~al.(2021)Liu, Xu, Wang, Chen, Li, Tan, Li, He, and Zhao}]{liu2021delightfultts}
Yanqing Liu, Zhihang Xu, Gang Wang, Kuan Chen, Bohan Li, Xu~Tan, Jinzhu Li, Lei He, and Sheng Zhao. 2021.
\newblock {DelightfulTTS}: The {Microsoft} speech synthesis system for blizzard challenge 2021.
\newblock \emph{arXiv preprint arXiv:2110.12612}.

\bibitem[{Liu et~al.(2022)Liu, Tian, Hu, Liu, Wu, Wang, Zhao, and Wang}]{liu2022controllable}
Zhengxi Liu, Qiao Tian, Chenxu Hu, Xudong Liu, Menglin Wu, Yuping Wang, Hang Zhao, and Yuxuan Wang. 2022.
\newblock Controllable and lossless non-autoregressive end-to-end text-to-speech.
\newblock \emph{arXiv preprint arXiv:2207.06088}.

\bibitem[{Liu et~al.(2024{\natexlab{b}})Liu, Wang, Inoue, Bai, and Li}]{liu2024autoregressive}
Zhijun Liu, Shuai Wang, Sho Inoue, Qibing Bai, and Haizhou Li. 2024{\natexlab{b}}.
\newblock Autoregressive diffusion transformer for text-to-speech synthesis.
\newblock \emph{arXiv preprint arXiv:2406.05551}.

\bibitem[{Liu et~al.(2025{\natexlab{b}})Liu, Wang, Zhu, Bi, and Li}]{liu2024e1}
Zhijun Liu, Shuai Wang, Pengcheng Zhu, Mengxiao Bi, and Haizhou Li. 2025{\natexlab{b}}.
\newblock {E1 TTS}: Simple and fast non-autoregressive {TTS}.
\newblock In \emph{IEEE International Conference on Acoustics, Speech and Signal Processing}, pages 1--5.

\bibitem[{Livingstone and Russo(2018)}]{livingstone2018ryerson}
Steven~R Livingstone and Frank~A Russo. 2018.
\newblock The ryerson audio-visual database of emotional speech and song ({RAVDESS}): A dynamic, multimodal set of facial and vocal expressions in {North} {American} {English}.
\newblock \emph{PloS One}, 13(5):e0196391.

\bibitem[{Loizou(2011)}]{loizou2011speech}
Philipos~C Loizou. 2011.
\newblock Speech quality assessment.
\newblock In \emph{Multimedia Analysis, Processing and Communications}, pages 623--654. Springer.

\bibitem[{L{\'o}pez et~al.(2018)L{\'o}pez, Quesada, and Guerrero}]{lopez2018alexa}
Gustavo L{\'o}pez, Luis Quesada, and Luis~A. Guerrero. 2018.
\newblock {Alexa vs. Siri vs. Cortana vs. Google} assistant: A comparison of speech-based natural user interfaces.
\newblock In \emph{Advances in Human Factors and Systems Interaction}, pages 241--250.

\bibitem[{Lu et~al.(2023)Lu, Wu, Wu, and Meng}]{lu2023speechtriplenet}
Hui Lu, Xixin Wu, Zhiyong Wu, and Helen Meng. 2023.
\newblock {SpeechTripleNet}: End-to-end disentangled speech representation learning for content, timbre and prosody.
\newblock In \emph{Proceedings of the 31st ACM International Conference on Multimedia}, pages 2829--2837.

\bibitem[{Lu et~al.(2022)Lu, Sisman, Liu, Zhang, and Li}]{lu2022visualtts}
Junchen Lu, Berrak Sisman, Rui Liu, Mingyang Zhang, and Haizhou Li. 2022.
\newblock {VisualTTS}: {TTS} with accurate lip-speech synchronization for automatic voice over.
\newblock In \emph{IEEE International Conference on Acoustics, Speech and Signal Processing}, pages 8032--8036.

\bibitem[{Lu et~al.(2025)Lu, Du, Sheng, Ai, and Ling}]{lu2025idea-tts}
Ye-Xin Lu, Hui-Peng Du, Zheng-Yan Sheng, Yang Ai, and Zhen-Hua Ling. 2025.
\newblock Incremental disentanglement for environment-aware zero-shot text-to-speech synthesis.
\newblock In \emph{IEEE International Conference on Acoustics, Speech and Signal Processing}, pages 1--5.

\bibitem[{Lyth and King(2024)}]{lyth2024natural}
Dan Lyth and Simon King. 2024.
\newblock Natural language guidance of high-fidelity text-to-speech with synthetic annotations.
\newblock \emph{arXiv preprint arXiv:2402.01912}.

\bibitem[{Marge et~al.(2022)Marge, Espy-Wilson, Ward, Alwan, Artzi, Bansal, Blankenship, Chai, Daum{\'e}~III, Dey et~al.}]{marge2022spoken}
Matthew Marge, Carol Espy-Wilson, Nigel~G Ward, Abeer Alwan, Yoav Artzi, Mohit Bansal, Gil Blankenship, Joyce Chai, Hal Daum{\'e}~III, Debadeepta Dey, and 1 others. 2022.
\newblock Spoken language interaction with robots: Recommendations for future research.
\newblock \emph{Computer Speech \& Language}, 71:101255.

\bibitem[{Meng et~al.(2024)Meng, Zhou, Liu, Chen, Han, Hu, Liu, Li, Zhao, Wu et~al.}]{meng2024autoregressive}
Lingwei Meng, Long Zhou, Shujie Liu, Sanyuan Chen, Bing Han, Shujie Hu, Yanqing Liu, Jinyu Li, Sheng Zhao, Xixin Wu, and 1 others. 2024.
\newblock Autoregressive speech synthesis without vector quantization.
\newblock \emph{arXiv preprint arXiv:2407.08551}.

\bibitem[{Min et~al.(2021)Min, Lee, Yang, and Hwang}]{min2021meta}
Dongchan Min, Dong~Bok Lee, Eunho Yang, and Sung~Ju Hwang. 2021.
\newblock {Meta-StyleSpeech}: Multi-speaker adaptive text-to-speech generation.
\newblock In \emph{International Conference on Machine Learning}, pages 7748--7759. PMLR.

\bibitem[{Mittag et~al.(2021)Mittag, Naderi, Chehadi, and Möller}]{mittag21_interspeech}
Gabriel Mittag, Babak Naderi, Assmaa Chehadi, and Sebastian Möller. 2021.
\newblock \href {https://doi.org/10.21437/Interspeech.2021-299} {Nisqa: A deep cnn-self-attention model for multidimensional speech quality prediction with crowdsourced datasets}.
\newblock In \emph{Interspeech 2021}, pages 2127--2131.

\bibitem[{Ning et~al.(2019)Ning, He, Wu, Xing, and Zhang}]{ning2019review}
Yishuang Ning, Sheng He, Zhiyong Wu, Chunxiao Xing, and Liang-Jie Zhang. 2019.
\newblock A review of deep learning based speech synthesis.
\newblock \emph{Applied Sciences}, 9(19):4050.

\bibitem[{Nishimura et~al.(2024)Nishimura, Hirose, Ohi, Nakayama, and Inoue}]{nishimura2024hall}
Yuto Nishimura, Takumi Hirose, Masanari Ohi, Hideki Nakayama, and Nakamasa Inoue. 2024.
\newblock {HALL-E}: Hierarchical neural codec language model for minute-long zero-shot text-to-speech synthesis.
\newblock \emph{arXiv preprint arXiv:2410.04380}.

\bibitem[{Nose et~al.(2007)Nose, Yamagishi, Masuko, and Kobayashi}]{nose2007style}
Takashi Nose, Junichi Yamagishi, Takashi Masuko, and Takao Kobayashi. 2007.
\newblock A style control technique for {HMM}-based expressive speech synthesis.
\newblock \emph{IEICE Transactions on Information and Systems}, 90(9):1406--1413.

\bibitem[{{OpenAI}(2022)}]{openai2022chatgpt}
{OpenAI}. 2022.
\newblock Introducing {ChatGPT}.
\newblock \url{https://openai.com/index/chatgpt/}.
\newblock Accessed: 2024-10-22.

\bibitem[{Park et~al.(2024{\natexlab{a}})Park, Kim, Shin, and Han}]{park2024dex}
Hyun~Joon Park, Jin~Sob Kim, Wooseok Shin, and Sung~Won Han. 2024{\natexlab{a}}.
\newblock {DEX-TTS}: Diffusion-based expressive text-to-speech with style modeling on time variability.
\newblock \emph{arXiv preprint arXiv:2406.19135}.

\bibitem[{Park et~al.(2024{\natexlab{b}})Park, Kim, Lee, Choi, Yeom, and Yoon}]{park2024nanovoice}
Nohil Park, Heeseung Kim, Che~Hyun Lee, Jooyoung Choi, Jiheum Yeom, and Sungroh Yoon. 2024{\natexlab{b}}.
\newblock {NanoVoice}: Efficient speaker-adaptive text-to-speech for multiple speakers.
\newblock \emph{arXiv preprint arXiv:2409.15760}.

\bibitem[{Peng et~al.(2024{\natexlab{a}})Peng, Wang, Xi, Li, Zhang, and Yu}]{peng2024survey}
Jing Peng, Yucheng Wang, Yu~Xi, Xu~Li, Xizhuo Zhang, and Kai Yu. 2024{\natexlab{a}}.
\newblock A survey on speech large language models.
\newblock \emph{arXiv preprint arXiv:2410.18908}.

\bibitem[{Peng et~al.(2024{\natexlab{b}})Peng, Huang, Li, Mohamed, and Harwath}]{peng2024voicecraft}
Puyuan Peng, Po-Yao Huang, Shang-Wen Li, Abdelrahman Mohamed, and David Harwath. 2024{\natexlab{b}}.
\newblock {V}oice{C}raft: Zero-shot speech editing and text-to-speech in the wild.
\newblock In \emph{Proceedings of the 62nd Annual Meeting of the Association for Computational Linguistics (Volume 1: Long Papers)}, pages 12442--12462.

\bibitem[{Prenger et~al.(2019)Prenger, Valle, and Catanzaro}]{prenger2019waveglow}
Ryan Prenger, Rafael Valle, and Bryan Catanzaro. 2019.
\newblock {WaveGlow}: A flow-based generative network for speech synthesis.
\newblock In \emph{IEEE International Conference on Acoustics, Speech and Signal Processing}, pages 3617--3621.

\bibitem[{Purcell and Munhall(2006)}]{purcell2006adaptive}
David~W Purcell and Kevin~G Munhall. 2006.
\newblock Adaptive control of vowel formant frequency: Evidence from real-time formant manipulation.
\newblock \emph{The Journal of the Acoustical Society of America}, 120(2):966--977.

\bibitem[{Rabiner(1968)}]{rabiner1968digital}
Lawrence~R Rabiner. 1968.
\newblock Digital-formant synthesizer for speech-synthesis studies.
\newblock \emph{The Journal of the Acoustical Society of America}, 43(4):822--828.

\bibitem[{Radford et~al.(2023)Radford, Kim, Xu, Brockman, McLeavey, and Sutskever}]{radford2023whisper}
Alec Radford, Jong~Wook Kim, Tao Xu, Greg Brockman, Christine McLeavey, and Ilya Sutskever. 2023.
\newblock Robust speech recognition via large-scale weak supervision.
\newblock In \emph{International Conference on Machine Learning}, pages 28492--28518.

\bibitem[{Rafailov et~al.(2023)Rafailov, Sharma, Mitchell, Manning, Ermon, and Finn}]{rafailov2023dpo}
Rafael Rafailov, Archit Sharma, Eric Mitchell, Christopher~D Manning, Stefano Ermon, and Chelsea Finn. 2023.
\newblock Direct preference optimization: Your language model is secretly a reward model.
\newblock \emph{Advances in Neural Information Processing Systems}, 36:53728--53741.

\bibitem[{Raffel et~al.(2020)Raffel, Shazeer, Roberts, Lee, Narang, Matena, Zhou, Li, and Liu}]{raffel2020t5}
Colin Raffel, Noam Shazeer, Adam Roberts, Katherine Lee, Sharan Narang, Michael Matena, Yanqi Zhou, Wei Li, and Peter~J Liu. 2020.
\newblock Exploring the limits of transfer learning with a unified text-to-text transformer.
\newblock \emph{Journal of Machine Learning Research}, 21(140):1--67.

\bibitem[{Ren et~al.(2021{\natexlab{a}})Ren, Hu, Tan, Qin, Zhao, Zhao, and Liu}]{ren2020fastspeech}
Yi~Ren, Chenxu Hu, Xu~Tan, Tao Qin, Sheng Zhao, Zhou Zhao, and Tie-Yan Liu. 2021{\natexlab{a}}.
\newblock {FastSpeech} 2: Fast and high-quality end-to-end text to speech.
\newblock In \emph{International Conference on Learning Representations}.

\bibitem[{Ren et~al.(2021{\natexlab{b}})Ren, Hu, Tan, Qin, Zhao, Zhao, and Liu}]{ren2020fastspeech2}
Yi~Ren, Chenxu Hu, Xu~Tan, Tao Qin, Sheng Zhao, Zhou Zhao, and Tie{-}Yan Liu. 2021{\natexlab{b}}.
\newblock {FastSpeech} 2: Fast and high-quality end-to-end text to speech.
\newblock In \emph{International Conference on Learning Representations}, pages 1--15.

\bibitem[{Ren et~al.(2019)Ren, Ruan, Tan, Qin, Zhao, Zhao, and Liu}]{ren2019fastspeech}
Yi~Ren, Yangjun Ruan, Xu~Tan, Tao Qin, Sheng Zhao, Zhou Zhao, and Tie-Yan Liu. 2019.
\newblock {FastSpeech}: Fast, robust and controllable text to speech.
\newblock \emph{Advances in Neural Information Processing Systems}, 32:1--10.

\bibitem[{Rezende and Mohamed(2015)}]{rezende2015variational}
Danilo Rezende and Shakir Mohamed. 2015.
\newblock Variational inference with normalizing flows.
\newblock In \emph{International Conference on Machine Learning}, pages 1530--1538.

\bibitem[{Ringeval et~al.(2013)Ringeval, Sonderegger, Sauer, and Lalanne}]{ringeval2013RECOLA}
Fabien Ringeval, Andreas Sonderegger, Juergen Sauer, and Denis Lalanne. 2013.
\newblock Introducing the {RECOLA} multimodal corpus of remote collaborative and affective interactions.
\newblock In \emph{10th IEEE International Conference and Workshops on Automatic Face and Gesture Recognition}, pages 1--8.

\bibitem[{Rix et~al.(2001)Rix, Beerends, Hollier, and Hekstra}]{Rix2001PESQ}
A.W. Rix, J.G. Beerends, M.P. Hollier, and A.P. Hekstra. 2001.
\newblock Perceptual evaluation of speech quality-a new method for speech quality assessment of telephone networks and codecs.
\newblock In \emph{2001 IEEE International Conference on Acoustics, Speech, and Signal Processing. Proceedings}, pages 749--752.

\bibitem[{Rong and Liu(2025)}]{rong2025seeing}
Yan Rong and Li~Liu. 2025.
\newblock Seeing your speech style: A novel zero-shot identity-disentanglement face-based voice conversion.
\newblock In \emph{Proceedings of the AAAI Conference on Artificial Intelligence}, volume~39, pages 25092--25100.

\bibitem[{Rong et~al.(2025)Rong, Yang, Lei, and Liu}]{rong2025dopamine}
Yan Rong, Shan Yang, Guangzhi Lei, and Li~Liu. 2025.
\newblock Dopamine audiobook: A training-free {MLLM} agent for emotional and human-like audiobook generation.
\newblock \emph{arXiv preprint arXiv:2504.11002}.

\bibitem[{Saeki et~al.(2022)Saeki, Xin, Nakata, Koriyama, Takamichi, and Saruwatari}]{saeki22c_interspeech}
Takaaki Saeki, Detai Xin, Wataru Nakata, Tomoki Koriyama, Shinnosuke Takamichi, and Hiroshi Saruwatari. 2022.
\newblock \href {https://doi.org/10.21437/Interspeech.2022-439} {Utmos: Utokyo-sarulab system for voicemos challenge 2022}.
\newblock In \emph{Interspeech 2022}, pages 4521--4525.

\bibitem[{Shen et~al.(2018)Shen, Pang, Weiss, Schuster, Jaitly, Yang, Chen, Zhang, Wang, Skerrv-Ryan et~al.}]{shen2018tacotron2}
Jonathan Shen, Ruoming Pang, Ron~J Weiss, Mike Schuster, Navdeep Jaitly, Zongheng Yang, Zhifeng Chen, Yu~Zhang, Yuxuan Wang, Rj~Skerrv-Ryan, and 1 others. 2018.
\newblock Natural {TTS} synthesis by conditioning {WaveNet} on mel spectrogram predictions.
\newblock In \emph{IEEE International Conference on Acoustics, Speech and Signal Processing}, pages 4779--4783.

\bibitem[{Shen et~al.(2024)Shen, Ju, Tan, Liu, Leng, He, Qin, sheng zhao, and Bian}]{shen2024naturalspeech2}
Kai Shen, Zeqian Ju, Xu~Tan, Eric Liu, Yichong Leng, Lei He, Tao Qin, sheng zhao, and Jiang Bian. 2024.
\newblock {NaturalSpeech} 2: Latent diffusion models are natural and zero-shot speech and singing synthesizers.
\newblock In \emph{International Conference on Learning Representations}, pages 1--25.

\bibitem[{Shi et~al.(2021)Shi, Bu, Xu, Zhang, and Li}]{shi2020aishell}
Yao Shi, Hui Bu, Xin Xu, Shaoji Zhang, and Ming Li. 2021.
\newblock {AISHELL}-3: A multi-speaker mandarin {TTS} corpus.
\newblock In \emph{Conference of the International Speech Communication Association}, pages 2756--2760.

\bibitem[{Shimizu et~al.(2024)Shimizu, Yamamoto, Kawamura, Shirahata, Doi, Komatsu, and Tachibana}]{shimizu2024prompttts++}
Reo Shimizu, Ryuichi Yamamoto, Masaya Kawamura, Yuma Shirahata, Hironori Doi, Tatsuya Komatsu, and Kentaro Tachibana. 2024.
\newblock {PromptTTS++}: Controlling speaker identity in prompt-based text-to-speech using natural language descriptions.
\newblock In \emph{IEEE International Conference on Acoustics, Speech and Signal Processing}, pages 12672--12676.

\bibitem[{Siuzdak(2024)}]{siuzdak2024vocos}
Hubert Siuzdak. 2024.
\newblock Vocos: Closing the gap between time-domain and {Fourier}-based neural vocoders for high-quality audio synthesis.
\newblock In \emph{The Twelfth International Conference on Learning Representations}.

\bibitem[{Skerry-Ryan et~al.(2018)Skerry-Ryan, Battenberg, Xiao, Wang, Stanton, Shor, Weiss, Clark, and Saurous}]{skerry2018towards}
RJ~Skerry-Ryan, Eric Battenberg, Ying Xiao, Yuxuan Wang, Daisy Stanton, Joel Shor, Ron Weiss, Rob Clark, and Rif~A Saurous. 2018.
\newblock Towards end-to-end prosody transfer for expressive speech synthesis with tacotron.
\newblock In \emph{International Conference on Machine Learning}, pages 4693--4702.

\bibitem[{Snyder et~al.(2018)Snyder, Garcia-Romero, Sell, Povey, and Khudanpur}]{Snyder2018xvectors}
David Snyder, Daniel Garcia-Romero, Gregory Sell, Daniel Povey, and Sanjeev Khudanpur. 2018.
\newblock {X-Vectors}: Robust dnn embeddings for speaker recognition.
\newblock In \emph{IEEE International Conference on Acoustics, Speech and Signal Processing}, pages 5329--5333.

\bibitem[{Song et~al.(2024)Song, Chen, Wang, Ma, and Chen}]{song2024ella}
Yakun Song, Zhuo Chen, Xiaofei Wang, Ziyang Ma, and Xie Chen. 2024.
\newblock {ELLA-V}: Stable neural codec language modeling with alignment-guided sequence reordering.
\newblock \emph{arXiv preprint arXiv:2401.07333}.

\bibitem[{Stanton et~al.(2018)Stanton, Wang, and Skerry-Ryan}]{stanton2018predicting}
Daisy Stanton, Yuxuan Wang, and RJ~Skerry-Ryan. 2018.
\newblock Predicting expressive speaking style from text in end-to-end speech synthesis.
\newblock In \emph{IEEE Spoken Language Technology Workshop}, pages 595--602. IEEE.

\bibitem[{Tabet and Boughazi(2011)}]{tabet2011speech}
Youcef Tabet and Mohamed Boughazi. 2011.
\newblock Speech synthesis techniques. a survey.
\newblock In \emph{International Workshop on Systems, Signal Processing and their Applications}, pages 67--70.

\bibitem[{Tae et~al.(2022)Tae, Kim, and Kim}]{tae2022editts}
Jaesung Tae, Hyeongju Kim, and Taesu Kim. 2022.
\newblock {EdiTTS}: Score-based editing for controllable text-to-speech.
\newblock In \emph{Annual Conference of the International Speech Communication Association}, pages 421--425.

\bibitem[{Tan et~al.(2021{\natexlab{a}})Tan, Deng, Yeung, Jiang, Chen, and Lee}]{tan2021editspeech}
Daxin Tan, Liqun Deng, Yu~Ting Yeung, Xin Jiang, Xiao Chen, and Tan Lee. 2021{\natexlab{a}}.
\newblock {EditSpeech}: A text based speech editing system using partial inference and bidirectional fusion.
\newblock In \emph{IEEE Automatic Speech Recognition and Understanding Workshop}, pages 626--633.

\bibitem[{Tan et~al.(2024)Tan, Chen, Liu, Cong, Zhang, Liu, Wang, Leng, Yi, He, Zhao, Qin, Soong, and Liu}]{tan2024naturalspeech}
Xu~Tan, Jiawei Chen, Haohe Liu, Jian Cong, Chen Zhang, Yanqing Liu, Xi~Wang, Yichong Leng, Yuanhao Yi, Lei He, Sheng Zhao, Tao Qin, Frank Soong, and Tie-Yan Liu. 2024.
\newblock {NaturalSpeech}: End-to-end text-to-speech synthesis with human-level quality.
\newblock \emph{IEEE Transactions on Pattern Analysis and Machine Intelligence}, 46(6):4234--4245.

\bibitem[{Tan et~al.(2021{\natexlab{b}})Tan, Qin, Soong, and Liu}]{tan2021survey}
Xu~Tan, Tao Qin, Frank Soong, and Tie-Yan Liu. 2021{\natexlab{b}}.
\newblock A survey on neural speech synthesis.
\newblock \emph{arXiv preprint arXiv:2106.15561}.

\bibitem[{Tokuda et~al.(2000)Tokuda, Yoshimura, Masuko, Kobayashi, and Kitamura}]{tokuda2000speech}
K.~Tokuda, T.~Yoshimura, T.~Masuko, T.~Kobayashi, and T.~Kitamura. 2000.
\newblock Speech parameter generation algorithms for {HMM}-based speech synthesis.
\newblock In \emph{IEEE International Conference on Acoustics, Speech, and Signal Processing}, volume~3, pages 1315--1318.

\bibitem[{{Toloka}(2024)}]{toloka2024CLESC}
{Toloka}. 2024.
\newblock Crowd labeled emotions and speech characteristics.
\newblock \url{https://huggingface.co/datasets/toloka/CLESC}.
\newblock Accessed: 2024-03-23.

\bibitem[{Triantafyllopoulos et~al.(2023)Triantafyllopoulos, Schuller, {\.I}ymen, Sezgin, He, Yang, Tzirakis, Liu, Mertes, Andr{\'e} et~al.}]{triantafyllopoulos2023overview_survey}
Andreas Triantafyllopoulos, Bj{\"o}rn~W Schuller, G{\"o}k{\c{c}}e {\.I}ymen, Metin Sezgin, Xiangheng He, Zijiang Yang, Panagiotis Tzirakis, Shuo Liu, Silvan Mertes, Elisabeth Andr{\'e}, and 1 others. 2023.
\newblock An overview of affective speech synthesis and conversion in the deep learning era.
\newblock \emph{Proceedings of the IEEE}, 111(10):1355--1381.

\bibitem[{Um et~al.(2020)Um, Oh, Byun, Jang, Ahn, and Kang}]{um2020emotional}
Se-Yun Um, Sangshin Oh, Kyungguen Byun, Inseon Jang, ChungHyun Ahn, and Hong-Goo Kang. 2020.
\newblock Emotional speech synthesis with rich and granularized control.
\newblock In \emph{IEEE International Conference on Acoustics, Speech and Signal Processing}, pages 7254--7258.

\bibitem[{Valle et~al.(2020)Valle, Shih, Prenger, and Catanzaro}]{valle2020flowtron}
Rafael Valle, Kevin Shih, Ryan Prenger, and Bryan Catanzaro. 2020.
\newblock Flowtron: an autoregressive flow-based generative network for text-to-speech synthesis.
\newblock \emph{arXiv preprint arXiv:2005.05957}.

\bibitem[{Van Den~Oord et~al.(2016)Van Den~Oord, Dieleman, Zen, Simonyan, Vinyals, Graves, Kalchbrenner, Senior, Kavukcuoglu et~al.}]{van2016wavenet}
Aaron Van Den~Oord, Sander Dieleman, Heiga Zen, Karen Simonyan, Oriol Vinyals, Alex Graves, Nal Kalchbrenner, Andrew Senior, Koray Kavukcuoglu, and 1 others. 2016.
\newblock {WaveNet}: A generative model for raw audio.
\newblock In \emph{9th {ISCA} Speech Synthesis Workshop}, volume~12, page 125.

\bibitem[{Vyas et~al.(2023)Vyas, Shi, Le, Tjandra, Wu, Guo, Zhang, Zhang, Adkins, Ngan et~al.}]{vyas2023audiobox}
Apoorv Vyas, Bowen Shi, Matthew Le, Andros Tjandra, Yi-Chiao Wu, Baishan Guo, Jiemin Zhang, Xinyue Zhang, Robert Adkins, William Ngan, and 1 others. 2023.
\newblock Audiobox: Unified audio generation with natural language prompts.
\newblock \emph{arXiv preprint arXiv:2312.15821}.

\bibitem[{Wang et~al.(2023{\natexlab{a}})Wang, Chen, Wu, Zhang, Zhou, Liu, Chen, Liu, Wang, Li et~al.}]{wang2023neural}
Chengyi Wang, Sanyuan Chen, Yu~Wu, Ziqiang Zhang, Long Zhou, Shujie Liu, Zhuo Chen, Yanqing Liu, Huaming Wang, Jinyu Li, and 1 others. 2023{\natexlab{a}}.
\newblock Neural codec language models are zero-shot text to speech synthesizers.
\newblock \emph{arXiv preprint arXiv:2301.02111}.

\bibitem[{Wang et~al.(2023{\natexlab{b}})Wang, Song, and Jha}]{wang2023generalizable}
Wenbin Wang, Yang Song, and Sanjay Jha. 2023{\natexlab{b}}.
\newblock Generalizable zero-shot speaker adaptive speech synthesis with disentangled representations.
\newblock In \emph{Annual Conference of the International Speech Communication Association 2023}, pages 4454--4458.

\bibitem[{Wang et~al.(2025{\natexlab{a}})Wang, Jiang, Ma, Zhang, Liu, Li, Liang, Zheng, Wang, Feng et~al.}]{wang2025sparktts}
Xinsheng Wang, Mingqi Jiang, Ziyang Ma, Ziyu Zhang, Songxiang Liu, Linqin Li, Zheng Liang, Qixi Zheng, Rui Wang, Xiaoqin Feng, and 1 others. 2025{\natexlab{a}}.
\newblock {Spark-TTS}: An efficient llm-based text-to-speech model with single-stream decoupled speech tokens.
\newblock \emph{arXiv preprint arXiv:2503.01710}.

\bibitem[{Wang et~al.(2025{\natexlab{b}})Wang, Zhan, Liu, Zeng, Guo, Zheng, Zhang, Zhang, Zhang, and Wu}]{wang2025maskgct}
Yuancheng Wang, Haoyue Zhan, Liwei Liu, Ruihong Zeng, Haotian Guo, Jiachen Zheng, Qiang Zhang, Xueyao Zhang, Shunsi Zhang, and Zhizheng Wu. 2025{\natexlab{b}}.
\newblock Mask{GCT}: Zero-shot text-to-speech with masked generative codec transformer.
\newblock In \emph{International Conference on Learning Representations}, pages 1--24.

\bibitem[{Wang et~al.(2019)Wang, Wang, Liang, and Yu}]{wang2019comic}
Yujia Wang, Wenguan Wang, Wei Liang, and Lap-Fai Yu. 2019.
\newblock Comic-guided speech synthesis.
\newblock \emph{ACM Transactions on Graphics}, 38(6):1--14.

\bibitem[{Wang et~al.(2017)Wang, Skerry{-}Ryan, Stanton, Wu, Weiss, Jaitly, Yang, Xiao, Chen, Bengio, Le, Agiomyrgiannakis, Clark, and Saurous}]{wang2017tacotron}
Yuxuan Wang, R.~J. Skerry{-}Ryan, Daisy Stanton, Yonghui Wu, Ron~J. Weiss, Navdeep Jaitly, Zongheng Yang, Ying Xiao, Zhifeng Chen, Samy Bengio, Quoc~V. Le, Yannis Agiomyrgiannakis, Rob Clark, and Rif~A. Saurous. 2017.
\newblock Tacotron: Towards end-to-end speech synthesis.
\newblock In \emph{Annual Conference of the International Speech Communication Association}, pages 4006--4010.

\bibitem[{Wang et~al.(2018)Wang, Stanton, Zhang, Ryan, Battenberg, Shor, Xiao, Jia, Ren, and Saurous}]{wang2018style}
Yuxuan Wang, Daisy Stanton, Yu~Zhang, RJ-Skerry Ryan, Eric Battenberg, Joel Shor, Ying Xiao, Ye~Jia, Fei Ren, and Rif~A Saurous. 2018.
\newblock {Style Tokens}: Unsupervised style modeling, control and transfer in end-to-end speech synthesis.
\newblock In \emph{International Conference on Machine Learning}, pages 5167--5176.

\bibitem[{Wang et~al.(2024)Wang, Wang, Li, and Huang}]{wang2024artspeech}
Zhongxu Wang, Yujia Wang, Mingzhu Li, and Hua Huang. 2024.
\newblock {ArtSpeech}: Adaptive text-to-speech synthesis with articulatory representations.
\newblock In \emph{Proceedings of the 32nd ACM International Conference on Multimedia}, pages 535--544.

\bibitem[{{Wikipedia}(2024)}]{enwiki2024wer}
{Wikipedia}. 2024.
\newblock Word error rate.
\newblock \url{https://en.wikipedia.org/wiki/Word_error_rate}.
\newblock Accessed: 2024-12-07.

\bibitem[{{Wikipedia}(2025)}]{enwiki2024mos}
{Wikipedia}. 2025.
\newblock Mean opinion score.
\newblock \url{https://en.wikipedia.org/wiki/Mean_opinion_score}.
\newblock Accessed: 2024-12-07.

\bibitem[{Woo et~al.(2023)Woo, Debnath, Hu, Chen, Liu, Kweon, and Xie}]{woo2023convnext}
Sanghyun Woo, Shoubhik Debnath, Ronghang Hu, Xinlei Chen, Zhuang Liu, In~So Kweon, and Saining Xie. 2023.
\newblock {ConvNeXt V2}: Co-designing and scaling convnets with masked autoencoders.
\newblock In \emph{Proceedings of the IEEE/CVF Conference on Computer Vision and Pattern Recognition}, pages 16133--16142.

\bibitem[{Wouters and Macon(2001)}]{wouters2001control}
Johan Wouters and Michael~W Macon. 2001.
\newblock Control of spectral dynamics in concatenative speech synthesis.
\newblock \emph{IEEE Transactions on Speech and Audio Processing}, 9(1):30--38.

\bibitem[{Xiao et~al.(2024)Xiao, Wang, Tan, He, Zhu, Zhao, and Lee}]{xiao2024contrastive}
Yujia Xiao, Xi~Wang, Xu~Tan, Lei He, Xinfa Zhu, Sheng Zhao, and Tan Lee. 2024.
\newblock Contrastive context-speech pretraining for expressive text-to-speech synthesis.
\newblock In \emph{Proceedings of the 32nd ACM International Conference on Multimedia}, pages 2099--2107.

\bibitem[{Xin et~al.(2024)Xin, Tan, Shen, Ju, Yang, Wang, Takamichi, Saruwatari, Liu, Li et~al.}]{xin2024rall}
Detai Xin, Xu~Tan, Kai Shen, Zeqian Ju, Dongchao Yang, Yuancheng Wang, Shinnosuke Takamichi, Hiroshi Saruwatari, Shujie Liu, Jinyu Li, and 1 others. 2024.
\newblock {RALL-E}: Robust codec language modeling with chain-of-thought prompting for text-to-speech synthesis.
\newblock \emph{arXiv preprint arXiv:2404.03204}.

\bibitem[{Yamagishi et~al.(2009)Yamagishi, Nose, Zen, Ling, Toda, Tokuda, King, and Renals}]{yamagishi2009robust}
Junichi Yamagishi, Takashi Nose, Heiga Zen, Zhen-Hua Ling, Tomoki Toda, Keiichi Tokuda, Simon King, and Steve Renals. 2009.
\newblock Robust speaker-adaptive {HMM}-based text-to-speech synthesis.
\newblock \emph{IEEE Transactions on Audio, Speech, and Language Processing}, 17(6):1208--1230.

\bibitem[{Yamagishi et~al.(2003)Yamagishi, Onishi, Masuko, and Kobayashi}]{yamagishi2003modeling}
Junichi Yamagishi, Koji Onishi, Takashi Masuko, and Takao Kobayashi. 2003.
\newblock Modeling of various speaking styles and emotions for {HMM}-based speech synthesis.
\newblock In \emph{Proceedings of the Annual Conference of the International Speech Communication Association}, pages 2461--2464.

\bibitem[{Yamamoto et~al.(2024)Yamamoto, Shirahata, Kawamura, and Tachibana}]{yamamoto2024description}
Ryuichi Yamamoto, Yuma Shirahata, Masaya Kawamura, and Kentaro Tachibana. 2024.
\newblock Description-based controllable text-to-speech with cross-lingual voice control.
\newblock \emph{arXiv preprint arXiv:2409.17452}.

\bibitem[{Yamamoto et~al.(2020)Yamamoto, Song, and Kim}]{yamamoto2020parallelwavegan}
Ryuichi Yamamoto, Eunwoo Song, and Jae-Min Kim. 2020.
\newblock Parallel {WaveGAN}: A fast waveform generation model based on generative adversarial networks with multi-resolution spectrogram.
\newblock In \emph{IEEE International Conference on Acoustics, Speech and Signal Processing}, pages 6199--6203.

\bibitem[{Yang et~al.(2024{\natexlab{a}})Yang, Huang, Wang, Guo, Chong, Liu, Wu, and Meng}]{yang2024simplespeech2}
Dongchao Yang, Rongjie Huang, Yuanyuan Wang, Haohan Guo, Dading Chong, Songxiang Liu, Xixin Wu, and Helen Meng. 2024{\natexlab{a}}.
\newblock {SimpleSpeech} 2: Towards simple and efficient text-to-speech with flow-based scalar latent transformer diffusion models.
\newblock \emph{arXiv preprint arXiv:2408.13893}.

\bibitem[{Yang et~al.(2023{\natexlab{a}})Yang, Liu, Huang, Tian, Weng, and Zou}]{yang2023hificodec}
Dongchao Yang, Songxiang Liu, Rongjie Huang, Jinchuan Tian, Chao Weng, and Yuexian Zou. 2023{\natexlab{a}}.
\newblock {HiFi-Codec}: Group-residual vector quantization for high fidelity audio codec.
\newblock \emph{arXiv preprint arXiv:2305.02765}.

\bibitem[{Yang et~al.(2024{\natexlab{b}})Yang, Liu, Huang, Weng, and Meng}]{yang2024instructtts}
Dongchao Yang, Songxiang Liu, Rongjie Huang, Chao Weng, and Helen Meng. 2024{\natexlab{b}}.
\newblock {InstructTTS}: Modelling expressive {TTS} in discrete latent space with natural language style prompt.
\newblock \emph{IEEE/ACM Transactions on Audio, Speech, and Language Processing}, 32:2913--2925.

\bibitem[{Yang et~al.(2023{\natexlab{b}})Yang, Tian, Tan, Huang, Liu, Chang, Shi, Zhao, Bian, Wu et~al.}]{yang2023uniaudio}
Dongchao Yang, Jinchuan Tian, Xu~Tan, Rongjie Huang, Songxiang Liu, Xuankai Chang, Jiatong Shi, Sheng Zhao, Jiang Bian, Xixin Wu, and 1 others. 2023{\natexlab{b}}.
\newblock {UniAudio}: An audio foundation model toward universal audio generation.
\newblock \emph{arXiv preprint arXiv:2310.00704}.

\bibitem[{Yang et~al.(2024{\natexlab{c}})Yang, Wang, Guo, Chen, Wu, and Meng}]{yang2024simplespeech}
Dongchao Yang, Dingdong Wang, Haohan Guo, Xueyuan Chen, Xixin Wu, and Helen Meng. 2024{\natexlab{c}}.
\newblock {SimpleSpeech}: Towards simple and efficient text-to-speech with scalar latent transformer diffusion models.
\newblock In \emph{Proceedings of the Annual Conference of the International Speech Communication Association}, pages 4398--4402.

\bibitem[{Yang et~al.(2025)Yang, Yang, Chen, Ma, Chen, Wang, Wang, Yang, Niu, Liu et~al.}]{yang2025emovoice}
Guanrou Yang, Chen Yang, Qian Chen, Ziyang Ma, Wenxi Chen, Wen Wang, Tianrui Wang, Yifan Yang, Zhikang Niu, Wenrui Liu, and 1 others. 2025.
\newblock {EmoVoice}: {LLM}-based emotional text-to-speech model with freestyle text prompting.
\newblock \emph{arXiv preprint arXiv:2504.12867}.

\bibitem[{Yang et~al.(2021)Yang, Bae, Bak, Kim, and Cho}]{yang2021ganspeech}
Jinhyeok Yang, Jae{-}Sung Bae, Taejun Bak, Young{-}Ik Kim, and Hoon{-}Young Cho. 2021.
\newblock {GANSpeech}: Adversarial training for high-fidelity multi-speaker speech synthesis.
\newblock In \emph{22nd Annual Conference of the International Speech Communication Association}, pages 2202--2206.

\bibitem[{Yang et~al.(2024{\natexlab{d}})Yang, Zuo, Su, Jiang, Li, Zhao, Chen, Wang, and Huai}]{yang2024mscenespeech}
Qian Yang, Jialong Zuo, Zhe Su, Ziyue Jiang, Mingze Li, Zhou Zhao, Feiyang Chen, Zhefeng Wang, and Baoxing Huai. 2024{\natexlab{d}}.
\newblock {MSceneSpeech}: A multi-scene speech dataset for expressive speech synthesis.
\newblock In \emph{Annual Conference of the International Speech Communication Association 2024}, pages 1845--1849.

\bibitem[{Yang et~al.(2024{\natexlab{e}})Yang, Ma, Liu, Li, Wang, Meng, Sun, Liang, Xu, Hu et~al.}]{yang2024ist-lm}
Yifan Yang, Ziyang Ma, Shujie Liu, Jinyu Li, Hui Wang, Lingwei Meng, Haiyang Sun, Yuzhe Liang, Ruiyang Xu, Yuxuan Hu, and 1 others. 2024{\natexlab{e}}.
\newblock Interleaved speech-text language models are simple streaming text to speech synthesizers.
\newblock \emph{arXiv preprint arXiv:2412.16102}.

\bibitem[{Yang et~al.(2022)Yang, Chen, Luo, Yang, Ye, Cheng, Xu, Jin, Zhang, Zhang et~al.}]{yang2022magicdata}
Zehui Yang, Yifan Chen, Lei Luo, Runyan Yang, Lingxuan Ye, Gaofeng Cheng, Ji~Xu, Yaohui Jin, Qingqing Zhang, Pengyuan Zhang, and 1 others. 2022.
\newblock Open {Source} {MagicData-RAMC}: A rich annotated mandarin conversational ({RAMC}) speech dataset.
\newblock \emph{arXiv preprint arXiv:2203.16844}.

\bibitem[{Ye et~al.(2024)Ye, Ju, Liu, Tan, Chen, Lu, Sun, Pan, Bian, He et~al.}]{ye2024flashspeech}
Zhen Ye, Zeqian Ju, Haohe Liu, Xu~Tan, Jianyi Chen, Yiwen Lu, Peiwen Sun, Jiahao Pan, Weizhen Bian, Shulin He, and 1 others. 2024.
\newblock {FlashSpeech}: Efficient zero-shot speech synthesis.
\newblock In \emph{Proceedings of the 32nd ACM International Conference on Multimedia}, pages 6998--7007.

\bibitem[{Yoshimura et~al.(1999)Yoshimura, Tokuda, Masuko, Kobayashi, and Kitamura}]{yoshimura1999simultaneous}
Takayoshi Yoshimura, Keiichi Tokuda, Takashi Masuko, Takao Kobayashi, and Tadashi Kitamura. 1999.
\newblock Simultaneous modeling of spectrum, pitch and duration in {HMM}-based speech synthesis.
\newblock In \emph{6th European Conference on Speech Communication and Technology}, pages 2347--2350.

\bibitem[{Yu et~al.(2020)Yu, Lu, Hu, Yu, Weng, Xu, Liu, Tuo, Kang, Lei et~al.}]{yu2020durian}
Chengzhu Yu, Heng Lu, Na~Hu, Meng Yu, Chao Weng, Kun Xu, Peng Liu, Deyi Tuo, Shiyin Kang, Guangzhi Lei, and 1 others. 2020.
\newblock {DurIAN}: Duration informed attention network for speech synthesis.
\newblock In \emph{Proceedings of the Annual Conference of the International Speech Communication Association}, pages 2027--2031.

\bibitem[{Zeghidour et~al.(2021)Zeghidour, Luebs, Omran, Skoglund, and Tagliasacchi}]{zeghidour2021soundstream}
Neil Zeghidour, Alejandro Luebs, Ahmed Omran, Jan Skoglund, and Marco Tagliasacchi. 2021.
\newblock {SoundStream}: An end-to-end neural audio codec.
\newblock \emph{IEEE/ACM Transactions on Audio, Speech, and Language Processing}, 30:495--507.

\bibitem[{Zen et~al.(2013)Zen, Senior, and Schuster}]{zen2013statistical}
Heiga Zen, Andrew Senior, and Mike Schuster. 2013.
\newblock Statistical parametric speech synthesis using deep neural networks.
\newblock In \emph{IEEE International Conference on Acoustics, Speech and Signal Processing}, pages 7962--7966.

\bibitem[{Zen et~al.(2009)Zen, Tokuda, and Black}]{zen2009statistical}
Heiga Zen, Keiichi Tokuda, and Alan~W Black. 2009.
\newblock Statistical parametric speech synthesis.
\newblock \emph{Speech Communication}, 51(11):1039--1064.

\bibitem[{Zhang et~al.(2022)Zhang, Lv, Guo, Shao, Yang, Xie, Xu, Bu, Chen, Zeng et~al.}]{zhang2022wenetspeech}
Binbin Zhang, Hang Lv, Pengcheng Guo, Qijie Shao, Chao Yang, Lei Xie, Xin Xu, Hui Bu, Xiaoyu Chen, Chenchen Zeng, and 1 others. 2022.
\newblock {WenetSpeech}: A 10000+ hours multi-domain mandarin corpus for speech recognition.
\newblock In \emph{IEEE International Conference on Acoustics, Speech and Signal Processing}, pages 6182--6186.

\bibitem[{Zhang et~al.(2023{\natexlab{a}})Zhang, Zhang, Zheng, Zhang, Qamar, Bae, and Kweon}]{zhang2023survey}
Chenshuang Zhang, Chaoning Zhang, Sheng Zheng, Mengchun Zhang, Maryam Qamar, Sung-Ho Bae, and In~So Kweon. 2023{\natexlab{a}}.
\newblock A survey on audio diffusion models: Text to speech synthesis and enhancement in generative {AI}.
\newblock \emph{arXiv preprint arXiv:2303.13336}.

\bibitem[{Zhang et~al.(2023{\natexlab{b}})Zhang, Li, Zhang, Zhan, Wang, Zhou, and Qiu}]{zhang2023speechgpt}
Dong Zhang, Shimin Li, Xin Zhang, Jun Zhan, Pengyu Wang, Yaqian Zhou, and Xipeng Qiu. 2023{\natexlab{b}}.
\newblock {S}peech{GPT}: Empowering large language models with intrinsic cross-modal conversational abilities.
\newblock In \emph{Findings of the Association for Computational Linguistics: EMNLP}, pages 15757--15773.

\bibitem[{Zhang et~al.(2025{\natexlab{a}})Zhang, Mehrish, Li, and Poria}]{zhang2025proemo}
Shaozuo Zhang, Ambuj Mehrish, Yingting Li, and Soujanya Poria. 2025{\natexlab{a}}.
\newblock {PROEMO}: Prompt-driven text-to-speech synthesis based on emotion and intensity control.
\newblock \emph{arXiv preprint arXiv:2501.06276}.

\bibitem[{Zhang et~al.(2024)Zhang, Zhang, Li, Zhou, and Qiu}]{zhang2023speechtokenizer}
Xin Zhang, Dong Zhang, Shimin Li, Yaqian Zhou, and Xipeng Qiu. 2024.
\newblock {SpeechTokenizer}: Unified speech tokenizer for speech language models.
\newblock In \emph{The Twelfth International Conference on Learning Representations}.

\bibitem[{Zhang et~al.(2025{\natexlab{b}})Zhang, Zhang, Peng, Tang, Manohar, Liu, Hwang, Li, Wang, Chan, Huang, Wu, and Ma}]{zhang2025vevo}
Xueyao Zhang, Xiaohui Zhang, Kainan Peng, Zhenyu Tang, Vimal Manohar, Yingru Liu, Jeff Hwang, Dangna Li, Yuhao Wang, Julian Chan, Yuan Huang, Zhizheng Wu, and Mingbo Ma. 2025{\natexlab{b}}.
\newblock Vevo: Controllable zero-shot voice imitation with self-supervised disentanglement.
\newblock In \emph{The Thirteenth International Conference on Learning Representations}.

\bibitem[{Zhang et~al.(2019)Zhang, Pan, He, and Ling}]{zhang2019learning}
Ya-Jie Zhang, Shifeng Pan, Lei He, and Zhen-Hua Ling. 2019.
\newblock Learning latent representations for style control and transfer in end-to-end speech synthesis.
\newblock In \emph{IEEE International Conference on Acoustics, Speech and Signal Processing}, pages 6945--6949.

\bibitem[{Zhang et~al.(2023{\natexlab{c}})Zhang, Liu, Lei, Chen, Yin, Xie, and Li}]{zhang2023promptspeaker}
Yongmao Zhang, Guanghou Liu, Yi~Lei, Yunlin Chen, Hao Yin, Lei Xie, and Zhifei Li. 2023{\natexlab{c}}.
\newblock Promptspeaker: Speaker generation based on text descriptions.
\newblock In \emph{IEEE Automatic Speech Recognition and Understanding Workshop}, pages 1--7. IEEE.

\bibitem[{Zhang et~al.(2023{\natexlab{d}})Zhang, Zhou, Wang, Chen, Wu, Liu, Chen, Liu, Wang, Li et~al.}]{zhang2023speak}
Ziqiang Zhang, Long Zhou, Chengyi Wang, Sanyuan Chen, Yu~Wu, Shujie Liu, Zhuo Chen, Yanqing Liu, Huaming Wang, Jinyu Li, and 1 others. 2023{\natexlab{d}}.
\newblock Speak foreign languages with your own voice: Cross-lingual neural codec language modeling.
\newblock \emph{arXiv preprint arXiv:2303.03926}.

\bibitem[{Zhao et~al.(2023)Zhao, Zhou, Li, Tang, Wang, Hou, Min, Zhang, Zhang, Dong et~al.}]{zhao2023survey}
Wayne~Xin Zhao, Kun Zhou, Junyi Li, Tianyi Tang, Xiaolei Wang, Yupeng Hou, Yingqian Min, Beichen Zhang, Junjie Zhang, Zican Dong, and 1 others. 2023.
\newblock A survey of large language models.
\newblock \emph{arXiv preprint arXiv:2303.18223}, 1(2).

\bibitem[{Zhou et~al.(2022)Zhou, Sisman, Liu, and Li}]{zhou2022emotional}
Kun Zhou, Berrak Sisman, Rui Liu, and Haizhou Li. 2022.
\newblock Emotional voice conversion: Theory, databases and {ESD}.
\newblock \emph{Speech Communication}, 137:1--18.

\bibitem[{Zhou et~al.(2024)Zhou, Qin, Jin, Zhou, Lei, Zhou, Wu, and Jia}]{zhou2024voxinstruct}
Yixuan Zhou, Xiaoyu Qin, Zeyu Jin, Shuoyi Zhou, Shun Lei, Songtao Zhou, Zhiyong Wu, and Jia Jia. 2024.
\newblock {VoxInstruct}: Expressive human instruction-to-speech generation with unified multilingual codec language modelling.
\newblock In \emph{Proceedings of the 32nd ACM International Conference on Multimedia}, pages 554--563.

\bibitem[{Zhu et~al.(2024)Zhu, Tian, and Xie}]{zhu2024kall-e}
Xinfa Zhu, Wenjie Tian, and Lei Xie. 2024.
\newblock Autoregressive speech synthesis with next-distribution prediction.
\newblock \emph{arXiv preprint arXiv:2412.16846}.

\end{thebibliography}

\appendix

\begin{table*}[!t]
    \centering
    \resizebox{1.0\textwidth}{!}{
    \begin{tabular}{p{3cm}p{4cm}p{4cm}p{8cm}}
        \toprule
        Control Strategy & Core Idea & Key Features & Pros \& Cons \\
        \midrule
        Style Tagging & Control specific attributes using predefined tags. & Direct control over attributes like emotion and pitch. & \textbf{Pros}: Simple to implement. \textbf{Cons}: Limited expressive diversity, cannot achieve fine-grained or combined control. \\
        \midrule
        Reference Speech Prompts & Use a short audio clip as a reference for style. & Enables zero-shot TTS and voice cloning; separates timbre and style. & \textbf{Pros}: High degree of personalization, more flexible control. \textbf{Cons}: Relies on high-quality reference audio, less intuitive control dimensions. \\
        \midrule
        Natural Language Descriptions & Describe desired voice characteristics in text. & User-friendly control via natural language (e.g., "speak calmly"). & \textbf{Pros}: High interpretability, user-friendly. \textbf{Cons}: Limited freedom and accuracy of description, potential for model misunderstanding. \\
        \midrule
        Instruction-Guided Control & Use LLMs to interpret free-form instructions. & Combines content and style instructions; can generate paralinguistic sounds. & \textbf{Pros}: Extremely high control precision and freedom, understands complex instructions. \textbf{Cons}: Strong dependency on LLMs, high system complexity. \\ 
        \bottomrule
    \end{tabular}
    }
    \caption{Summary of the pros and cons of each control strategy.}
    \label{tab:appd_control_strategies}
\end{table*}

\begin{table*}[!t]
\centering
\resizebox{1.0\textwidth}{!}{
\begin{threeparttable}
    \begin{tabular}{cccccccccccccccc}
    \toprule
    \multicolumn{1}{c}{\multirow{2}{*}[-3pt]{Dataset}} & \multicolumn{1}{c}{\multirow{2}{*}[-3pt]{\makecell{Hours\\(at least)}}} & \multicolumn{1}{c}{\multirow{2}{*}[-3pt]{\makecell{\#Speakers\\(at least)}}} & \multicolumn{11}{c}{Labels} & \multicolumn{1}{c}{\multirow{2}{*}[-3pt]{\makecell{Lang}}} & \multicolumn{1}{c}{\multirow{2}{*}[-3pt]{\makecell{Release\\Time}}} \\ 
    \cmidrule(rl){4-14}
    & & & Pit. & Ene. & Spe. & Age & Gen. & Emo. & Emp. & Acc. & Top. & Des. & Dia. & & \\
    \midrule
    IEMOCAP~\cite{busso2008iemocap} & 12 & 10 & \cmark & \cmark & \cmark & & \cmark & \cmark &  &  &  &  &  & en & 2008 \\
    RECOLA~\cite{ringeval2013RECOLA} &3.8 &46 & &&&&& \cmark  & &&&&& fr & 2013 \\
    RAVDESS~\cite{livingstone2018ryerson} & / & 24 &  & &  & \cmark &  & \cmark &  &  & & & & en & 2018 \\
    CMU-MOSEI~\cite{bagher2018cmu-mosei} & 65 & 1,000 & &&&& & \cmark & &&&&& en & 2018 \\
    Taskmaster-1~\cite{byrne2019taskmaster}& / & / &  &  &  &  &  &  &  &  &  &  &  \cmark & en&2019 \\
    AISHELL-3~\cite{shi2020aishell} & 85 & 218 & & & & \cmark & \cmark & & & \cmark & & & & zh & 2020 \\
    Common Voice~\cite{ardila2020commonvoice} & 2,500 & 50,000 &  &  &  & \cmark & \cmark &  &  & \cmark & &  &  & multi & 2020 \\
    ESD~\cite{zhou2022emotional} &29&10&  &  &  & &  & \cmark &  &  &  &  &  & en,zh &2021 \\
    GigaSpeech~\cite{chen2021gigaspeech} & 10,000 & / & &  &  &  &  &  &  &  & \cmark &  &  & en & 2021 \\
    WenetSpeech~\cite{zhang2022wenetspeech}& 10,000 & / &  &  &  &  &  &  &  & &\cmark & & & zh & 2021 \\
    PromptSpeech~\cite{guo2023prompttts}& / & / & \cmark & \cmark & \cmark & &  \cmark& &  &  &  & \cmark &  & en & 2022 \\
    MagicData-RAMC~\cite{yang2022magicdata} & 180 & 663 &  &  &  & &  & &  &  & \cmark &  & \cmark & zh & 2022 \\
    DailyTalk~\cite{lee2023dailytalk}& 20 & 2&  &  &  &  &  & \cmark &  &  & \cmark &  & \cmark & en&2023 \\
    TextrolSpeech~\cite{ji2024textrolspeech} & 330 & 1,324 & \cmark & \cmark & \cmark &  & \cmark & \cmark &  &  &  & \cmark &  & en  & 2023\\
    CLESC~\cite{toloka2024CLESC} & $<$1 & / & \cmark & \cmark & \cmark & & & \cmark & & & & & & en & 2024 \\
    VccmDataset~\cite{ji2024controlspeech}& 330& 1,324& \cmark & \cmark & \cmark &  & \cmark & \cmark &  &  &  & \cmark &  & en&2024\\
    MSceneSpeech~\cite{yang2024mscenespeech} & 13 & 13 &  &  &  &  & &  &  &  & \cmark &  &  & zh & 2024 \\
    Parler-TTS~\cite{lyth2024natural} & 50,000 & / & \cmark &  & \cmark &  & \cmark & \cmark &  & \cmark &  & \cmark &  & en & 2024\\
    SpeechCraft~\cite{jin2024speechcraft} & 2,391 & 3,200 & \cmark & \cmark & \cmark & \cmark & \cmark & \cmark & \cmark & & \cmark & \cmark & & en,zh & 2024 \\
    \bottomrule
    \end{tabular}
    Abbreviations: Pit(ch), Ene(rgy)=volume, Spe(ed), Gen(der), Emo(tion), Emp(hasis), Acc(ent), Top(ic), Des(cription), Dia(logue).
\end{threeparttable}
}
\caption{A summary of publicly available speech datasets for controllable TTS.}
\label{tab:datasets}
\end{table*}

\begin{table}[!t]
\setlength{\tabcolsep}{1pt}
\centering
\resizebox{1.0\columnwidth}{!}{
\begin{threeparttable}
    \begin{tabular}{cccc}
    \toprule
    Metric & Type & Eval Target & GT Required \\ 
    \midrule
    MCD~\cite{kominek2008synthesizer}$\downarrow$ & Objective & Acoustic similarity & \cmark \\
    FDSD~\cite{Bińkowski2020High}$\downarrow$ & Objective & Acoustic similarity & \cmark \\
    WER~\cite{enwiki2024wer}$\downarrow$ & Objective & Intelligibility & \cmark \\
    Cosine~\cite{Brecht20ECAPA-TDNN}$\downarrow$ & Objective & Speaker similarity & \cmark \\
    PESQ~\cite{Rix2001PESQ}$\uparrow$ & Objective & Perceptual quality & \cmark \\
    \midrule
    MOS~\cite{enwiki2024mos}$\uparrow$ & Subjective & Preference & \\
    CMOS~\cite{loizou2011speech}$\uparrow$ & Subjective & Preference & \\
    AB Test & Subjective & Preference & \\
    ABX Test & Subjective & Perceptual similarity & \cmark \\
    \bottomrule
    \end{tabular}
    GT: Ground truth, $\downarrow$: Lower is better, $\uparrow$: Higher is better.
\end{threeparttable}}
\caption{Widely used evaluation metrics.}
\label{tab:eval_metrics}
\end{table}

\section{Appendix} \label{sec:appendix}

\subsection{The History of Controllable TTS}
\label{sec:appd_history}

Controllable TTS aims to steer various aspects of synthesized speech, including pitch, energy, speed, prosody, timbre, emotion, gender, and speaking style.
This subsection briefly reviews its development, from early methods to recent advancements.

\noindent\textbf{Early Methods.}
Early controllable TTS systems were primarily based on rule-based, concatenative, and statistical approaches.
Rule-based systems, such as formant synthesis~\cite{rabiner1968digital,allen1987mitalk,purcell2006adaptive}, used handcrafted rules to adjust acoustic parameters like pitch and duration, enabling basic prosody control.
Concatenative systems~\cite{Hunt1996373,wouters2001control,bulut2002expressive} generated speech by stitching pre-recorded speech units together, allowing prosody modifications through pitch and timing adjustments.
Later, Hidden Markov model (HMM)-based statistical methods~\cite{nose2007style,ling2009integrating,tokuda2000speech} modeled the relationship between linguistic features and acoustic outputs, offering greater flexibility in controlling prosody and speaking rate.
These systems also introduced speaker adaptation~\cite{yamagishi2009robust} and limited emotional control~\cite{yamagishi2003modeling}, and require less storage and provide smoother transitions than concatenative methods.

\noindent\textbf{Neural Synthesis.}
The emergence of deep learning revolutionized TTS, leading to the development of neural model-based systems capable of producing more natural, expressive, and controllable speech.
Early models like WaveNet~\cite{van2016wavenet} and Tacotron~\cite{wang2017tacotron} demonstrated the potential for prosody control through explicit conditioning~\cite{shen2018tacotron2,ren2020fastspeech2}.
Neural TTS further enhanced speaker control through speaker embeddings and adaptation techniques~\cite{fan2015multi,casanova2022yourtts}, while advances in emotional modeling~\cite{lei2022msemotts,um2020emotional} enabled the synthesis of speech with specific affective tones.
Recent models have also achieved manipulation of timbre~\cite{wang2025maskgct,shen2024naturalspeech2} and style~\cite{li2025styletts,huang2022generspeech}, fostering the research in zero-shot TTS and voice cloning~\cite{cooper2020zero}.
In addition, methods for fine-grained content control~\cite{peng2024voicecraft,tan2021editspeech} have made it possible to emphasize or edit specific words in synthesized speech.

\noindent\textbf{LLM-based Synthesis.}
More recently, LLM-based approaches have further advanced controllable TTS.
Leveraging models like BERT~\cite{devlin2018bert}, GPT~\cite{brown2020gpt3}, T5~\cite{raffel2020t5}, and PaLM~\cite{chowdhery2023palm}, LLMs bring superior context modeling and intuitive control to speech synthesis~\cite{guo2023prompttts,zhou2024voxinstruct}.
By interpreting natural language prompts, such as describing a speaker's emotion, age, or style, LLMs can infer nuanced attributes and steer the generation process accordingly.
This enables dynamic, fine-grained control over prosody, emotion, style, and speaker identity~\cite{yang2024instructtts,gao2024emo}, marking a big step toward more flexible and intuitive TTS systems.

\begin{figure*}[!tb]
    \centering
    \includegraphics[width=\linewidth]{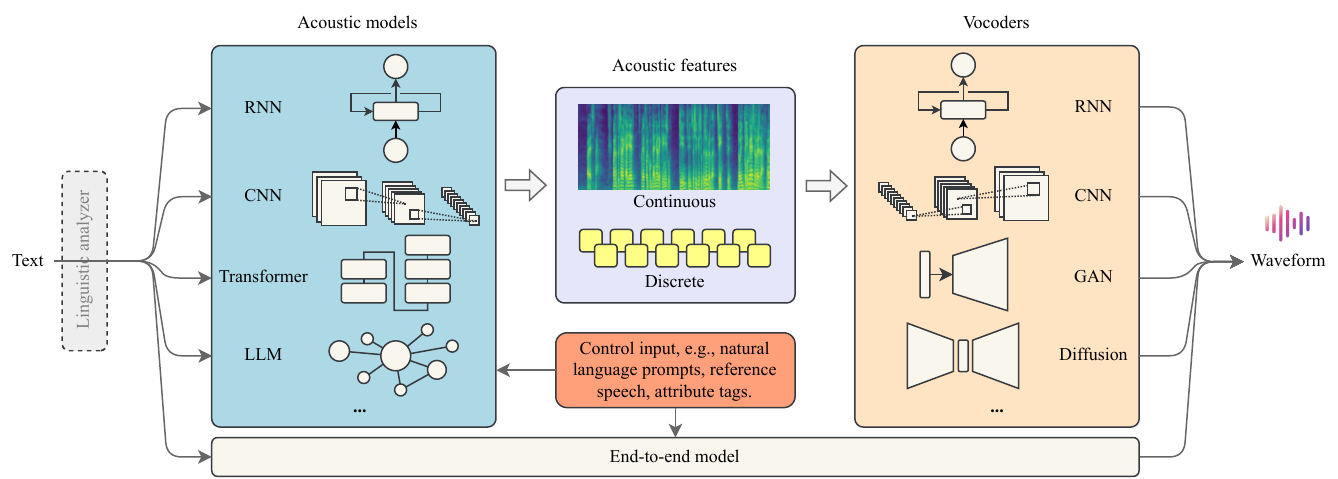}
    \caption{General pipeline of controllable TTS from the perspective of network structure. Linguistic analysis is necessary for parametric and a few neural methods but is no longer needed for most modern neural methods. In this paper, we only review neural model-based controllable TTS methods and do not investigate acoustic features (e.g., MFCC~\cite{fukada1992adaptive}, LSP~\cite{itakura1975line}, F0~\cite{kawahara1999restructuring}) used in early TTS methods.}
    \label{fig:pipeline}
\end{figure*}

\begin{table*}[!t]
\centering
\resizebox{1.0\textwidth}{!}{
\begin{threeparttable}
    \begin{tabular}{cccc}
        \toprule
        Method & Modeling & Code & Year \\
        \midrule
        VQ-Wav2Vec~\cite{baevski2019vqwav2vec} & SSCP & \url{https://github.com/facebookresearch/fairseq/tree/main/examples/wav2vec\#vq-wav2vec} & 2019 \\
        Wav2Vec 2.0~\cite{baevski2020wav2vec} & SSCP & \url{https://github.com/facebookresearch/fairseq/tree/main/examples/wav2vec} & 2019 \\
        HuBERT~\cite{hsu2021hubert} & SSCP & \url{https://github.com/facebookresearch/fairseq/tree/main/examples/hubert} & 2021 \\
        Whisper Encoder~\cite{radford2023whisper} & SSCP & \url{https://github.com/openai/whisper} & 2022 \\
        Data2vec~\cite{baevski2022data2vec} & SSCP & \url{https://github.com/facebookresearch/fairseq/tree/main/examples/data2vec} & 2022 \\
        W2v-BERT 2.0~\cite{barrault2023seamless} & SSCP & \url{https://huggingface.co/facebook/w2v-bert-2.0} & 2023 \\
        \midrule
        SoundStream~\cite{zeghidour2021soundstream} & RVQ-GAN & \url{https://github.com/wesbz/SoundStream} & 2021 \\
        Encodec~\cite{defossez2022encodec} & RVQ-GAN & \url{https://github.com/facebookresearch/encodec} & 2022 \\
        HiFi-Codec~\cite{yang2023hificodec} & RVQ-GAN & \url{https://github.com/yangdongchao/AcademiCodec} & 2023 \\
        SpeechTokenizer~\cite{zhang2023speechtokenizer} & RVQ-GAN & \url{https://github.com/ZhangXInFD/SpeechTokenizer} & 2023 \\
        Descript Audio Codec~\cite{kumar2024dac} & RVQ-GAN & \url{https://github.com/descriptinc/descript-audio-codec} & 2023 \\
        Mimi Codec~\cite{defossez2024moshi} & RVQ-GAN & \url{https://github.com/kyutai-labs/moshi} & 2024 \\
        WavTokenizer~\cite{ji2025wavtokenizer} & VQ-GAN & \url{https://github.com/jishengpeng/WavTokenizer} & 2024 \\
        \bottomrule
    \end{tabular}
    SSCP: Self-supervised context (token) prediction, RVQ: Residual vector quantization~\cite{zeghidour2021soundstream}.
\end{threeparttable}}
\caption{Popular open-source speech quantization and tokenization methods.}
\label{tab:speech_feature}
\end{table*}

\subsection{Overview of the TTS Pipeline}
\label{sec:appd_pipeline}

In this section, we provide an overview of the general pipeline that supports controllable TTS technologies.
Fig.~\ref{fig:pipeline} depicts the general pipeline of controllable TTS, containing various model architectures and feature representations.

A TTS pipeline generally contains three key components, i.e., linguistic analyzer, acoustic model, and speech vocoder, where a conditional input, e.g., prompts, can be processed for controllable speech synthesis.
\emph{Linguistic analyzer} aims to extract linguistic features, e.g., phoneme duration and position, syllable stress, and utterance level, from the input text, which is a necessary step in HMM-based methods~\cite{yoshimura1999simultaneous,tokuda2000speech} and a few neural model-based methods~\cite{zen2013statistical,fan2014tts}, but is time-consuming and error-prone.
\emph{Acoustic model} is a parametric or neural model that predicts the acoustic features from the input texts.
Modern neural acoustic models like Tacotron~\cite{wang2017tacotron} and later works~\cite{ren2019fastspeech,ren2020fastspeech2,jeong2021difftts} directly take character~\cite{chen2015joint} or word embeddings~\cite{almeida2019word} as input, which is much more efficient than previous methods.
\emph{Speech vocoder} is the last component that converts the intermediate acoustic features into a waveform that can be played back.
This step bridges the gap between the acoustic features and the actual sounds produced, helping to generate high-quality, natural-sounding speech~\cite{van2016wavenet,kong2020hifigan}.
Besides, some end-to-end methods use a single model to encode the input and decode the speech waveforms without generating intermediate features like mel-spectrograms.
One can refer to ~\citet{tan2021survey} for a more comprehensive and detailed review of acoustic models and vocoders.

\subsection{Speech Quantization vs. Tokenization.}
\label{sec:appd_feature}

It is worth noting that quantization and tokenization serve distinct purposes in speech processing.
Quantization is primarily used for high-fidelity compression, reducing the precision of numerical representations (e.g., from 32-bit floating point to 8-bit integers) while preserving model performance.
In speech synthesis, quantization is often used in waveform generation (e.g., codec-based approaches like EnCodec~\cite{defossez2022encodec}) and neural vocoders to compress audio signals without significant loss of perceptual quality.
Tokenization, on the other hand, is a discretization process that segments continuous data into meaningful units.
In speech tasks, tokenization extracts semantically relevant representations such as phonemes, characters, or learned speech units (e.g., HuBERT~\cite{hsu2021hubert} and Wav2Vec 2.0~\cite{baevski2020wav2vec}).
This makes tokenization particularly suitable for speech-to-text (ASR), TTS, and multimodal NLP tasks, where aligning speech with textual information is crucial.
Tokenization also facilitates training language models on speech data by enabling linguistic or learned unit-based processing rather than raw audio waveform modeling.
Table \ref{tab:speech_feature} in Appendix~\ref{sec:appd_feature} summarizes popular open-source speech quantization and tokenization methods.
Table~\ref{tab:methods_all} summarizes the acoustic features of representative methods.

\subsection{Evaluation Metric Computations}
\label{sec:appd_metrics}

The performance of controllable TTS often requires objective and subjective evaluation. We introduce common evaluation metrics in this subsection.

\textbf{Objective Evaluation Metrics.}
Objective metrics offer automated and reproducible evaluations. Mel Cepstral Distortion (MCD)~\cite{kominek2008synthesizer} measures the spectral distance between synthesized and reference speech, reflecting how closely the generated audio matches the target in terms of acoustic features.
A lower MCD value indicates a higher similarity between synthesized and reference speech, meaning better speech synthesis quality. Typically, an MCD value below 4 suggests good quality, while values above 6 may indicate significant distortion.
The MCD is computed as follows:
\begin{equation}
    MCD = \frac{10}{\ln 10} \cdot \sqrt{2 \sum_{d=1}^{D} (c_d^{(syn)} - c_d^{(ref)})^2},
\end{equation}
where $c_d^{(syn)}$ represents the d-th Mel Cepstral Coefficient (MCC) of the synthesized speech, $c_d^{(ref)}$ represents the d-th MCC of the reference speech, $D$ is the number of MCC, and $\frac{10}{\ln 10} \approx 4.342$ is a constant factor that converts the logarithm to a decibel scale.

Fréchet DeepSpeech Distance (FDSD)~\cite{Bińkowski2020High} is another metric designed to evaluate the quality and naturalness of synthesized speech. It is inspired by the Fréchet Inception Distance (FID)~\cite{heusel2017fid} used in image generation but adapted to speech by leveraging a deep speech recognition model.
FDSD measures the statistical distance between the distributions of real (reference) and synthesized speech in the feature space of a pretrained speech recognition model, such as Deep Speech~\cite{hannun2014deepspeech}. By comparing the mean and covariance of the extracted feature representations, FDSD provides a perceptually relevant assessment of speech synthesis quality.
A lower FDSD means the synthesized speech is more similar to real speech.
FDSD can be computed as:
\begin{equation}
    FDSD = ||\mu_s - \mu_r||^2 + \text{Tr}(\Sigma_s + \Sigma_r - 2(\Sigma_s \Sigma_r)^{1/2}),
\end{equation}
where $\mu_s$ and $\Sigma_s$ are the mean and covariance of the embeddings from the synthesized speech, $\mu_r$ and $\Sigma_r$ are the mean and covariance of the embeddings from the real (reference) speech, $||\mu_s - \mu_r||^2$ represents the squared Euclidean distance between the means, $\text{Tr}(\cdot)$ denotes the trace of a matrix, and $(\Sigma_s \Sigma_r)^{1/2}$ is the geometric mean of the covariance matrices.

For intelligibility, the Word Error Rate (WER)~\cite{enwiki2024wer} is used.
It measures the difference between the recognized transcript and the reference transcript by computing the number of errors made in the transcription process.
WER is computed as:
\begin{equation}
    WER = \frac{S + D + I}{N},
\end{equation}
where $S$ is the number of substitutions (wrong word in place of the correct word), $D$ is the number of deletions (missed words), $I$ is the number of insertions (extra words added), and $N$ is the total number of words in the reference transcript.

Cosine similarity (on speaker embeddings) measures similarity between the speaker embeddings of synthesized and reference speech. It can be used to evaluate zero-shot TTS (voice cloning) methods, where higher values indicate better speaker similarity.
Given two speaker embeddings, $\mathbf{e_1}$ and $\mathbf{e_2}$, their cosine similarity is defined as:
\begin{equation}
    CosSim(\mathbf{e_1}, \mathbf{e_2}) = \frac{\mathbf{e_1} \cdot \mathbf{e_2}}{\|\mathbf{e_1}\| \|\mathbf{e_2}\|},
\end{equation}
where speaker embeddings can be extracted from a pre-trained speaker embedding model (e.g., ECAPA-TDNN~\cite{Brecht20ECAPA-TDNN} and x-vectors~\cite{Snyder2018xvectors}).

Perceptual Evaluation of Speech Quality (PESQ)~\cite{Rix2001PESQ} is another objective metric designed to evaluate speech quality by comparing degraded audio with a clean reference.
It is widely used in telecommunications and speech synthesis.
PESQ models human auditory perception, producing a score in the range $[-0.5,-4.5]$ that reflects intelligibility and distortion under various conditions, including noise or compression.
PESQ involves complex perceptual modeling, its core components can be summarized as:
\begin{equation}
    PESQ = a_0 + a_1 \cdot D_{frame} + a_2 \cdot D_{time},
\end{equation}
where $D_{frame}$ is the frame-by-frame perceptual distortion, $D_{time}$ is the time-domain distortion, and $a_0, a_1, a_2$ are regression coefficients. One can refer to \cite{Rix2001PESQ} for details.

Signal-to-Noise Ratio (SNR) measures the ratio of signal power to noise power. A higher SNR indicates a cleaner signal with less noise, while a lower SNR suggests that noise is dominating the signal.
However, in TTS, noise can come from different sources, such as artifacts from vocoders, neural network distortions, or background noise in dataset recordings.
A direct computation of SNR in TTS requires a reference clean speech signal ($x[n]$), a synthesized (or noisy) speech signal ($y[n]$), and extracting the noise component ($e[n] = y[n] - x[n]$) from the synthesized signal.
The SNR for TTS systems can be computed as:
\begin{equation}
    SNR = 10 \log_{10} \left( \frac{P_{\text{signal}}}{P_{\text{noise}}} \right),
\end{equation}
where $P_{\text{signal}} = \frac{1}{N} \sum_{n=1}^{N} x[n]^2$ and $P_{\text{noise}} = \frac{1}{N} \sum_{n=1}^{N} e[n]^2$.

\textbf{Subjective Evaluation Metrics.}
The Mean Opinion Score (MOS)~\cite{enwiki2024mos} is the most commonly used subjective metric. In MOS evaluations, listeners rate various aspects, such as naturalness, expressiveness, quality, intelligibility, et al., of synthesized speech on a scale from 1 to 5, where higher scores indicate better quality.
MOS captures human perception effectively, but is expensive for large-scale evaluations.

Comparison Mean Opinion Score (CMOS)~\cite{loizou2011speech} further evaluates relative quality differences between two TTS audio samples.
Participants listen to paired samples and rate their preference on a scale (e.g., -3 to +3, where negative values favor the first sample).
CMOS is used to measure subtle improvements in TTS systems, complementing absolute MOS ratings.
MOS and CMOS scores are computed as the average scores across all listeners:
\begin{equation}
    MOS/CMOS = \frac{1}{N} \sum_{i=1}^{N} s_i,
\end{equation}
where $s_i$ is the score given by the $i$-th listener, and $N$ is the number of listeners.

AB and ABX tests are also popular in evaluating TTS methods.
An AB test involves presenting two versions of a synthesized speech (from different TTS models) to human listeners and asking them to choose which they prefer. The goal is to assess which model produces better-sounding speech based on certain criteria, such as naturalness, intelligibility, or clarity.
In an ABX test, listeners compare two synthesized speech samples to a reference speech sample and determine which one is closer in terms of timbre, prosody, emotion, and other relevant features. ABX tests are widely used in evaluating zero-shot TTS methods.
The AB/ABX test score for a model $m$ is:
\begin{equation}
    Score_{AB}/Score_{ABX} = \frac{N_m}{N},
\end{equation}
where $N_m$ represents the number of listeners who prefer the speech synthesized by model $m$, and $N$ denotes the total number of listeners.

Table~\ref{tab:eval_metrics} summarizes widely used metrics for TTS.

\subsection{A Google Gemini-Based Experimental Evaluation of TTS Controllability}
\label{sec:evaluation}

We designed an evaluation pipeline using Gemini to assess synthesized speech in terms of \textbf{instruction following}, \textbf{naturalness}, and \textbf{expressiveness}, because these dimensions are not well captured by traditional metrics. Conventional scores like MCD, WER, PESQ, speaker similarity, and MOS/CMOS are excluded, as our goal is to explore the feasibility of using multimodal large language models (MLLMs) as subjective evaluators.

\subsubsection{Implementation details}

\textbf{Models.}
Due to time constraints, we only evaluate a total of 10 models: 8 open-source systems (F5-TTS, CosyVoice, CosyVoice2, Vevo, SparkTTS, MaskGCT, PromptTTS, and VoxInstruct) and 2 commercial TTS systems (ElevenLabs and MiniMax TTS).

\textbf{Tasks.}
Zero-shot TTS and description-based synthesis. For each model, we synthesize 20 speech samples (10 in English and 10 in Chinese) for each task.

\begin{figure*}[t]
    \centering
    \includegraphics[width=\linewidth]{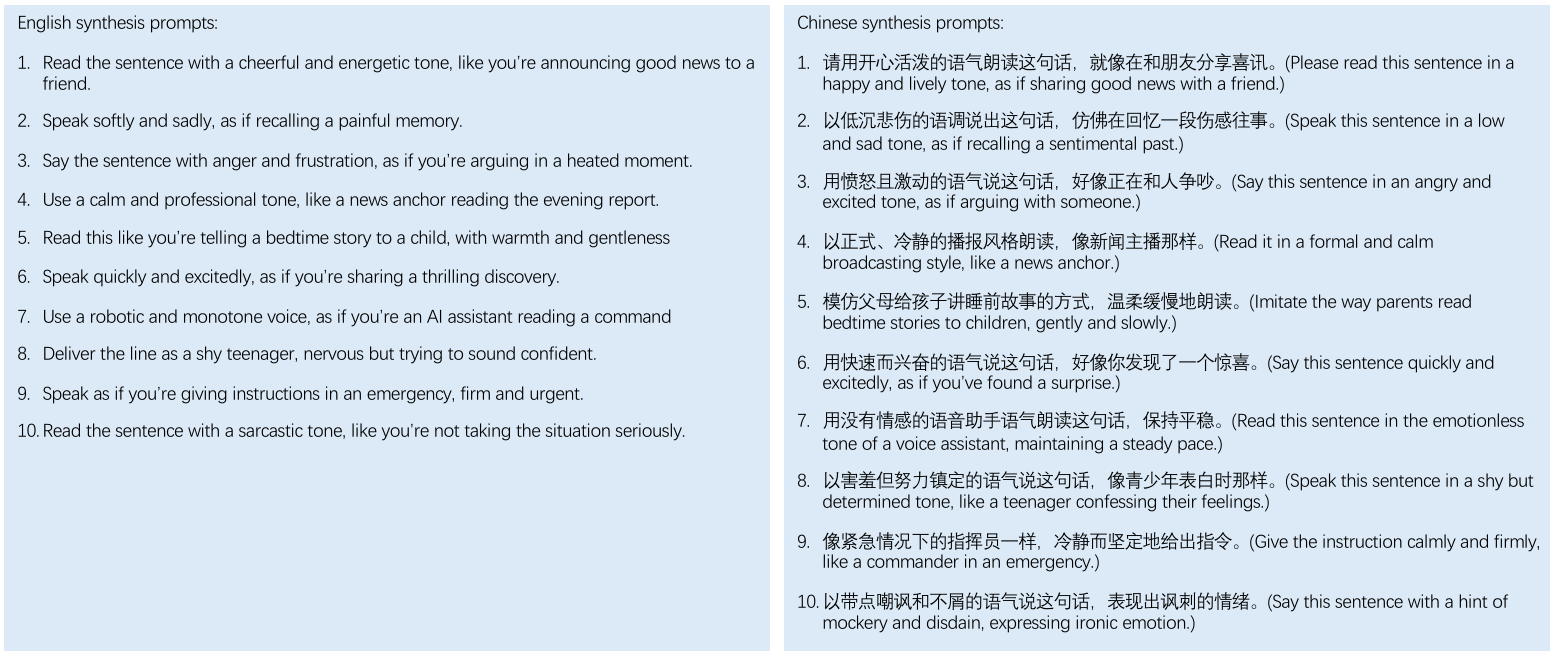}
    \caption{Textual descriptions generated by ChatGPT}
    \label{fig:descriptions}
\end{figure*}

\textbf{Dataset.}
For zero-shot TTS, we sampled 10 English utterances from the MSP-Podcast dataset and 10 Chinese utterances from Emo-Emilia to serve as reference speech prompts. For description-based synthesis, we used ChatGPT to generate diverse textual descriptions as shown in Fig.~\ref{fig:descriptions}.

\textbf{Metrics Clarification}:

\begin{itemize}
    \item Instruction Following
    \begin{itemize}
        \item Purpose: To assess how accurately the synthesized audio follows the given instruction regarding speech characteristics such as timing, emphasis, and pacing.
        \item Focus: Measures the controllability and fidelity of the model in executing user-specified directives.
    \end{itemize}
    \item Naturalness
    \begin{itemize}
        \item Purpose: To evaluate how natural the audio sounds—whether it resembles human speech or exhibits synthetic, robotic qualities.
        \item Focus: Measures the perceptual audio quality and realism of the synthesized speech.
    \end{itemize}
    \item Expressiveness
    \begin{itemize}
        \item Purpose: To judge the emotional richness and prosodic variation in the audio, such as tone, intensity, and nuance.
        \item Focus: Measures the model’s ability to convey expressive and emotionally engaging speech.
    \end{itemize}
\end{itemize}

\textbf{Gemini Prompts.}

The prompt we use for the evaluation is as follows:

\textit{SYSTEM\_PROMPT}: You are a strict quality evaluator for synthesized speech. Given an audio file of a speech sample, its transcript, and an instruction describing the intended speech characteristics, please rate the audio based on the following three aspects, using the defined criteria. Output ONLY a JSON dictionary with the keys instruction\_following, naturalness, and expressiveness, each assigned an integer value from 1 to 5. The evaluation rubrics are as follows:

\begin{enumerate}
    \item \textbf{Instruction Following (1--5):}
    \begin{itemize}
        \item 1 point: The audio completely ignores the instruction; it does not follow the intended timing, emphasis, or pacing. 
        \item 2 points: It loosely follows the instructions but misses key elements or timing in parts. 
        \item 3 points: Generally follows the instruction with minor lapses in emphasis or pacing. 
        \item 4 points: Clearly follows the instruction with only slight deviations. 
        \item 5 points: Perfectly follows every aspect of the instruction with clear emphasis and precise pacing.
    \end{itemize}
    
    \item \textbf{Naturalness (1--5):}
    \begin{itemize}
        \item 1 point: The audio sounds fully synthetic or robotic; extremely unnatural. 
        \item 2 points: Noticeably synthetic; some unnatural artifacts remain. 
        \item 3 points: Moderately natural with occasional synthetic artifacts. 
        \item 4 points: Largely natural sounding with minor imperfections. 
        \item 5 points: Completely natural; indistinguishable from a human recording.
    \end{itemize}
    
    \item \textbf{Expressiveness (1--5):}
    \begin{itemize}
        \item 1 point: The audio is flat and monotone; no emotional variation. 
        \item 2 points: Minimal expressiveness; emotions are weak or inconsistent. 
        \item 3 points: Reasonably expressive with some highlights, but could be stronger. 
        \item 4 points: Clearly expressive with only slight under- or over-emphasis. 
        \item 5 points: Exceptionally expressive; full emotional richness and nuance.
    \end{itemize}
\end{enumerate}

\textit{USER\_PROMPT}: The synthesized speech is \{audio\}. The transcript of the audio is: ``\{transcript\}". The instruction for the audio is: ``\{instruction\}".

\begin{table*}[t]
    \centering
    \resizebox{1.0\textwidth}{!}{
    \begin{tabular}{ccccc}
        \toprule
        Task & Method & \makecell{Instruction Following\\ (Gemeni 2.5 Flash / Human)} & \makecell{Naturalness\\ (Gemeni 2.5 Flash / Human)} & \makecell{Expressiveness\\ (Gemeni 2.5 Flash / Human)} \\ 
        \midrule
        \multirow{6}{*}{Zero-shot} & F5-TTS & - & 4.27±0.87 & 4.21±0.78 \\ 
        ~ & CosyVoice & - & 4.20±0.82 & 4.17±0.83 \\ 
        ~ & CosyVoice 2 & - & 4.25±0.58 & 4.20±0.63 \\ 
        ~ & Vevo & - & 4.43±0.55 & 4.32±0.75 \\ 
        ~ & SparkTTS & - & 3.68±0.80 & 3.83±0.79 \\ 
        ~ & MaskGCT & - & 3.91±0.88 & 4.08±0.81 \\ 
        \midrule
        \multirow{6}{*}{Instruction-based} & CosyVoice & 4.81±0.28 & 4.92±0.24 & 4.78±0.29 \\ 
        ~ & CosyVoice 2 & 4.61±0.49 & 4.85±0.31 & 4.63±0.52 \\ 
        ~ & EmoVoice & 4.67±0.44 & 4.80±0.36 & 4.67±0.44 \\ 
        ~ & VoxInstruct & 4.45±0.50 & 4.83±0.32 & 4.50±0.52 \\ 
        ~ & ElevenLabs & 4.52±0.67 & 4.85±0.31 & 4.63±0.57 \\ 
        ~ & MiniMax TTS & 4.67±0.36 & 4.87±0.27 & 4.63±0.44 \\ 
        \bottomrule
    \end{tabular}
    }
    \caption{The evaluation of the controllability of open-source and commercial TTS systems.}
    \label{tab:eval_results}
\end{table*}

\begin{table}[t]
    \centering
    \resizebox{1.0\linewidth}{!}{
    \begin{tabular}{cccc}
    \toprule
        ~ & Instruction Following & Naturalness & Expressiveness \\ 
        \midrule
        NISQA~\cite{mittag21_interspeech} & - & 0.01 & -0.03 \\ 
        UTMOS~\cite{saeki22c_interspeech} & - & -0.10 & -0.17 \\ 
        Ours & 0.12 & 0.17 & 0.14 \\ 
        \bottomrule
    \end{tabular}
    }
    \caption{The alignment between model-based evaluation and human preference.}
    \label{tab:result_alignment}
\end{table}

\subsubsection{Results: Model-level Performance Comparison}

As shown in Table~\ref{tab:eval_results}, in the zero-shot setting, among the six models, Vevo performs best in both naturalness (4.43±0.55) and expressiveness (4.32±0.75), indicating strong general quality without explicit guidance. CosyVoice, CosyVoice 2, and F5-TTS follow closely with similar scores (~4.2), while SparkTTS and MaskGCT lag behind, especially in naturalness.

In the instruction-based setting, all models show a clear improvement across all metrics. CosyVoice achieves the highest overall scores, with instruction following at 4.81±0.28, naturalness at 4.92±0.24, and expressiveness at 4.78±0.29. Other strong performers include MiniMax TTS and EmoVoice, both exceeding 4.6 in most dimensions. Even the lowest-scoring instruction-based method (VoxInstruct) outperforms the best zero-shot model in every aspect.

\subsubsection{Results: The Reliability of Multimodal LLM-based Evaluation}

We also compare the proposed metrics with existing automated evaluation methods, namely NISQA~\cite{mittag21_interspeech} and UTMOS~\cite{saeki22c_interspeech}. Specifically, we use 96 synthesized samples to compute the Pearson correlation coefficients between the predicted scores from each method and human ratings, aiming to assess how well each method aligns with human perception. 

As shown in Table~\ref{tab:result_alignment}, although the absolute Pearson correlation coefficients of our method are relatively modest, our approach consistently outperforms both NISQA and UTMOS across all three evaluation dimensions: instruction following, naturalness, and expressiveness.

These results suggest that existing automated metrics like NISQA and UTMOS, which are primarily designed for general speech quality assessment, may not capture nuanced attributes such as speaker intent or expressive delivery in controllable TTS tasks. 

In contrast, our metric, tailored for instruction-based synthesis evaluation, better reflects human judgments, particularly in aspects beyond raw audio quality. This supports the need for task-specific evaluation frameworks when benchmarking modern controllable TTS systems.

\subsubsection{Conclusion}

To some extent, the proposed MLLM-based evaluation pipeline is able to predict human-aligned scores for instruction following, naturalness, and expressiveness. We also find that it offers promising potential for automated evaluation of controllable TTS.

In future work, we plan to enhance our survey by designing a more robust and reliable MLLM-based evaluation framework and conducting a comprehensive benchmark of existing controllable TTS methods.
 
\end{document}